\documentclass{article}

\PassOptionsToPackage{numbers, sort&compress}{natbib}
\usepackage[preprint]{neurips_2026}

\usepackage[utf8]{inputenc} 
\usepackage[T1]{fontenc}    
\usepackage{url}            
\usepackage{booktabs}       
\usepackage{amsfonts}       
\usepackage{nicefrac}       
\usepackage{microtype}      
\usepackage{xcolor}         
\usepackage{bm}
\usepackage{graphicx}
\usepackage[table]{xcolor} 
\usepackage{wrapfig}
\usepackage{amsmath}
\usepackage{siunitx}
\usepackage{multirow}
\usepackage{subcaption}
\usepackage{lipsum}
\usepackage{wrapfig}
\usepackage{pifont}
\usepackage{algorithm}
\usepackage{algpseudocode}
\usepackage{bbm}
\usepackage{float}
\usepackage{placeins}
\usepackage[most]{tcolorbox}
\usepackage[pagebackref=true]{hyperref}

\usepackage{multirow}
\usepackage{adjustbox}
\definecolor{customblue}{HTML}{4169E1}
\hypersetup{
    colorlinks=true,   
    citecolor=customblue,    
    linkcolor=customblue,    
    urlcolor=customblue      
}

\definecolor{checkgreen}{RGB}{34,139,34}
\definecolor{crossred}{RGB}{200,40,40}
\definecolor{partialorange}{RGB}{230,140,20}

\newcommand{\cmark}{\textcolor{checkgreen}{\ding{51}}}
\newcommand{\xmark}{\textcolor{crossred}{\ding{55}}}
\newcommand{\pmark}{\textcolor{partialorange}{$\sim$}}

\newcommand{\bcmark}{\textbf{\textcolor{checkgreen}{\ding{51}}}}

\definecolor{bestblue}{HTML}{FCD0E9}      
\definecolor{secondblue}{RGB}{148, 189, 227}   
\definecolor{thirdblue}{RGB}{184, 228, 255}    

\newcommand{\best}[1]{\multicolumn{1}{c}{\cellcolor{bestblue}\textbf{#1}}}

\title{Fixed-Point Masked Generative Modeling}

\author{%
\begingroup
\setlength{\tabcolsep}{5pt}
\renewcommand{\arraystretch}{0.92}
\small
\begin{tabular}{@{}c@{\hspace{0.35cm}}c@{\hspace{0.35cm}}c@{\hspace{0.35cm}}c@{\hspace{0.35cm}}c@{}}
{\bfseries Andrea Miele\thanks{Correspondence to andrea.miele.pro@gmail.com. Code available at \href{https://github.com/andreamiele/fp-mgm/}{https://github.com/andreamiele/fp-mgm/}.}}
&
{\bfseries Yiming Qin}
&
{\bfseries Alba Carballo-Castro}
&
{\bfseries Justin Deschenaux}
&
{\bfseries Pascal Frossard}
\\[-1pt]
{\normalfont\mdseries\footnotesize LTS4, EPFL}
&
{\normalfont\mdseries\footnotesize LTS4, EPFL}
&
{\normalfont\mdseries\footnotesize LTS4, EPFL}
&
{\normalfont\mdseries\footnotesize CLAIRE, EPFL}
&
{\normalfont\mdseries\footnotesize LTS4, EPFL}
\end{tabular}
\endgroup
}

\newcommand{\mask}{[\textsc{mask}]~}

\begin{document}

\maketitle

\begin{abstract}
Masked Generative Models (MGMs) enable parallel decoding and achieve strong performance across modalities, but require full-sequence bidirectional transformers at every step, making training costly and degrading quality under low sampling budgets.
Existing work improves efficiency via better samplers or cheaper fixed-depth denoisers, but they still allocate a fixed amount of denoiser computation to each refinement step.
We introduce Fixed-Point Masked Generative Models (FP-MGMs), which replace part of the denoiser with a fixed-point solver over shared attention layers to enable adaptive depth with fewer parameters.
To make it more effective for masked generation, we first introduce a cross-step consistency loss, which aligns hidden representations at neighboring denoising steps and, second, three-state reuse (3SR) which warm-starts the solver using the previous solution by treating differently unchanged, still-masked, and newly revealed tokens respectively. Together, these components define our complete training-to-inference framework for fixed-point masked generation, \emph{CoFRe}.
We also show that pre-trained MGMs can be converted into FP-MGMs with short fine-tuning, avoiding full retraining.
Across modalities, CoFRe improves the quality and cost trade-off. On OpenWebText, CoFRe reduces parameters by 38.8\%, training time by 11.5\%, and VRAM by 16.9\%, while improving generative perplexity from 830.8 to 101.8 at a budget of $96$ transformer-block forward passes, compared to MDLM.
In ImageNette, CoFRe reduces training time by 48.6\% and VRAM by 50.7\%, while improving FID in all sample budgets tested. Overall, CoFRe offers a practical framework for cheaper training and stronger low-budget masked generation.

\end{abstract}

\vspace{-0.3em}
\begin{figure}[htbp]
    \centering
    \begin{subfigure}[b]{0.48\textwidth}
        \centering
        \includegraphics[width=\textwidth]{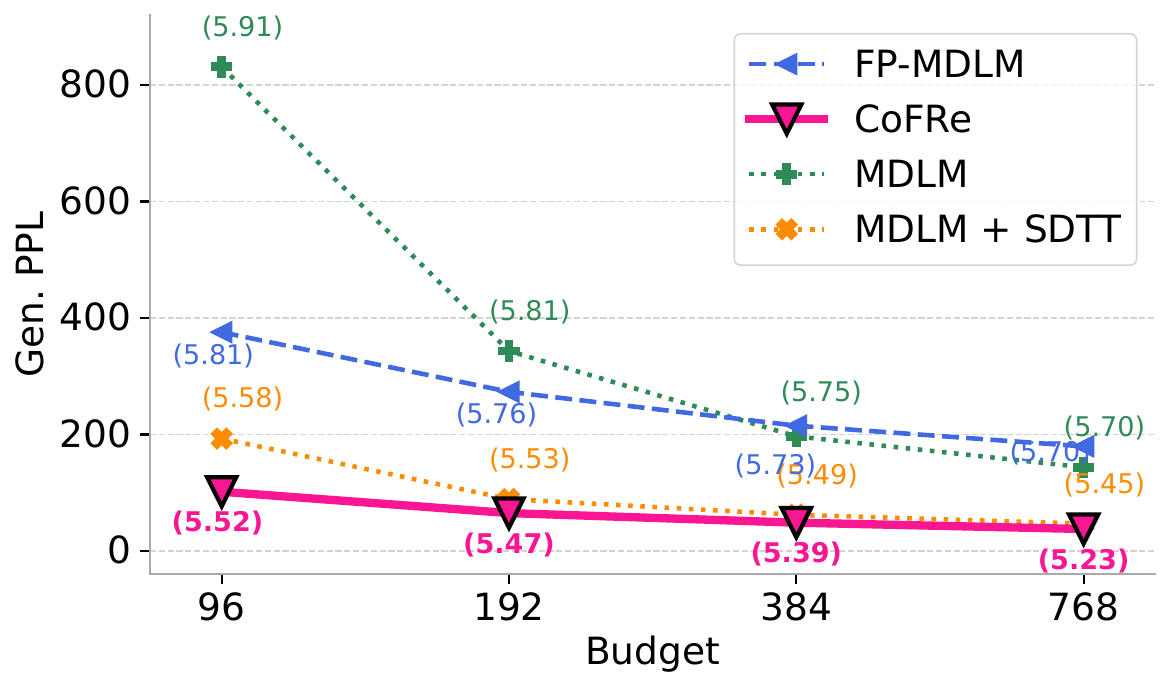}

        \label{fig:consistency_left}
    \end{subfigure}
    \hfill 
    \begin{subfigure}[b]{0.48\textwidth}
        \centering
        \includegraphics[width=\textwidth]{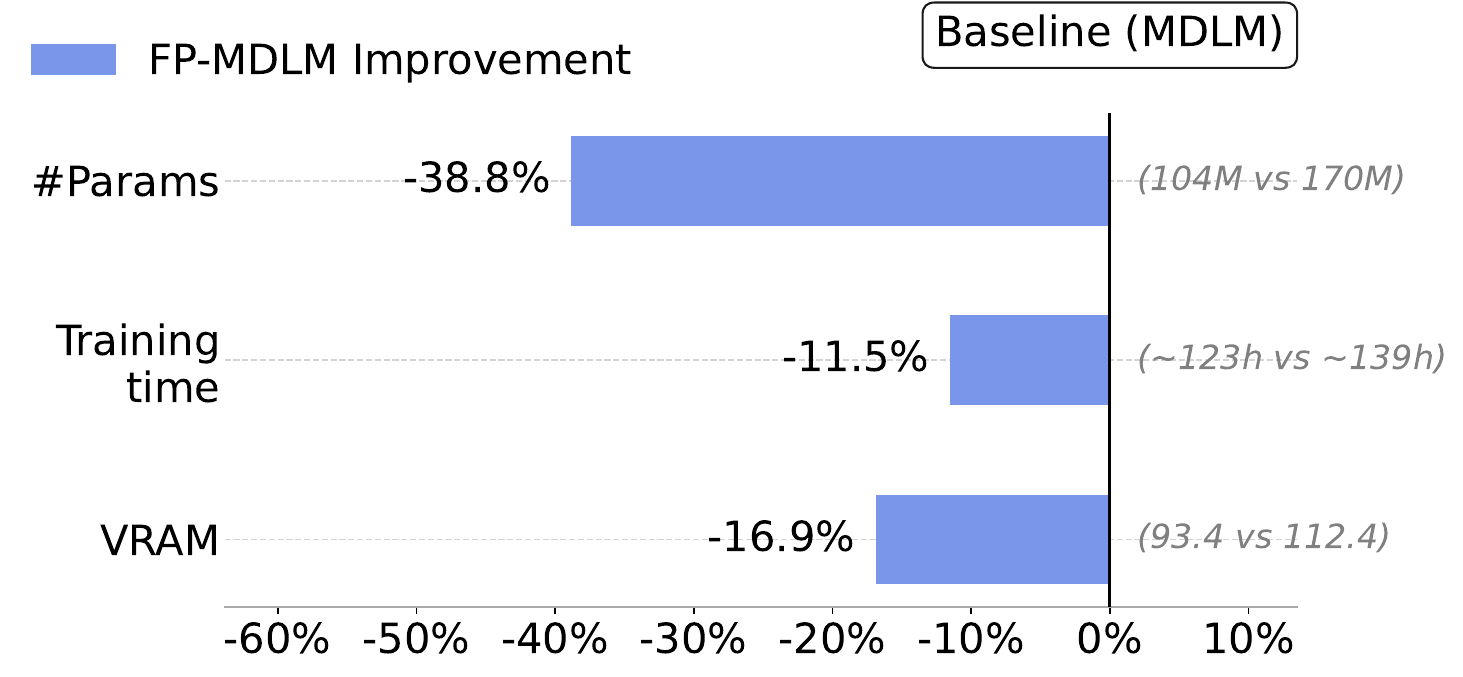}

        \label{fig:consistency_right}
    \end{subfigure}
    \caption{\textbf{FP-MDLM and CoFRe improve the quality--cost trade-off on OWT.}
(Left) Generative perplexity across forward-pass budgets, with entropy in parentheses; CoFRe gives the best quality at all shown budgets.
(Right) Relative to MDLM, FP-MDLM and CoFRe use fewer parameters, less training time, and less VRAM.}
    \label{fig:consistency}
    \vspace{-10pt}
\end{figure}

\section{Introduction}
\label{sec:introduction}

\begin{table}[htbp]
\centering
\normalsize
\vspace{-5pt}
\setlength{\tabcolsep}{4pt}
\renewcommand{\arraystretch}{1.05}
\setlength{\abovecaptionskip}{3pt}
\setlength{\textfloatsep}{3pt}

\resizebox{0.99\linewidth}{!}{%
\begin{tabular}{@{}l
>{\centering\arraybackslash}p{0.16\linewidth}
>{\centering\arraybackslash}p{0.16\linewidth}
>{\centering\arraybackslash}p{0.20\linewidth}
@{}}
\toprule
\textbf{Method family} 
& \shortstack{\textbf{Adaptive network}\\\textbf{depth}} 
& \shortstack{\textbf{Cheaper}\\\textbf{training}} 
& \shortstack{\textbf{Strong low-budget}\\\textbf{generation}} \\
\midrule
\shortstack[l]{Sampler improvements\\[-1pt]
\scriptsize\citep{besnierHaltonSchedulerMasked2025, pengPathPlanningMasked2025a, hayakawaDemystifyingMaskGITSampler2025, chenMaskedDiffusionModels2026, wang2025remasking, heReasoningLatentTokens2026}}
& \xmark & \xmark & \cmark \\

\shortstack[l]{Efficient fixed-depth architectures\\[-1pt]
\scriptsize\citep{deschenauxPartitionGenerativeModeling2026, youEffectiveEfficientMasked2026}}
& \xmark & \pmark & \xmark \\

\shortstack[l]{Controllable-depth / looped / DEQ-style models\\[-1pt]
\scriptsize\citep{giannou2023loopedtransformersprogrammablecomputers, zhang2026expertchoiceroutingenablesadaptive, dehghaniUniversalTransformers2019, baiDeepEquilibriumModels2019, Prairie2026ParcaeSL, jolicoeurmartineau2025morerecursivereasoningtiny}}
& \cmark & \pmark & \xmark \\

\rowcolor{bestblue} \textbf{Ours: FP-MGMs} 
& \bcmark & \bcmark & \bcmark \\

\bottomrule
\end{tabular}
}
\caption{Comparison of prior approaches and FP-MGMs. \pmark\ denotes partial coverage. We only focus here on discrete data.}
\label{tab:limitations_comparison}

\vspace{-1em}
\end{table}

Masked generative models (MGMs) generate sequences by iteratively denoising masked tokens, enabling parallel decoding and strong generation quality across modalities. Prominent examples include MDLMs for language \citep{sahoo_simple_2024, shiSimplifiedGeneralizedMasked2025, ou2026absorbingdiscretediffusionsecretly} and MaskGIT for images \citep{changMaskGITMaskedGenerative2022}, with other related masked-generation approaches extending to video, audio, and multimodal generation \citep{yuMAGVITMaskedGenerative2023, villegasPhenakiVariableLength2022, comunitaSpecMaskGITMaskedGenerative2024, mizrahi4MMassivelyMultimodal2023}.
However, MGMs are computationally expensive \citep{sedykhNotAllDenoising2026, deschenauxPartitionGenerativeModeling2026}, since each refinement step runs a full bidirectional transformer pass over the entire sequence. Training therefore consumes large amounts of VRAM and is notably slow. Furthermore, sampling under a low compute budget, i.e., the total number of transformer-block forward passes, produces poor-quality samples 
\citep{deschenaux2024promisesoutlookschallengesdiffusion}. Thus, improving MGMs requires controlling not only the number of denoising steps, but also the cost and effective depth of the denoiser passes at each denoising step.

Prior work addresses these issues and improves MGMs efficiency along three main axes. First, alternative samplers improve sample quality at fixed compute by changing which tokens are revealed or updated at each refinement step \citep{besnierHaltonSchedulerMasked2025, pengPathPlanningMasked2025a, hayakawaDemystifyingMaskGITSampler2025, chenMaskedDiffusionModels2026, wang2025remasking, heReasoningLatentTokens2026}. Second, efficient but fixed-depth architectures reduce architectural waste, for example by avoiding computation on \mask tokens \citep{deschenauxPartitionGenerativeModeling2026}, or by designing more efficient masked generative backbones \citep{youEffectiveEfficientMasked2026}. Finally, adaptive routing and looped transformers, and DEQ-style models provide controllable effective depth by repeatedly applying shared modules or dynamically allocating computation \citep{giannou2023loopedtransformersprogrammablecomputers, zhang2026expertchoiceroutingenablesadaptive, dehghaniUniversalTransformers2019, baiDeepEquilibriumModels2019, Prairie2026ParcaeSL, jolicoeurmartineau2025morerecursivereasoningtiny}.

However, these directions leave an important gap for masked generative modeling, as detailed in Tab.~\ref{tab:limitations_comparison}. Sampler improvements change the denoising trajectory but usually leave the training procedure and per-step denoiser unchanged, so the compute spent at each sampling step remains fixed independently of step difficulty. Efficient, fixed-depth masked architectures reduce compute, but still rely on a denoiser with fixed-capacity. This suggests that improving MGM efficiency not only requires changing the sampler, but also acting inside the denoiser: the model should reuse parameters to reduce backbone cost, while still allowing different refinement steps to use different amounts of computation.
Fixed-point / DEQ-style layers provide a natural substrate for this goal: they replace an explicit stack of distinct layers with repeated applications of a shared block, whose equilibrium is used as the layer output~\citep{baiDeepEquilibriumModels2019}. Fixed-Point Diffusion Models further show that this idea can improve the quality--cost trade-off in continuous diffusion denoisers~\citep{bai_fixed_2024}. Yet masked generation introduces discrete, token-wise state changes, so continuous-diffusion techniques do not directly transfer and can be suboptimal. This motivates our FP-MGM framework, \emph{CoFRe}, which combines fixed-point denoisers with additional mechanisms tailored to masked generation: cross-step consistency and token-aware three-state reuse.

\paragraph{Contributions}
We organize our contributions around three claims.

\textbf{(1) Fixed-point masked denoisers improve the quality-cost trade-off.}
Compared to using a fixed stack of distinct transformer layers, a fixed-point layer repeatedly applies a shared block and treats the resulting equilibrium as the layer output. Since the same block is reused across solver iterations, the model can increase or decrease its effective depth by changing the number of iterations, without adding parameters. We introduce Fixed-Point Masked Generative Models (FP-MGMs), which replace the middle layers of a masked denoiser with a weight-sharing fixed-point block. We apply this approach to two representative MGMs: MDLM \citep{sahoo_simple_2024} for text, yielding FP-MDLM, and MaskGIT \citep{changMaskGITMaskedGenerative2022} for images, yielding FP-MaskGIT. In both cases, the fixed-point denoiser reduces the number of parameters, training time, and memory, while allowing the effective denoiser depth to vary through the number of fixed-point iterations. This addresses limitations (1) and (2) (Fig.~{\ref{fig:consistency}}, Right).

\textbf{(2) Three-state reuse makes FP-MDLM practical.}
Warm-starting from a fixed-point solution from the last denoising step is a standard way to make FP models efficient.
However, in MGMs, the input sequence changes abruptly across refinement steps as tokens are revealed or replaced. As a result, the previous fixed-point solution is not equally reliable across positions.
We therefore introduce a three-state reuse rule, referred to as \textit{3SR}, that treats unchanged visible tokens, still-masked tokens, and newly revealed tokens differently.
By design, unchanged visible tokens fully reuse the previous fixed-point solution, still-masked tokens partially reuse it, and newly revealed tokens rely more on the current input injection.

\textbf{(3) Cross-time regularization is key for strong low-budget generation.}
Architecture and reuse alone are not sufficient for strong low-budget sampling: abrupt changes in the input space also induce non-smooth changes in the representation space across denoising steps.
We introduce $\mathcal{L}_{\mathrm{CONS}}$, a cross-step consistency loss that aligns the representations of a noisier student state and a cleaner teacher state. Empirically, this loss behaves like cross-time self-distillation, sharpening masked-token predictions and driving most of the low-budget generation gains. This addresses limitation (3) (Fig.~{\ref{fig:consistency}}, Left). Together, the fixed-point denoiser, cross-step consistency, and 3SR define \emph{CoFRe}: a complete training-to-inference recipe.
Finally, we also show that pretrained MDLM checkpoints can be converted into FP-MDLMs with a short distillation stage, avoiding full retraining from scratch. These components improve low-budget sampling and reduce the cost of obtaining strong FP-MDLM checkpoints, addressing limitations (2) and (3).

We evaluate FP-MDLM on OpenWebText (OWT) \citep{Gokaslan2019OpenWeb} and downstream tasks \citep{eval-harness}, and FP-MaskGIT on ImageNette \citep{howard2019imagenette, ImageNet}. On OWT, FP-MDLM reduces parameters by 38.8\%, training time by 11.5\%, and VRAM by 16.9\% relative to MDLM, while improving low-budget generative perplexity from 830.8 to 375.6 at budget 96. With cross-step consistency and 3SR, our model CoFRe further improves over MDLM+SDTT \citep{deschenaux2024beyond} in the low-budget regime, reducing generative perplexity from 193.1 to 101.8 at budget 96 and from 47.0 to 37.8 at budget 768.
On ImageNette, CoFRe reduces training time by 48.6\% and VRAM by 50.7\% relative to MaskGIT-Large, while improving FID across all reported budgets.
We also show that a pretrained MGM can be efficiently converted into a more compute-efficient FP architecture: improving over the 1M-step FP-MDLM baseline at every sampling budget, with only $4\%$ of the original pretraining steps. Overall, CoFRe makes masked generative models cheaper to train, easier to adapt, and stronger under limited sampling budgets.

\section{Background}
\label{sec:background}
In this paper, we use $\mathbf{x}=(x^1,\dots,x^d)\in\mathcal{V}^d$ to denote a clean sequence of length $d$ over vocabulary $\mathcal{V}:=[V]$. We denote by $\mathbf{z}_t$ a corrupted version of $\mathbf{x}$ at noise level or refinement state $t$, and by \mask the special mask token. A masked denoiser maps $(\mathbf{z}_t,t)$ to token logits $\boldsymbol{\ell}_\theta(\mathbf{z}_t,t)$, from which predictions or samples are obtained.

\subsection{Discrete generative models and masked generative models}

Discrete generative models learn distributions over sequences of categorical variables, such as text tokens or quantized image latents. 

In this work, we focus on MGMs, which corrupt data by replacing tokens with a special mask token \mask and train a denoiser parameterized by $\theta$ to recover the clean sequence. The denoiser outputs logits $\boldsymbol{\ell}_\theta(\mathbf{z}_t,t)$, and the training objective is
\begin{equation}
\label{eq:loss_mgm}
\mathcal{L}_{\text{MGM}}
:=
\mathbb{E}_{\mathbf{x}\sim \mathcal{D},\, t\sim\mathcal{U}[0,1]}
\left[
w(t)\,
\mathrm{CE}_{\mathcal{M}_t}
\left(
\boldsymbol{\ell}_\theta(\mathbf{z}_t,t),
\mathbf{x}
\right)
\right].
\end{equation}

where $w:[0,1]\to \mathbb{R}^+$ is a weighting function, 
$\mathcal{M}_t=\{i:z_t^i=\mask\}$ is the set of masked positions, 
and $\mathrm{CE}_{\mathcal{M}_t}(\boldsymbol{\ell}_\theta(\mathbf{z}_t,t),\mathbf{x})$ denotes the cross-entropy between the predicted logits and the clean tokens, evaluated only on positions in $\mathcal{M}_t$.
Sampling starts from a fully masked sequence and iteratively reveals subsets of tokens using repeated denoiser evaluations.

\paragraph{MDLM} 
MDLMs \citep{sahoo_simple_2024, shiSimplifiedGeneralizedMasked2025, ou2026absorbingdiscretediffusionsecretly} instantiate MGMs for language. MDLMs define an absorbing-state discrete diffusion process in which clean tokens are independently replaced by \mask tokens according to a time-dependent noise schedule. Training corresponds to the MGM objective in Eq.~\ref{eq:loss_mgm} with the MDLM weighting $w(t)=\frac{\alpha_t'}{1-\alpha_t}$, where $\alpha_t$ is the noise schedule giving the probability that a token remains clean at time $t$, using the notation of \citet{sahoo_simple_2024}.
To keep predictions coherent across noise levels during accelerated sampling, auxiliary temporal objectives such as consistency regularization or self-distillation through time (SDTT) \citep{deschenaux2024beyond} are commonly used.

\paragraph{MaskGIT}
MaskGIT \citep{changMaskGITMaskedGenerative2022} is a MGM for image tokens, typically operating in the latent space of a pretrained tokenizer. Unlike MDLM, which follows a diffusion-time masking process, MaskGIT relies on confidence-based iterative decoding: at each step, it predicts masked tokens, scores their confidence, and permanently reveals a subset of high-confidence positions. Following \citet{besnierHaltonSchedulerMasked2025}, we use Halton low-discrepancy schedules \citep{haltonRadicalInverse1964} to obtain more uniform spatial coverage; further details are given in Appendix~\ref{appendix:halton}.

\subsection{Deep equilibrium models, and fixed-point diffusion models}

A fixed point of a map $F_\theta$ is a state $\mathbf{h}^\star$ that remains unchanged after applying the map:
$
\mathbf{h}^\star = F_\theta(\mathbf{h}^\star; \mathbf{u}),
$
where $\mathbf{u}$ is an external input. 
An $n_\text{th}$ fixed-point layer uses this equilibrium state as its output, and approximates it by iterating a shared transformation,
$
\mathbf{h}^{n+1} = F_\theta(\mathbf{h}^n; \mathbf{u}),
$
or by using a numerical solver such as Broyden's method or Anderson acceleration. 
This can be viewed as a weight-sharing network whose effective depth is controlled by the number of solver iterations.

Deep Equilibrium Models (DEQs) \citep{baiDeepEquilibriumModels2019} use this principle to define hidden representations implicitly, with gradients computed by implicit differentiation or approximate Jacobian-free methods. 
Fixed-Point Diffusion Models (FPDMs)~\citep{bai_fixed_2024} have shown that fixed-point denoisers can improve the quality--cost trade-off in continuous diffusion by replacing part of the denoiser with an implicit weight-sharing layer. At each diffusion timestep, the denoiser solves a fixed-point problem over hidden representations, and nearby timesteps often have similar solutions, enabling efficient warm starts from previous fixed-point states. This makes fixed-point layers a natural substrate for parameter-efficient and controllable-depth denoisers.
Directly adapting FPDMs to masked generation is not sufficient, however. Masked refinement changes the input state discretely and token-wise: some positions remain visible, some remain masked, and others are newly revealed or replaced. Thus, previous fixed-point solutions are not uniformly reusable, and low-budget generation can still suffer from cross-step representation drift. We therefore introduce FP-MGMs together with CoFRe, a complete training-to-inference framework that adds cross-step consistency and token-aware three-state reuse to fixed-point masked denoisers.

\section{Fixed-Point Denoising Networks for Masked Sequence Modeling}
\label{sec:method}
Having introduced masked generative models and the fixed-point perspective in the previous sections, we now describe how to combine them into controllable-depth denoisers designed specifically for discrete masked generation.

\begin{figure}[t]
    \centering
    \includegraphics[width=\textwidth]{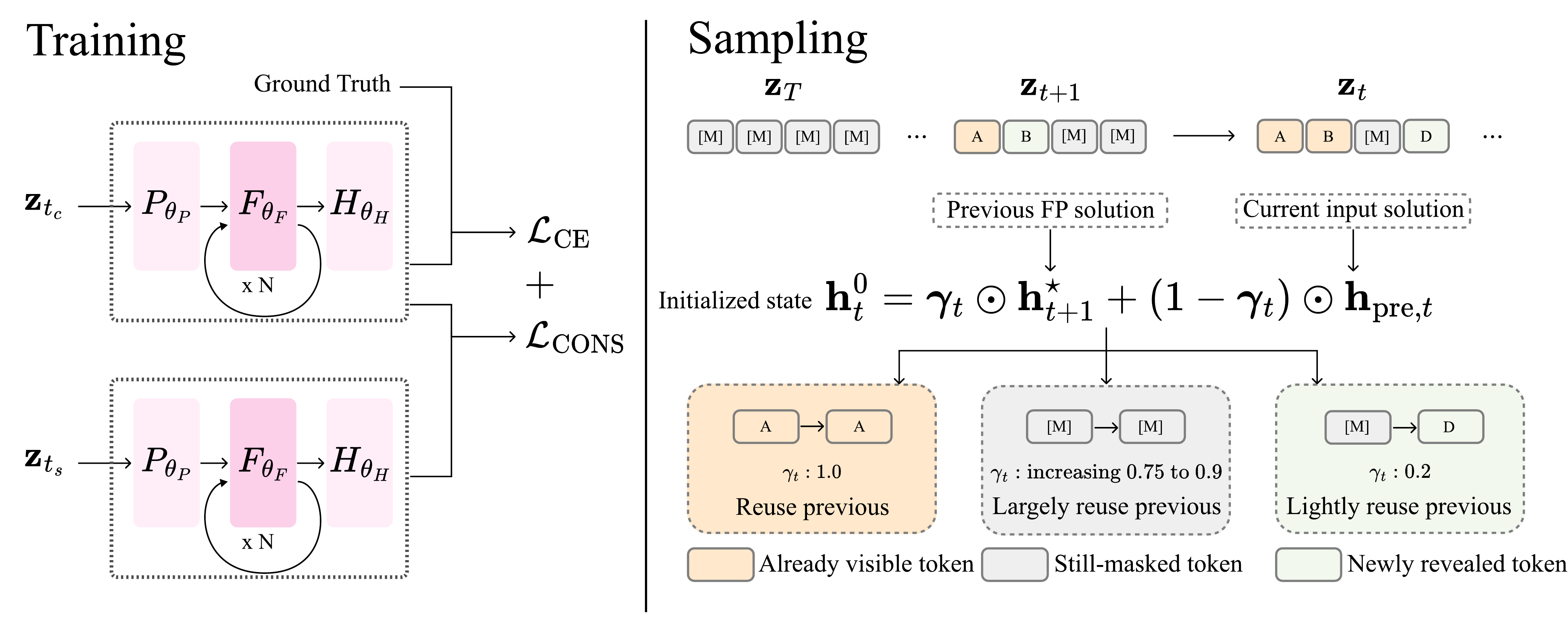}
    \caption{\textbf{Training and sampling for fixed-point masked generative models.}
(Left) During training, FP-MGMs keep the masked modeling objective while replacing the middle transformer stack with an iterated shared fixed-point block. For \textit{cross-step consistency}, correlated masks from the same clean sequence define a noisier student state and cleaner teacher state ($t_c<t_s$); the model is trained with the base cross-entropy loss plus $\mathcal{L}_{\mathrm{CONS}}$ to align their hidden representations.
(Right) During sampling, the fixed-point solver is warm-started from the previous denoising step using \textit{three-state reuse}: visible tokens reuse fully, still-masked tokens partially reuse, and newly revealed tokens rely more on the current pre-layer representation.}
    \label{fig:adaptation}

\end{figure}

\subsection{Fixed-point MGMs}
\label{sub:fpmgm}

A standard masked generative denoiser maps $(\mathbf{z}_t,t)$ to token logits through a finite stack of transformer layers. 
We instead decompose the denoiser into four parts: an explicit preprocessing stack $P_{\theta_P}$, an input-conditioning projection $G_{\theta_G}$, an implicit fixed-point block $F_{\theta_F}$, and an explicit postprocessing stack $H_{\theta_H}$. 
These respectively produce the initial hidden state, transform it into a conditioning signal for the fixed-point layers, solve for the denoising representation, and map this representation to logits:
\vspace{-0.5em}
\begin{equation}
\mathbf{h}_{\mathrm{pre},t}
= P_{\theta_P}(\mathbf{z}_t,t),
\;
\tilde{\mathbf{h}}_t
= G_{\theta_G}(\mathbf{h}_{\mathrm{pre},t}),
\;
\mathbf{h}_t^\star
= \operatorname{Fix}\!\left(F_{\theta_F}(\cdot;\tilde{\mathbf{h}}_t,t)\right),
\;
\boldsymbol{\ell}_\theta(\mathbf{z}_t,t)
= H_{\theta_H}(\mathbf{h}_t^\star,t).
\end{equation}
where $\theta=\{\theta_P,\theta_G,\theta_F,\theta_H\}$ and
$\boldsymbol{\ell}_\theta(\mathbf{z}_t,t)$ are the output token logits.
If no separate projection is used, $G_{\theta_G}$ is the identity.
In practice, the fixed point is approximated by $N$ iterations:
\begin{equation}
    \mathbf{h}_t^{0}=\mathbf{h}_{\mathrm{pre},t},
    \qquad
    \mathbf{h}_t^{n+1}
    =
    F_{\theta_F}(\mathbf{h}_t^n;\tilde{\mathbf{h}}_t,t),
    \qquad n=0,\dots,N-1.
\end{equation}
Let $K_{\mathrm{pre}}$, $K_{\mathrm{fp}}$, and $K_{\mathrm{post}}$ denote the number of transformer blocks in the preprocessing stack, the fixed-point block, and the postprocessing stack, respectively. A refinement step then uses 
$K_{\mathrm{pre}} + N K_{\mathrm{fp}} + K_{\mathrm{post}}$ transformer-block evaluations, while only parameterizing 
$K_{\mathrm{pre}} + K_{\mathrm{fp}} + K_{\mathrm{post}}$ distinct layers. 
Thus, weight sharing reduces parameter count, while the number of solver iterations controls the effective denoiser depth. 
The original MGM objective and sampling rule are unchanged; only the architecture used to compute the logits is modified.

\paragraph{Model variants}
We apply FP-MGMs to two masked generative models with different transformer denoisers: MDLM, which uses a diffusion transformer, and MaskGIT, which uses a bidirectional masked-token transformer. We detail ablations on architecture parameters in Appendices~\ref{appendix:experimental_details} and \ref{appendix:additional_results}.

\subsection{Training a FP-MGM}

Training uses the same task objective as the original masked generative model; our main modification is architectural. 

Following \citet{bai_fixed_2024}, we train with Stochastic Jacobian-Free Backpropagation (SJFB), which avoids backpropagating through the full solver trajectory. 
At each training step, we sample $N_{\mathrm{ng}}\sim\mathcal{U}\{0,\dots,4\}$ no-gradient iterations and $N_{\mathrm{g}}\sim\mathcal{U}\{3,\dots,6\}$ gradient-tracked iterations, where $\mathcal{U}$ denotes a discrete uniform distribution, and with $N=N_{\mathrm{ng}}+N_{\mathrm{g}}$, with $N$ defined in Section~\ref{sub:fpmgm}. 
The no-gradient iterations move the hidden state closer to the fixed point solution without storing activations, while the gradient-tracked iterations provide a tractable training signal for the fixed-point block.
The resulting logits are passed to the original MGM loss. 
Full hyperparameter details are given in Appendix~\ref{appendix:tuning_protocol_details} and more details about SJFB in Appendix~\ref{appendix:sjfb}.

\paragraph{Cross-step consistency regularization}
\label{method:consistency}

The fixed-point architecture improves efficiency by reducing the cost of each denoising step, while strong low-budget generation requires more than cheaper updates. First, each update must predict the clean data accurately, since errors can quickly accumulate when only a few denoising steps are used. Second, fixed-point solutions should be reusable across adjacent denoising states, so the model does not need to resolve each state from scratch.
Our lagged logit analysis highlights the first challenge: low-budget denoising can exhibit substantial cross-step logit drift, motivating an additional consistency signal to align student-step predictions with cleaner future states, improving prediction accuracy with noisier data (Figure~\ref{fig:lagged_consistency}). Moreover, in masked sequence generation, tokens may be revealed between steps, so the previous fixed-point solution is a useful but imperfect warm start for the next state, further motivating stabilizing predictions across adjacent steps.
Specifically, we add a short post-training stage, aligning the representation of a noisier student state with that of a cleaner teacher state from the same trajectory, using correlated masks, i.e., nested masks where the teacher context is always at least as clean as the student's. 
For a clean sequence $\mathbf{x}$, we construct a noisier student input $(\mathbf{z}_{t_s},t_s)$ and a cleaner teacher input $(\mathbf{z}_{t_c},t_c)$ from the same underlying example, with $t_c<t_s$. We then add a consistency term to the base MDLM objective,
$
    \mathcal{L}
    =
    \mathcal{L}_{\mathrm{MDLM}}
    +
    \lambda \mathcal{L}_{\mathrm{CONS}}.
$
In our main experiments, the consistency term is an MSE loss on hidden states,
$
    \mathcal{L}_{\mathrm{CONS}}
    =
    \left\|
    \mathbf{h}_{s} - \mathrm{sg}(\mathbf{h}_{c})
    \right\|_2^2,
$
where $\mathbf{h}_s$ and $\mathbf{h}_c$ are the student and teacher final tokenwise pre-logit hidden states, after the FP and postprocessing blocks, and $\mathrm{sg}(\cdot)$ denotes stop-gradient.

Although this loss is applied in representation space, it also helps improve the model output: empirically, it behaves like cross-time self-distillation, sharpening masked-token predictions and improving low-budget prediction quality. More details about the loss choice, training criterion, and correlated masks are in Appendices~\ref{appendix:lcons_training_dynamics}, \ref{appendix:loss-type-consistency}, and \ref{appendix:correlated-masks}.

\subsection{Sampling with three-state reuse}

At inference time, sampling follows the standard masked denoising process over a fixed number of denoising steps. 
Since each step requires solving a fixed-point problem, the solver initialization directly affects how much useful denoising can be obtained under a limited forward-pass budget. Following the allocation ablation in Appendix~\ref{appendix:budget_allocation_strategies}, we use a decreasing fixed-point iteration schedule, which allocates more solver steps early in the denoising trajectory.

At denoising step $t$, the fixed-point block is conditioned on the current input injection $\tilde{\mathbf{h}}_t=G_{\theta_G}(\mathbf{h}_{\mathrm{pre},t})$, where $\mathbf{h}_{\mathrm{pre},t}=P_{\theta_P}(\mathbf{z}_t,t)$.
Without reuse, the solver is initialized from the current pre-layer output, $\mathbf{h}_t^0=\mathbf{h}_{\mathrm{pre},t}$.
A natural reuse strategy is to warm-start the solver from the fixed-point solution of the previous denoising step, $\mathbf{h}_t^0=\mathbf{h}_{t+1}^\star$.
The fixed-point problem is unchanged; only the solver initialization differs.

\begin{wrapfigure}[18]{r}{0.48\textwidth}
    \centering
    \vspace{-8pt}
    \includegraphics[width=0.47\textwidth]{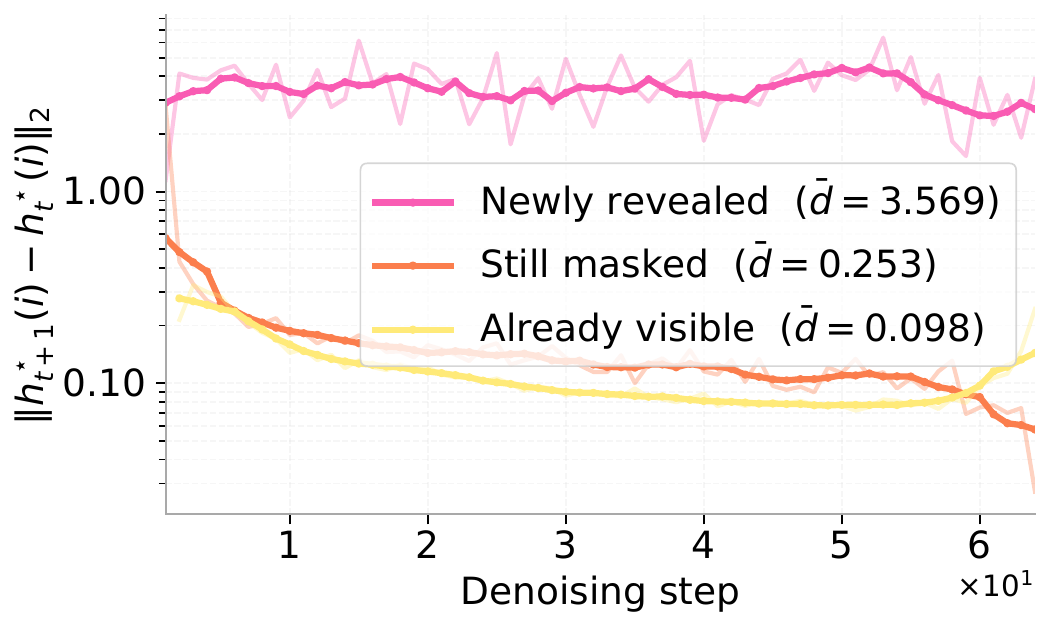}
    \caption{\textbf{Token transition type determines how reusable fixed-point states are.}
Newly revealed tokens move much more than stable tokens, motivating strong reuse for visible tokens, partial reuse for masked tokens, and weak reuse for newly revealed tokens.}
\label{fig:no_reuse_hstar_distance_by_transition}
    \vspace{-10pt}
\end{wrapfigure}

However, full reuse applies the same initialization rule to all positions, implicitly assuming that the previous fixed-point solution remains equally well aligned with the current fixed-point problem. 
In masked denoising, this assumption is violated in a token-dependent way: unchanged visible tokens preserve their local evidence, still-masked tokens keep the same local mask symbol but receive an updated context, and newly revealed tokens undergo a local conditioning shift. 
Thus, the initialization error induced by reuse is not uniform across positions. For each denoising transition under no reuse, we measure the tokenwise movement of the solved fixed-point state,
$\|\mathbf{h}_{t+1}^\star(i)-\mathbf{h}_{t}^\star(i)\|_2$, and group positions by their discrete transition type (Figure~\ref{fig:no_reuse_hstar_distance_by_transition}). 
Newly revealed tokens undergo the largest representation shifts, while already visible tokens move the least. Still-masked tokens' movement significantly decreases as denoising progresses, reflecting that their conditional context changes less once more tokens are visible; this supports stronger reuse at lower noise levels.

Inspired by this, we introduce a \emph{three-state reuse} rule,
$
    \mathbf{h}_t^0
    =
    \boldsymbol{\gamma}_t \odot \mathbf{h}_{t+1}^\star
    +
    (1-\boldsymbol{\gamma}_t)\odot \mathbf{h}_{\mathrm{pre},t}
$
where the token-wise modulation coefficient $\gamma_t^i$ is broadcast over hidden dimensions and is defined as:
$$
\gamma_t^i =
\begin{cases}
1, 
& \text{if position } i \text{ is an unchanged visible token}, \\[2pt]
\gamma_{\mathrm{mask}}, 
& \text{if position } i \text{ is still masked}, \\[2pt]
0.2, 
& \text{if position } i \text{ is newly revealed}.
\end{cases}
$$
Thus, stable visible positions inherit the previous fixed-point solution, still-masked positions use partial reuse, and changed positions move closer to the current pre-layer output. For still-masked tokens, we increase the partial-reuse coefficient as denoising progresses, linearly interpolating from $\gamma_{\mathrm{mask,min}}$ to $\gamma_{\mathrm{mask,max}}$. We select the coefficients of $\boldsymbol{\gamma}t$, including $\gamma{\mathrm{mask,min}}$ and $\gamma_{\mathrm{mask,max}}$, by grid search analyzing the generation quality and diversity metrics. The full coefficient schedule, tuning procedure, and sampling algorithm are given in Appendix~\ref{appendix:tuning_protocol_other_hp} and Algorithm~\ref{alg:three_state_reuse}.

\subsection{Pretrained model conversion and short adaptation}
\label{sec:conversion-adaptation}

Finally, we show that FP-MDLMs need not be trained from scratch -- given a pretrained MDLM, we are able to convert it into a more parameter-efficient FP-MDLM by mapping selected transformer layers to the preprocessing, fixed-point, and postprocessing blocks, with FP-specific projections initialized close to identity.

We then use a short teacher-student adaptation stage, similar in spirit to network distillation, to transfer the behavior of the original MDLM into the converted FP architecture. The original MDLM is kept frozen as the teacher. Using correlated masks from the same clean sequence, the converted FP-MDLM is trained with the base MDLM cross-entropy loss plus a temperature-scaled KL loss on student-masked positions $\mathcal{M}_s$:
$
    \mathcal{L}
    =
    \mathcal{L}_{\mathrm{base}}
    +
    \lambda \tau^2
    \frac{1}{|\mathcal{M}_s|}
    \sum_{i\in\mathcal{M}_s}
    \mathrm{KL}\!\left(
    p_c^\tau(i)\,\|\,p_s^\tau(i)
    \right).
$
This validates that transformer denoisers can be distilled into fixed-point denoisers with only short adaptation; details are given in Appendix~\ref{appendix:conversion_details}.

\section{Experiments}
\label{sec:experiments}

We organize the experiments around three questions. First, does CoFRe improve the end-to-end quality--cost trade-off in language and image generation (Section~\ref{sub:fp-mgm-quality-cost-tradeoff})? Second, can a pretrained MDLM be converted into an effective FP-MDLM with a short adaptation stage? Third, which components are responsible for the gains? We answer the first question with the main OWT and ImageNette results in Table~\ref{tab:main_results_language_image}, the second with a 40k-step checkpoint-adaptation experiment (Section~\ref{sub:model_adaptation}), and the third through ablations on three-state reuse, the consistency objective, and the adaptation initialization (Section~\ref{sub:ablations}). Additional base-model comparisons and extended sweeps are reported in Appendix~\ref{appendix:additional_results}.

\subsection{CoFRe improves the quality-cost trade-off}
\label{sub:fp-mgm-quality-cost-tradeoff}

\paragraph{Experimental setup.}
For language modeling, we evaluate on OWT \citep{Gokaslan2019OpenWeb} with context length 1024, sentence packing, and the GPT-2 tokenizer. We follow the MDLM training setup of \citet{sahoo_simple_2024}; CoFRe uses the same data, tokenizer, and objective, but replaces the middle transformer stack with a fixed-point block. We report generative perplexity (Gen. PPL, via GPT-2 Large) as a measure of quality, and unigram entropy as a measure of diversity and uncertainty \citep{sahoo_simple_2024}, across fixed transformer-block budgets; see Appendix~\ref{appendix:metrics} for details. Following the allocation ablation in Appendix~\ref{appendix:budget_allocation_strategies}, we use a decreasing fixed-point iteration schedule.

For image generation, we evaluate on ImageNette \citep{howard2019imagenette, ImageNet} at 256×256 resolution. Following \citet{besnierHaltonSchedulerMasked2025}, images are tokenized into 16×16=256 latent tokens using the ImageFolder VQ-4096/XQGAN-4096 tokenizer~\citep{li2024imagefolder,li2024xq}. We compare MaskGIT-Large and CoFRe under the same setup, evaluating FID (realism and alignment with the data distribution)~\citep{heusel2018ganstrainedtimescaleupdate} and IS (quality and diversity)~\citep{salimans2016improvedtechniquestraininggans}, as well as latency, training time, and VRAM; further details are in Appendix~\ref{appendix:metrics}.

\begin{table*}[t]
\centering
\scriptsize
\setlength{\tabcolsep}{3.2pt}
\renewcommand{\arraystretch}{1.08}
\sisetup{detect-weight=true, detect-inline-weight=math}

\resizebox{\linewidth}{!}{%
\begin{tabular}{@{}
r
S[table-format=3.3]
S[table-format=1.3]
S[table-format=3.3]
S[table-format=1.3]
@{\hspace{8pt}\vrule width 0.8pt\hspace{8pt}}
r
S[table-format=3.4]
S[table-format=2.4]
S[table-format=3.4]
S[table-format=2.4]
@{}}
\toprule
\multicolumn{5}{c}{\textbf{Language generation on OWT}} &
\multicolumn{5}{c}{\textbf{Image generation on ImageNette}} \\
\cmidrule(lr){1-5}\cmidrule(lr){6-10}

\textbf{Budget}
& \multicolumn{2}{c}{\shortstack{\textbf{MDLM + SDTT}\\ \scriptsize Train: $\approx$139h + SDTT\\ \scriptsize VRAM: 112.4 GiB/GPU}}
& \multicolumn{2}{c}{\shortstack{\textbf{CoFRe}\\ \scriptsize Train: $\approx$123h + 30k\\ \scriptsize VRAM: 93.44 GiB/GPU}}
&
\textbf{Budget}
& \multicolumn{2}{c}{\shortstack{\textbf{MaskGIT-Large}\\ \scriptsize Train: 17h46m\\ \scriptsize VRAM: 72.45 GiB}}
& \multicolumn{2}{c}{\shortstack{\textbf{CoFRe}\\ \scriptsize Train: 9h08m\\ \scriptsize VRAM: 35.74 GiB}} \\
\cmidrule(lr){2-3}\cmidrule(lr){4-5}\cmidrule(lr){7-8}\cmidrule(lr){9-10}
& {\textbf{Gen. PPL} $\downarrow$}
& {\textbf{Entropy} $\uparrow$}
& {\textbf{Gen. PPL} $\downarrow$}
& {\textbf{Entropy} $\uparrow$}
&
& {\textbf{FID} $\downarrow$}
& {\textbf{IS} $\uparrow$}
& {\textbf{FID} $\downarrow$}
& {\textbf{IS} $\uparrow$} \\
\midrule

96
& 193.050 & \best{5.580}
& \best{101.791} & 5.434
&
48
& 174.0856 & 9.2860
& \best{96.7331} & \best{14.4074} \\

192
& 89.170 & \best{5.530}
& \best{65.182} & 5.380
&
96
& 117.6439 & 13.3696
& \best{51.0077} & \best{15.9572} \\

384
& 62.290 & \best{5.490}
& \best{48.755} & 5.283
&
192
& 54.6172 & \best{16.0220}
& \best{27.6242} & 15.0822 \\

768
& 47.040 & \best{5.450}
& \best{37.846} & 5.142
&
384
& 30.0202 & \best{14.6473}
& \best{22.8381} & 14.4567 \\

\bottomrule
\end{tabular}%
}
\caption{\textbf{Main quality--cost results for language (Left) and image generation (Right).}
Budgets count transformer-block forward passes. For language, the training/VRAM values indicate the main backbone cost; SDTT uses additional short distillation stages. Entropy is reported to contextualize diversity; these values correspond to the selected operating point for each budget, while Appendix~\ref{appendix:tradeoff-steps-iter} provides the broader quality--diversity landscape obtained by varying the allocation between denoising steps and fixed-point iterations.
}
\vspace{-1em}
\label{tab:main_results_language_image}
\end{table*}

\paragraph{Results.}

Table~\ref{tab:main_results_language_image} reports the main quality--cost comparison for both modalities. We emphasize that the reported CoFRe numbers correspond to one point on a broader quality--diversity trade-off: Appendix~\ref{appendix:tradeoff-steps-iter} sweeps the allocation between denoising steps and fixed-point iterations, showing that the strongest CoFRe configurations improve generative perplexity without relying on a degenerate entropy regime. For language, CoFRe improves generative perplexity over MDLM+SDTT at every reported budget, reducing from 193.1 to 101.8 at budget 96 and from 47.0 to 37.8 at budget 768. Beyond these gains, CoFRe also reduces backbone cost: CoFRe uses 93.44 GiB/GPU and approximately 123h of training, compared to 112.4 GiB/GPU and approximately 139h for MDLM before the additional SDTT stage (Figure~\ref{fig:consistency}).

The same pattern holds for image generation. CoFRe improves FID over MaskGIT-Large at every reported budget, for example from 174.1 to 96.7 at budget 48 and from 30.0 to 22.8 at budget 384. It also reduces training time from 17h46m to 9h08m and VRAM from 72.45\,GiB to 35.74\,GiB. Overall, Table~\ref{tab:main_results_language_image} shows that fixed-point denoisers improve the quality--cost trade-off across modalities, and that CoFRe turns this efficiency gain into stronger low-budget generation. Further work and additional experiments in Appendix~\ref{appendix:additional_results}.

\subsection{
Adapting a pretrained MDLM checkpoint into FP-MDLM}
\label{sub:model_adaptation}

\paragraph{Experimental setup}

We use a pretrained MDLM teacher from \citet{sahoo_simple_2024} trained on OWT with sequence length 1024. The converted FP-MDLM is initialized from this checkpoint as described in Section~\ref{sec:conversion-adaptation}, and then adapted on OWT with the same sequence length. During adaptation, we use the KL consistency loss on correlated masked inputs. Then during post-training, the KL coefficient
is linearly warmed up from 0 to 0.1 over 5k global steps and then kept constant at 0.1 for the remainder of adaptation. We compare the generative perplexity and unigram entropy of base FP-MDLM and adapted FP-MDLM across different budgets.

\begin{wrapfigure}[15]{r}{0.48\textwidth}
    \centering
    \vspace{-8pt}
    \includegraphics[width=0.47\textwidth]{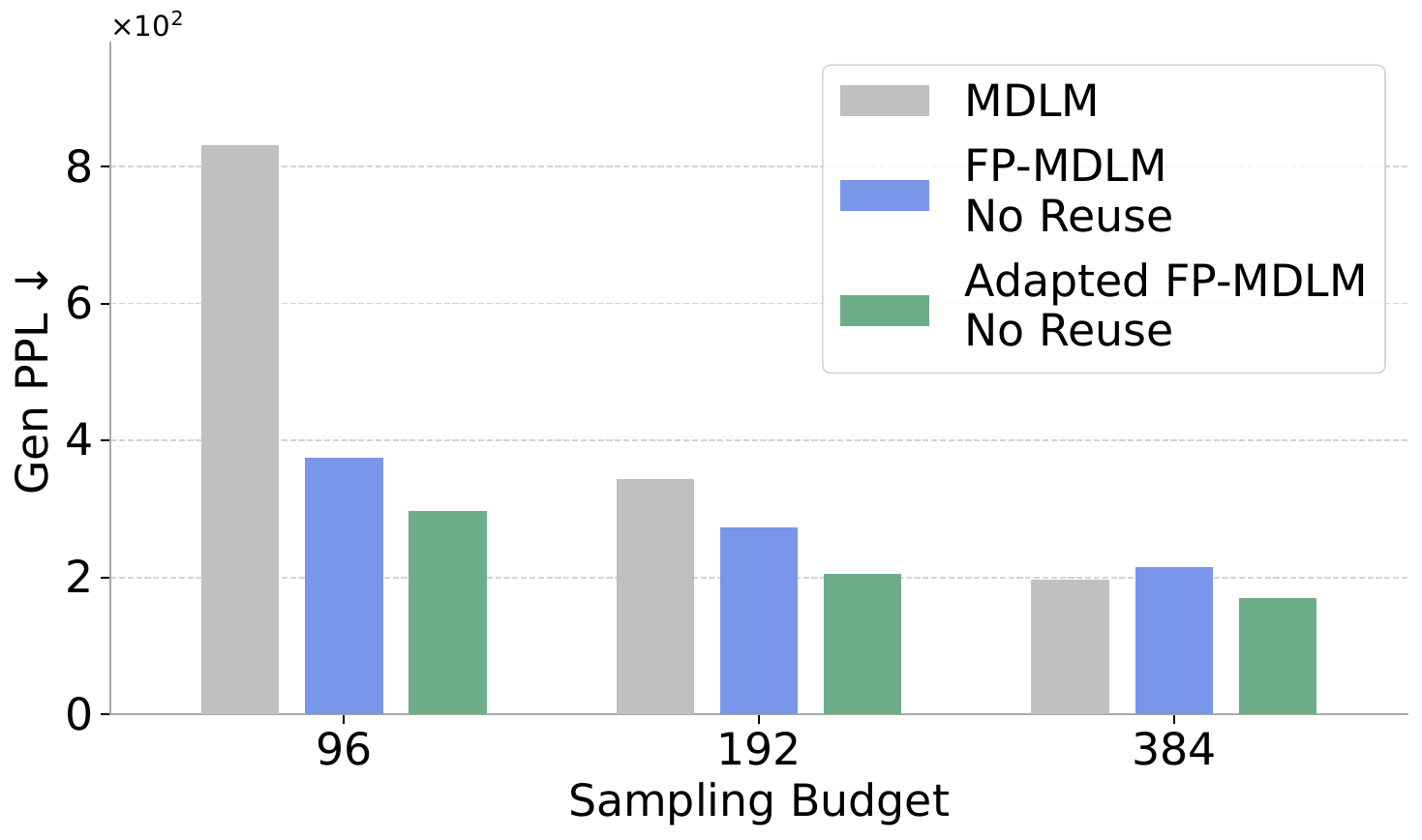}
    \caption{\textbf{Short from-scratch adaptation improves FP-MDLM on OWT.} Adapted FP-MDLM improves generation quality at every budget tested.}
    \label{fig:adapted_fp_mdlm_gen_ppl}
    \vspace{-10pt}
\end{wrapfigure}

\paragraph{Results}
Figure~\ref{fig:adapted_fp_mdlm_gen_ppl} and Table~\ref{tab:fp_mdlm_component_ablation} shows that a pretrained MDLM checkpoint can be converted into a stronger FP-MDLM with only 40k adaptation steps. We first compare the models without reuse in order to isolate the effect of the adaptation on the vanilla fixed-point model itself. In this setting, the adapted checkpoint improves generative perplexity at every budget, from 375.6 to 296.8 at budget 96 and from 179.7 to 149.6 at budget 768, while keeping entropy close to the baseline.
The adaptation also improves the behaviour of reuse. For the baseline FP-MDLM, reuse is inconsistent and can hurt generation quality at larger budgets. After adaptation, however, reuse becomes beneficial in the medium- and high-budget regimes: both full reuse and three-state reuse improve over no reuse at budgets 192, 384, and 768. The strongest results are obtained with three-state reuse, which reaches generative perplexities of 192.2, 149.4, and 131.3 at these budgets. Overall, these results show that pretrained MDLM checkpoints can be turned into effective FP-MDLM generators with a short adaptation stage, and that this adaptation restores the benefit of reuse when the sampling budget is sufficiently large. More details and results in Table~\ref{tab:fp_mdlm_component_ablation}.

\subsection{Ablations}
\label{sub:ablations}

\begin{wrapfigure}[13]{r}{0.47\textwidth}
    \centering
    \vspace{5pt}
    \includegraphics[width=0.47\textwidth]{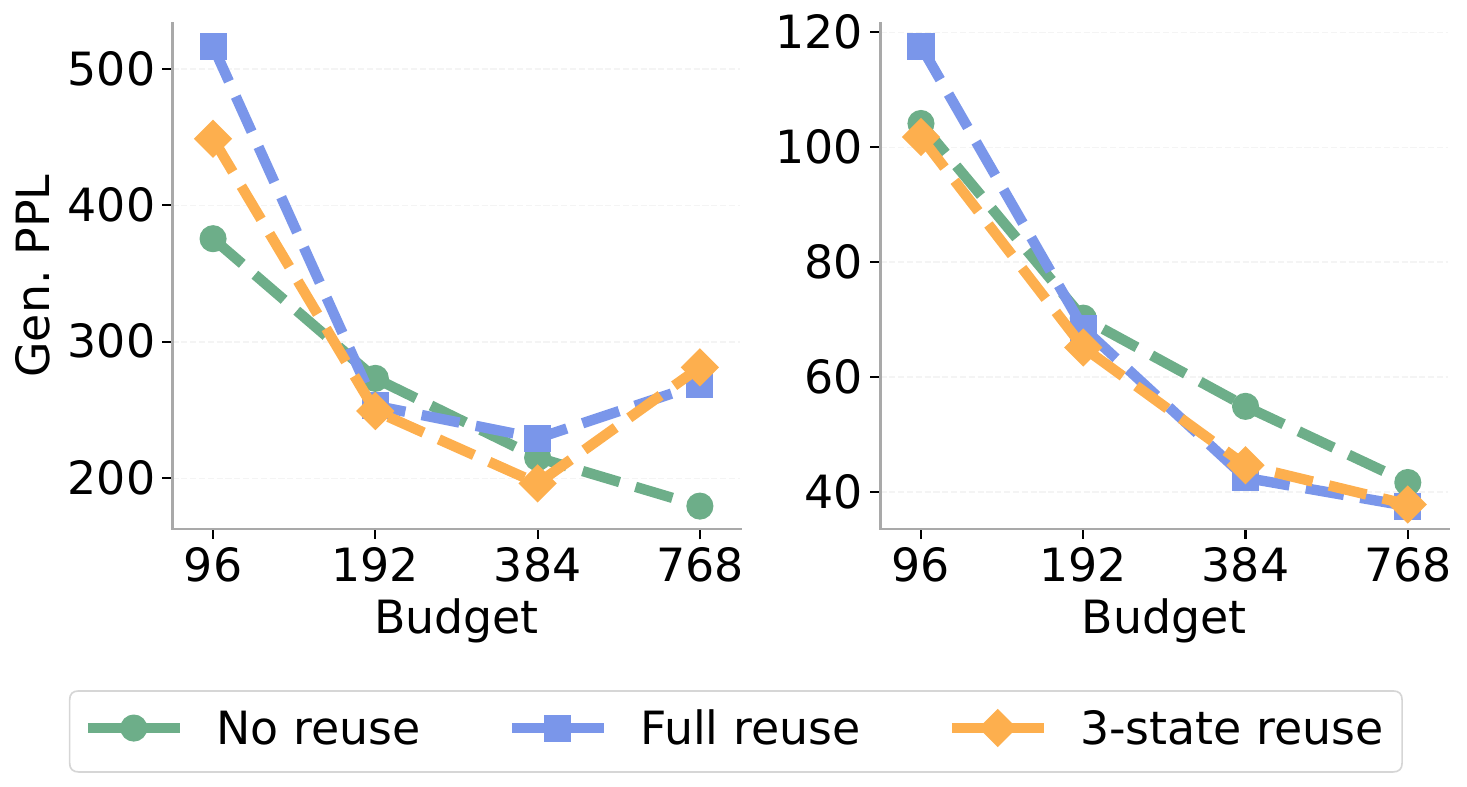}
    \caption{\textbf{Effect of different warm-start of the fixed-point on FP-MDLM base (Left) and FP-MDLM+$\mathcal{L}_{\mathrm{CONS}}$ (Right).}
}
    \label{fig:three_state}
\end{wrapfigure}

We isolate two design choices that are not covered by the main end-to-end results: two main ingredients of CoFRe: three-state reuse and consistency loss, and the pretrained layer initialization
used during checkpoint adaptation. Each ablation changes only the component under study while keeping the training or sampling protocol fixed.

\subsubsection{Three-state reuse}
\label{sub:solution-reuse}

\paragraph{Experimental setup}
We evaluate inference-time solution reuse by keeping the FP-MDLM checkpoint and the rest of the sampling setup fixed, and varying only the initialization of the fixed-point solver. We compare three settings: no reuse, full reuse, and three-state reuse, and report GPT-2 Large generative perplexity together with sample entropy. We also analyze how the initialization to solved distance compares across the three regimes.

\paragraph{Results}
Figure~\ref{fig:three_state} separates inference-time reuse from consistency post-training. On the base FP-MDLM (left), 3SR improves over full reuse mainly at low and medium budgets, with the gap reducing at larger budgets. Figure~\ref{fig:no_reuse_hstar_distance_by_transition} explains this behavior: both reuse variants reduce the initialization-to-solution distance, but full reuse treats all tokens uniformly, including newly revealed tokens whose fixed-point states change the most. In contrast, 3SR weakens reuse for newly revealed tokens, partially reuses still-masked tokens, and strongly reuses stable visible tokens.

With $\mathcal{L}_{\mathrm{CONS}}$ (right), the same trend holds, but all methods achieve much lower generative perplexity. Thus, $\mathcal{L}_{\mathrm{CONS}}$ and 3SR are complementary: consistency improves prediction quality, while 3SR provides better token-aware solver initialization. Additional results in Appendices~\ref{appendix:owt_full_results}, \ref{appendix:budget_allocation_strategies}, and~\ref{appendix:3sr}.

\subsubsection{Adaptation initialization.}

\paragraph{Experimental setup}
We compare two 40k-step FP-MDLM adaptation runs on OWT with sequence length 1024. The \emph{Initialized} model is initialized from a pretrained MDLM checkpoint using the layer-mapping procedure described in Section~\ref{sec:conversion-adaptation}, while the \emph{Not initialized} model uses the same FP-MDLM architecture and adaptation objective but does not use this pretrained layer initialization. Both models are adapted with the same frozen teacher, correlated-mask KL objective, optimizer and sampling settings.

\begin{wraptable}[10]{r}{0.5\textwidth}
\centering
\vspace{-0.25cm}
\scriptsize
\caption{\textbf{Initializing from a pretrained MDLM improves FP-MDLM adaptation with 3SR.}}
\label{tab:gen_ppl_3sr_noinit_init}
\setlength{\tabcolsep}{2.2pt}
\renewcommand{\arraystretch}{0.55}
\sisetup{detect-weight=true, detect-inline-weight=math}
\begin{tabular*}{\linewidth}{@{\extracolsep{\fill}} ll c c c c @{}}
\toprule
& & \multicolumn{4}{c}{\textbf{Budget}} \\
\cmidrule(l){3-6}
\textbf{Model} & \textbf{Metric} & \textbf{96} & \textbf{192} & \textbf{384} & \textbf{768}\\
\midrule
\multirow{2}{*}{\shortstack[l]{\textbf{No init}}}
    & Gen PPL $\downarrow$ & 298.643 & 192.227 & 149.427 & 131.291 \\
    & Entropy $\uparrow$ & \best{5.722} & \best{5.658} & 5.611 & \best{5.597} \\
\cmidrule(lr){1-6}
\multirow{2}{*}{\shortstack[l]{\textbf{Init}}}
    & Gen PPL $\downarrow$ & \best{286.403} & \best{184.708} & \best{147.497} & \best{126.872} \\
    & Entropy $\uparrow$ & 5.695 & 5.649 & \best{5.619} & 5.572 \\
\bottomrule
\end{tabular*}
\end{wraptable}

\paragraph{Results}
Tables~\ref{tab:gen_ppl_3sr_noinit_init} and \ref{tab:gen_ppl_noinit_scratch} show that pretrained initialization improves short adaptation. Without reuse, generative perplexity drops from 296.8 to 276.0 at budget 96 and from 149.6 to 139.0 at budget 768. With 3SR, initialization is consistently better from budget 192 onward, reaching 126.9 at budget 768. The initialized run also trains faster (Figure~\ref{fig:init_vs_noinit_loss}), showing that the layer mapping provides a better starting point.

\section{Related work}


\paragraph{Efficient training of MGMs}
PUMA \citep{kimStopTrainingWorst2026} aligns training with inference-time unmasking patterns. DiffuGPT/DiffuLLaMA \citep{gongScalingDiffusionLanguage2025b} and Dream \citep{yeDream7BDiffusion2025} adapt pretrained autoregressive models into bidirectional diffusion models, rather than training from scratch. We reduce training cost via a weight-shared fixed-point solver, and show that pre-trained MGMs can be adapted into FP-MGMs.

\paragraph{Efficient few-step generation with MGMs}
Unmasking schedules and token-order policies select which positions to update at each step \citep{besnierHaltonSchedulerMasked2025, jazbecLearningUnmaskingPolicies2026, hongImprovingDiscreteDiffusion2026, pengPathPlanningMasked2025a, kimTrainWorstPlan2025, liuThinkWhileYou2024}. Discrete solvers and timestep schedules reduce the number of sampling steps \citep{luxembourgPlanSpeedDilated2025, renFastSolversDiscrete2025, forestiImprovedSamplingSchedules2026a, parkJumpYourSteps2024}. Distillation compresses many-step teachers into few-step students \citep{deschenaux2024beyond, zhuDi$mathttM$ODistillingMasked2025, hoogeboomBeyondSingleTokens2026, liIDLMInverseDistilled2026, sahooDiffusionDuality2025}. Self-speculative decoding produces non-factorized predictions over masked positions in a single forward pass by drafting and validating tokens \citep{campbellSelfSpeculativeMaskedDiffusions2026}. PGM \citep{deschenauxPartitionGenerativeModeling2026} removes explicit mask tokens to improve the throughput during sampling. CDLM \citep{kimCDLMConsistency2025} reduces the number of sampling steps via a consistency objective and uses block-wise causal attention to enable KV caching. We instead keep bidirectional attention, reducing the per-step cost via a weight-shared fixed-point solver, and add cross-step consistency and three-state reuse.

\paragraph{Implicit depth and looped models}
Universal \citep{dehghaniUniversalTransformers2019} and looped Transformers \citep{Prairie2026ParcaeSL, yuSpiralFormerLoopedTransformers2026, jeddiLoopFormer2026} repeatedly apply shared blocks for adaptive depth. DEQ \citep{baiDeepEquilibriumModels2019} solve for the fixed point of a shared layer via implicit differentiation, and the Generative Equilibrium Transformer \citep{gengOneStepDiffusionDistillation2023} uses a DEQ for one-step diffusion distillation. In autoregressive LLMs, Mixture-of-Recursions and Relaxed Recursive Transformers~\citep{baeMixtureRecursions2025, baeRelaxedRecursiveTransformers2024} use different depths for different tokens. In continuous diffusion, Fixed-Point Diffusion Models \citep{bai_fixed_2024} combine an implicit solver with state reuse. The implicit-depth methods above apply to continuous diffusion or autoregressive language models. MGMs differ because the local conditioning of each position changes non-uniformly across denoising steps in one step, in an arbitrary order. We add two mechanisms to the fixed-point solver. (1) The three-state reuse rule handles clean, masked, and newly decoded tokens differently. (2) The cross-step consistency behaves like self-distillation and substantially improves in low-budget generative perplexity.

\section{Conclusion}
\label{sec:conclusion}

We introduce Fixed-Point Masked Generative Models (FP-MGMs), which replace part of the denoising transformer with an implicit weight-sharing block. When applied to MDLM and MaskGIT, FP-MGMs reduce parameters, training time, and memory while improving performance on low-budget generation.The fixed-point architecture provides the efficiency gains, but effective low-budget generation also requires stabilizing training and reuse across denoising steps. Cross-step consistency drives low-budget generation quality, while three-state reuse enables token-aware warm starts; together, they make CoFRe a complete training-to-inference recipe. We also show that pretrained MDLM checkpoints can be converted into a fixed-point model with only short adaptation. Overall, CoFRe offers a practical path toward cheaper and more flexible masked generative models.

\begin{ack}
Yiming Qin and Alba Carballo-Castro were supported by the Swiss National Science Foundation (SNSF grant 10001445). Justin Deschenaux has received funding from the Swiss State Secretariat for Education, Research and Innovation (SERI).
\end{ack}

\bibliographystyle{plainnat} 
\bibliography{references}


\appendix
\clearpage
\tableofcontents
\clearpage

\section{Limitations and future work}
\label{appendix:limitations}

While this work advances masked generative modeling by introducing fixed-point denoisers, cross-step consistency regularization, and three-state reuse, several limitations remain. We discuss these limitations below, both to clarify the scope of our current results and to highlight directions for future research.

\paragraph{Scale and scope.}
Our experiments are limited to OWT-scale language modeling and ImageNette image generation. These settings allow controlled comparisons, but do not yet show whether FP-MGMs scale to larger language models, larger image datasets, or multimodal generation. Evaluating FP-MGMs at larger model and data scales is an important direction for future work.

\paragraph{Additional tuning and adaptation.}
FP-MGMs introduce extra design choices, including where to place the fixed-point block, how many solver iterations to use, how to set reuse coefficients, and how to allocate the sampling budget across denoising steps. We ablate these choices, but the current recipe remains partly heuristic. In addition, the best FP-MDLM results require a short consistency post-training stage, whose duration must be chosen carefully to avoid over-sharpening and entropy collapse.

\paragraph{Generality and practical speedups.}
Three-state reuse is designed for monotonic masked decoding, where tokens are revealed and then remain fixed; samplers that remask or revise visible tokens may require different reuse rules. Moreover, our main compute metric is transformer-block forward passes, which is hardware-independent but does not always translate directly into wall-clock gains because fixed-point solvers add control-flow overhead. Future work should develop adaptive stopping rules, optimized implementations, and reuse strategies for more general masked-generation trajectories.

\section{Ethics statement}
\label{appendix:ethics_statement}

The objective of this work is to improve the efficiency of masked generative models by introducing fixed-point denoisers, cross-step consistency regularization, and three-state reuse. Masked generative models are relevant to a broad range of applications, including language generation, image synthesis, video, audio, and multimodal modeling. Improvements in their training and sampling efficiency may therefore reduce the computational cost of developing and deploying generative models, making such models more accessible to researchers with limited compute resources.

At the same time, FP-MGMs inherit the broader risks of generative models. More efficient generation can lower the cost of producing synthetic text or images, which may amplify existing concerns around misinformation, spam, copyright misuse, or the generation of biased and harmful content. Our experiments are limited to moderate-scale language and image benchmarks, and the generated samples still exhibit failure modes such as repetition, factual inconsistency, and topic drift. As a result, we do not view the current models as directly suitable for high-stakes applications such as medical, legal, or policy decision-making.

Overall, this work primarily contributes an architectural and algorithmic efficiency improvement. We expect its main near-term impact to be methodological, by providing a route toward cheaper training and stronger low-budget masked generation. Future work should evaluate FP-MGMs at larger scales and study how efficiency gains interact with safety, bias, memorization, and misuse risks in practical deployments.

\section{Additional Method Details}
\label{appendix:method_details}

\subsection{Stochastic Jacobian-Free Backpropagation}
\label{appendix:sjfb}

Training an FP-MGM requires differentiating through the implicit fixed-point block, whose output at denoising state $t$ is defined as the hidden-state solution
\[
\mathbf{h}_t^\star
=
F_{\theta_F}(\mathbf{h}_t^\star;\tilde{\mathbf{h}}_t,t),
\]
where $\tilde{\mathbf{h}}_t=G_{\theta_G}(\mathbf{h}_{\mathrm{pre},t})$ is the input-conditioning signal produced by the preceding explicit layers. In principle, one can backpropagate through this equilibrium using implicit differentiation, which gives a gradient involving the inverse Jacobian term
\[
\left(
I -
\frac{\partial F_{\theta_F}(\mathbf{h}_t^\star;\tilde{\mathbf{h}}_t,t)}
{\partial \mathbf{h}_t^\star}
\right)^{-1}.
\]
However, explicitly forming or solving this Jacobian system is computationally expensive and can be unstable at scale. Jacobian-Free Backpropagation (JFB) \citep{fungJFBJacobianFreeBackpropagation2021} avoids this cost by first computing an approximate fixed point without storing intermediate activations, and then applying one additional fixed-point iteration with gradients enabled; the backward pass is therefore performed only through this final step, giving an approximate gradient of the form:
\[
\frac{\partial \mathcal{L}}{\partial \theta_F}
\approx
\frac{\partial \mathcal{L}}{\partial \mathbf{h}_t^\star}
\frac{\partial F_{\theta_F}(\mathbf{h}_t^\star;\tilde{\mathbf{h}}_t,t)}
{\partial \theta_F}.
\]

Stochastic Jacobian-Free Backpropagation (S-JFB) \citep{bai_fixed_2024} generalizes this idea by unrolling a random number of fixed-point iterations during training. At each training step, S-JFB samples two integers $n \sim \mathcal{U}\{0,\ldots,N\}$ and $m \sim \mathcal{U}\{1,\ldots,M\}$. It first performs $n$ fixed-point iterations under a stop-gradient/no-gradient context, producing an approximate equilibrium while avoiding the memory cost of storing these intermediate states. It then performs $m$ additional iterations with gradient tracking enabled, and the loss is backpropagated only through these last $m$ unrolled iterations. The hyperparameters $N$ and $M$ therefore control the maximum number of fixed-point iterations used without and with gradients, respectively. Compared with standard one-step JFB, S-JFB is slightly more expensive because it backpropagates through multiple final iterations rather than only one, but it remains much cheaper than full implicit differentiation or fully unrolled explicit networks. Its stochasticity also exposes the model to different approximation depths during training, which makes the fixed-point layer more robust and empirically improves optimization compared with the deterministic one-step JFB baseline.

\subsection{Three-State Reuse Details}
\label{appendix:3sr}

Three-state reuse is the inference-time mechanism we propose to warm-start the fixed-point solver across masked denoising steps. Unlike full reuse, which initializes every token from the previous fixed-point solution, 3SR accounts for the fact that token states evolve non-uniformly during sampling: some tokens remain visible and unchanged, some remain masked but receive updated context, and others are newly revealed. This appendix gives the exact token-wise interpolation rule used to initialize the solver, the visible-fraction-dependent coefficient schedule, and the complete sampling procedure.

\paragraph{Three-state reuse schedule.}
We detail here the schedule used by 3SR. We base this schedule on the results obtained when looking at the distance between $\mathbf{h}^{\star}_t$ and $\mathbf{h}^{\star}_{t+1}$, as plotting on Figure~\ref{fig:no_reuse_hstar_distance_by_transition}. We analyze this distance as it shows, for each type of transition, how far the previous fixed-point solution is from the current fixed-point solution, when the initialization is made without reuse. We therefore use the following schedule:
 Unchanged visible tokens use full reuse, with $\gamma_t=1.0$. Still-masked tokens use
$\gamma_t=
\gamma_{\mathrm{masked}} =
\gamma_{\mathrm{mask,min}}
+ \bigl(\gamma_{\mathrm{mask,max}} - \gamma_{\mathrm{mask,min}}\bigr)\,v_t,
$
while newly revealed tokens use
$
\gamma_t=\gamma_{\mathrm{changed}}=0.2$
where $
v_t = \frac{1}{d}\sum_{i=1}^d \mathbbm{1}[z_t^i \neq \mask]
$ is the fraction of visible tokens at step $t$. 
In our base setting, we use $\gamma_{\mathrm{mask,min}}=0.75$, $\gamma_{\mathrm{mask,max}}=0.90$, $\gamma_{\mathrm{changed}}=0.2$. We tune these hyperparameters using a grid search. 

To select the reuse coefficients, we run a sweep over the masked-token and newly revealed-token interpolation parameters. Tables~\ref{tab:fp_mdlm_fixed_gamma_sweep} and~\ref{tab:fp_mdlm_changed_gamma_075_090} report these sweeps across sampling budgets. Table~\ref{tab:fp_mdlm_fixed_gamma_sweep} varies the reuse range for still-masked tokens and the maximum reuse assigned to newly revealed tokens, while Table~\ref{tab:fp_mdlm_changed_gamma_075_090} fixes the masked-token range and varies the newly revealed-token reuse coefficient more finely. Across these sweeps, performance is relatively robust within a moderate range of coefficients, supporting the use of a simple hand-tuned 3SR schedule rather than a learned or highly budget-specific policy.

\FloatBarrier
\begin{algorithm}[H]
\caption{FP-MGM sampling with three-state reuse}
\label{alg:three_state_reuse}
\begin{algorithmic}[1]
\Require Schedule $1=\tau_T>\cdots>\tau_0=0$, initial state $\mathbf{z}_{\tau_T}=\mask^d$
\Require Preprocessing stack $P_{\theta_P}$, input-conditioning projection $G_{\theta_G}$, fixed-point block $F_{\theta_F}$, postprocessing stack $H_{\theta_H}$
\Require Per-step solver iterations $N_i$
\Require Reuse parameters $\gamma_{\mathrm{mask,min}},\gamma_{\mathrm{mask,max}},\gamma_{\mathrm{changed}}$

\State $\mathbf{h}^{\star}_{\mathrm{prev}}\gets \emptyset$
\State $\mathbf{z}_{\mathrm{prev}}\gets \emptyset$

\For{$i=T,T-1,\dots,1$}
    \State $\mathbf{h}_{\mathrm{pre},\tau_i}\gets P_{\theta_P}(\mathbf{z}_{\tau_i},\tau_i)$
    \State $\tilde{\mathbf{h}}_{\tau_i}\gets G_{\theta_G}(\mathbf{h}_{\mathrm{pre},\tau_i})$

    \If{$\mathbf{h}^{\star}_{\mathrm{prev}}=\emptyset$}
        \State $\mathbf{h}^{0}_{\tau_i}\gets \mathbf{h}_{\mathrm{pre},\tau_i}$
    \Else
        \State $v_{\tau_i}\gets \frac{1}{d}\sum_{j=1}^{d}\mathbbm{1}[z_{\tau_i}^j\neq \mask]$
        \State $\gamma_{\mathrm{mask}}\gets
        \gamma_{\mathrm{mask,min}}
        +(\gamma_{\mathrm{mask,max}}-\gamma_{\mathrm{mask,min}})v_{\tau_i}$

        \State Define token-wise reuse coefficients $\gamma_{\tau_i}^j$ as
        \Statex
        \[
        \gamma_{\tau_i}^j =
        \begin{cases}
        1,
        & \text{if } z_{\tau_i}^j=z_{\mathrm{prev}}^j\neq\mask,\\[2pt]
        \gamma_{\mathrm{mask}},
        & \text{if } z_{\tau_i}^j=z_{\mathrm{prev}}^j=\mask,\\[2pt]
        \gamma_{\mathrm{changed}},
        & \text{otherwise}.
        \end{cases}
        \]
        \State $\mathbf{h}^{0}_{\tau_i}\gets
        \boldsymbol{\gamma}_{\tau_i}\odot \mathbf{h}^{\star}_{\mathrm{prev}}
        +(1-\boldsymbol{\gamma}_{\tau_i})\odot \mathbf{h}_{\mathrm{pre},\tau_i}$
    \EndIf

    \For{$n=0,\dots,N_i-1$}
        \State $\mathbf{h}^{n+1}_{\tau_i}\gets
        F_{\theta_F}(\mathbf{h}^{n}_{\tau_i};\tilde{\mathbf{h}}_{\tau_i},\tau_i)$
    \EndFor

    \State $\mathbf{h}^{\star}_{\tau_i}\gets \mathbf{h}^{N_i}_{\tau_i}$
    \State $\boldsymbol{\ell}_{\theta}(\mathbf{z}_{\tau_i},\tau_i)
    \gets H_{\theta_H}(\mathbf{h}^{\star}_{\tau_i},\tau_i)$
    \State Sample $\mathbf{z}_{\tau_{i-1}}$ using the original MGM transition rule and logits $\boldsymbol{\ell}_{\theta}(\mathbf{z}_{\tau_i},\tau_i)$
    \State $\mathbf{h}^{\star}_{\mathrm{prev}}\gets \mathbf{h}^{\star}_{\tau_i}$
    \State $\mathbf{z}_{\mathrm{prev}}\gets \mathbf{z}_{\tau_i}$
\EndFor

\State \Return $\mathbf{z}_{\tau_0}$
\end{algorithmic}
\end{algorithm}
\FloatBarrier

\FloatBarrier
\begin{table}[!t]
\centering
\small
\caption{Results for \texttt{fp\_mdlm} with fixed strategy across budgets. Constant settings for all runs: visible $\gamma = 1.0$, changed $\gamma_{\min}=0.0$. Values are shown to four decimal places.}
\label{tab:fp_mdlm_fixed_gamma_sweep}
\setlength{\tabcolsep}{1.2pt}
\renewcommand{\arraystretch}{0.72}
\sisetup{detect-weight=true, detect-inline-weight=math}
\resizebox{0.83\linewidth}{!}{
\begin{tabular*}{\linewidth}{@{\extracolsep{\fill}} l c l c c c c @{}}
\toprule
& & & \multicolumn{4}{c}{\textbf{Budget}} \\
\cmidrule(l){4-7}
\textbf{Masked $\gamma$ range} & \textbf{Changed $\gamma_{\max}$} & \textbf{Metric} & \textbf{96} & \textbf{192} & \textbf{384} & \textbf{768}\\
\midrule

\multirow{6}{*}{\shortstack[l]{$[0.60, 0.90]$}}
& \multirow{2}{*}{0.1} & Gen PPL $\downarrow$ & 94.1776 & 83.0655 & 47.4160 & 37.0117\\
& & Entropy $\uparrow$ & 5.5104 & 5.4820 & 5.3554 & 5.2280\\
\cmidrule(lr){2-7}
& \multirow{2}{*}{0.2} & Gen PPL $\downarrow$ & 95.3269 & 81.8788 & 47.2272 & 36.9850\\
& & Entropy $\uparrow$ & 5.5110 & 5.4766 & 5.3562 & 5.2260\\
\cmidrule(lr){2-7}
& \multirow{2}{*}{0.3} & Gen PPL $\downarrow$ & 96.1154 & 81.8405 & 46.8567 & 36.9900\\
& & Entropy $\uparrow$ & 5.5181 & 5.4817 & 5.3548 & 5.2394\\
\midrule

\multirow{6}{*}{\shortstack[l]{$[0.60, 0.95]$}}
& \multirow{2}{*}{0.1} & Gen PPL $\downarrow$ & 94.7594 & 82.7298 & 47.1981 & 36.7854\\
& & Entropy $\uparrow$ & 5.5115 & 5.4771 & 5.3496 & 5.2242\\
\cmidrule(lr){2-7}
& \multirow{2}{*}{0.2} & Gen PPL $\downarrow$ & 95.0930 & 81.3275 & 47.4469 & 37.2802\\
& & Entropy $\uparrow$ & 5.5126 & 5.4770 & 5.3526 & 5.2504\\
\cmidrule(lr){2-7}
& \multirow{2}{*}{0.3} & Gen PPL $\downarrow$ & 95.4836 & 83.2949 & 47.0793 & 36.8499\\
& & Entropy $\uparrow$ & 5.5122 & 5.4893 & 5.3624 & 5.2257\\
\midrule

\multirow{6}{*}{\shortstack[l]{$[0.75, 0.90]$}}
& \multirow{2}{*}{0.1} & Gen PPL $\downarrow$ & 97.2015 & 81.5204 & 45.9489 & 36.8647\\
& & Entropy $\uparrow$ & 5.5201 & 5.4752 & 5.3500 & 5.2429\\
\cmidrule(lr){2-7}
& \multirow{2}{*}{0.2} & Gen PPL $\downarrow$ & 97.8571 & 80.2788 & 46.4174 & 36.0494\\
& & Entropy $\uparrow$ & 5.5267 & 5.4656 & 5.3382 & 5.2489\\
\cmidrule(lr){2-7}
& \multirow{2}{*}{0.3} & Gen PPL $\downarrow$ & 96.8045 & 80.8193 & 46.2536 & 36.4045\\
& & Entropy $\uparrow$ & 5.5187 & 5.4681 & 5.3523 & 5.2345\\
\midrule

\multirow{6}{*}{\shortstack[l]{$[0.75, 0.95]$}}
& \multirow{2}{*}{0.1} & Gen PPL $\downarrow$ & 98.9172 & 80.3999 & 46.2176 & 36.8800\\
& & Entropy $\uparrow$ & 5.5264 & 5.4639 & 5.3496 & 5.2355\\
\cmidrule(lr){2-7}
& \multirow{2}{*}{0.2} & Gen PPL $\downarrow$ & 98.4236 & 81.4000 & 46.7018 & 37.2678\\
& & Entropy $\uparrow$ & 5.5262 & 5.4768 & 5.3453 & 5.2457\\
\cmidrule(lr){2-7}
& \multirow{2}{*}{0.3} & Gen PPL $\downarrow$ & 97.5960 & 80.5485 & 46.8595 & 36.5579\\
& & Entropy $\uparrow$ & 5.5210 & 5.4687 & 5.3531 & 5.2423\\
\midrule

\multirow{6}{*}{\shortstack[l]{$[0.85, 0.90]$}}
& \multirow{2}{*}{0.1} & Gen PPL $\downarrow$ & 98.6561 & 80.5434 & 47.7492 & 36.8783\\
& & Entropy $\uparrow$ & 5.5192 & 5.4795 & 5.3718 & 5.2484\\
\cmidrule(lr){2-7}
& \multirow{2}{*}{0.2} & Gen PPL $\downarrow$ & 99.3588 & 79.7671 & 47.5099 & 36.3204\\
& & Entropy $\uparrow$ & 5.5297 & 5.4734 & 5.3559 & 5.2056\\
\cmidrule(lr){2-7}
& \multirow{2}{*}{0.3} & Gen PPL $\downarrow$ & 98.3399 & 79.6512 & 47.7224 & 37.3283\\
& & Entropy $\uparrow$ & 5.5244 & 5.4711 & 5.3672 & 5.2439\\
\midrule

\multirow{6}{*}{\shortstack[l]{$[0.85, 0.95]$}}
& \multirow{2}{*}{0.1} & Gen PPL $\downarrow$ & 98.4593 & 79.7479 & 47.3798 & 37.1501\\
& & Entropy $\uparrow$ & 5.5246 & 5.4711 & 5.3686 & 5.2551\\
\cmidrule(lr){2-7}
& \multirow{2}{*}{0.2} & Gen PPL $\downarrow$ & 99.1255 & 78.1578 & 46.5278 & 36.6537\\
& & Entropy $\uparrow$ & 5.5231 & 5.4590 & 5.3563 & 5.2321\\
\cmidrule(lr){2-7}
& \multirow{2}{*}{0.3} & Gen PPL $\downarrow$ & 98.2182 & 79.6038 & 47.8894 & 36.5796\\
& & Entropy $\uparrow$ & 5.5240 & 5.4698 & 5.3720 & 5.2358\\
\bottomrule
\end{tabular*}}
\end{table}
\FloatBarrier

\FloatBarrier
\begin{table}[!t]
\centering
\footnotesize
\caption{Results for \texttt{fp\_mdlm} with fixed strategy for masked $\gamma$ range $[0.75, 0.90]$. Constant settings for all runs: visible $\gamma = 1.0$. Values are shown to four decimal places.}
\label{tab:fp_mdlm_changed_gamma_075_090}
\setlength{\tabcolsep}{1.0pt}
\renewcommand{\arraystretch}{0.70}
\sisetup{detect-weight=true, detect-inline-weight=math}
\resizebox{\linewidth}{!}{
\begin{tabular*}{\linewidth}{@{\extracolsep{\fill}} l c c l c c c c @{}}
\toprule
& & & & \multicolumn{4}{c}{\textbf{Budget}} \\
\cmidrule(l){5-8}
\textbf{Masked $\gamma$ range}
& \boldmath$\gamma_{\mathrm{changed},\min}$
& \boldmath$\gamma_{\mathrm{changed},\max}$
& \textbf{Metric}
& \textbf{96}
& \textbf{192}
& \textbf{384}
& \textbf{768} \\
\midrule

\multirow{16}{*}{$[0.75, 0.90]$}

& \multirow{2}{*}{0.00}
& \multirow{2}{*}{0.00}
& Gen PPL $\downarrow$ & 101.4907 & 62.5693 & 42.3873 & 37.8086\\
& & & Entropy $\uparrow$ & 5.4348 & 5.3472 & 5.1515 & 5.1822\\
\cmidrule(lr){2-8}

& \multirow{2}{*}{0.10}
& \multirow{2}{*}{0.10}
& Gen PPL $\downarrow$ & 100.9282 & 62.0509 & 41.9018 & 37.2647\\
& & & Entropy $\uparrow$ & 5.4300 & 5.3373 & 5.1356 & 5.1530\\
\cmidrule(lr){2-8}

& \multirow{2}{*}{0.20}
& \multirow{2}{*}{0.20}
& Gen PPL $\downarrow$ & 101.1447 & 60.5738 & 40.7417 & 37.1662\\
& & & Entropy $\uparrow$ & 5.4351 & 5.3447 & 5.0733 & 5.1485\\
\cmidrule(lr){2-8}

& \multirow{2}{*}{0.25}
& \multirow{2}{*}{0.25}
& Gen PPL $\downarrow$ & 99.7740 & 61.9484 & 41.2334 & 37.7585\\
& & & Entropy $\uparrow$ & 5.4340 & 5.3493 & 5.1347 & 5.1718\\
\cmidrule(lr){2-8}

& \multirow{2}{*}{0.30}
& \multirow{2}{*}{0.30}
& Gen PPL $\downarrow$ & 100.9554 & 61.3730 & 42.3543 & 36.9675\\
& & & Entropy $\uparrow$ & 5.4388 & 5.3411 & 5.0945 & 5.1459\\
\cmidrule(lr){2-8}

& \multirow{2}{*}{0.00}
& \multirow{2}{*}{0.10}
& Gen PPL $\downarrow$ & 97.2015 & 81.5204 & 45.9489 & 36.8647\\
& & & Entropy $\uparrow$ & 5.5201 & 5.4752 & 5.3500 & 5.2429\\
\cmidrule(lr){2-8}

& \multirow{2}{*}{0.00}
& \multirow{2}{*}{0.20}
& Gen PPL $\downarrow$ & 97.8571 & 80.2788 & 46.4174 & 36.0494\\
& & & Entropy $\uparrow$ & 5.5267 & 5.4656 & 5.3382 & 5.2489\\
\cmidrule(lr){2-8}

& \multirow{2}{*}{0.00}
& \multirow{2}{*}{0.30}
& Gen PPL $\downarrow$ & 96.8045 & 80.8193 & 46.2536 & 36.4045\\
& & & Entropy $\uparrow$ & 5.5187 & 5.4681 & 5.3523 & 5.2345\\

\bottomrule
\end{tabular*}}
\end{table}
\FloatBarrier

\subsection{Pretrained MDLM Conversion Details}
\label{appendix:conversion_details}

In this part, we give further details and context on how we initialize the FP-MDLM model when we adapt it, as presented in Section~\ref{sub:model_adaptation}, and why these design choices are motivated.

\subsubsection{Analysis of the pretrained MDLM checkpoint}
\label{appendix:analysis_pretrained_mdlm}

\paragraph{Layer similarity analysis with CKA.}
To better understand how to convert a pretrained MDLM into a FP model, we analyze the similarity of hidden representations across transformer layers. We use Linear Centered Kernel Alignment (CKA)~\citep{kornblith2019similarityneuralnetworkrepresentations}, a standard representation-similarity measure that compares whether two layers encode examples with similar geometry. For each timestep $t$, we collect the residual-stream activations of every transformer layer on held-out OWT batches. For layer $l$, we flatten batch and sequence dimensions to obtain a feature matrix
$$
    X_l^{(t)} \in \mathbb{R}^{N \times d},
$$
where each row corresponds to one token representation. Given two centered feature matrices $X$ and $Y$, Linear CKA is defined as
$$
    \mathrm{CKA}(X,Y)
    =
    \frac{
        \left\| X^\top Y \right\|_F^2
    }{
        \left\| X^\top X \right\|_F
        \left\| Y^\top Y \right\|_F
    }.
$$
CKA is close to $1$ when two layers induce very similar representation geometry over the same token samples, and close to $0$ when their representations are largely unrelated. We therefore use CKA to identify redundant groups of layers and potential boundaries between qualitatively different representation regimes.
\FloatBarrier
\begin{figure}[H]
    \centering
    \begin{subfigure}[b]{0.48\textwidth}
        \centering
        \includegraphics[width=\textwidth]{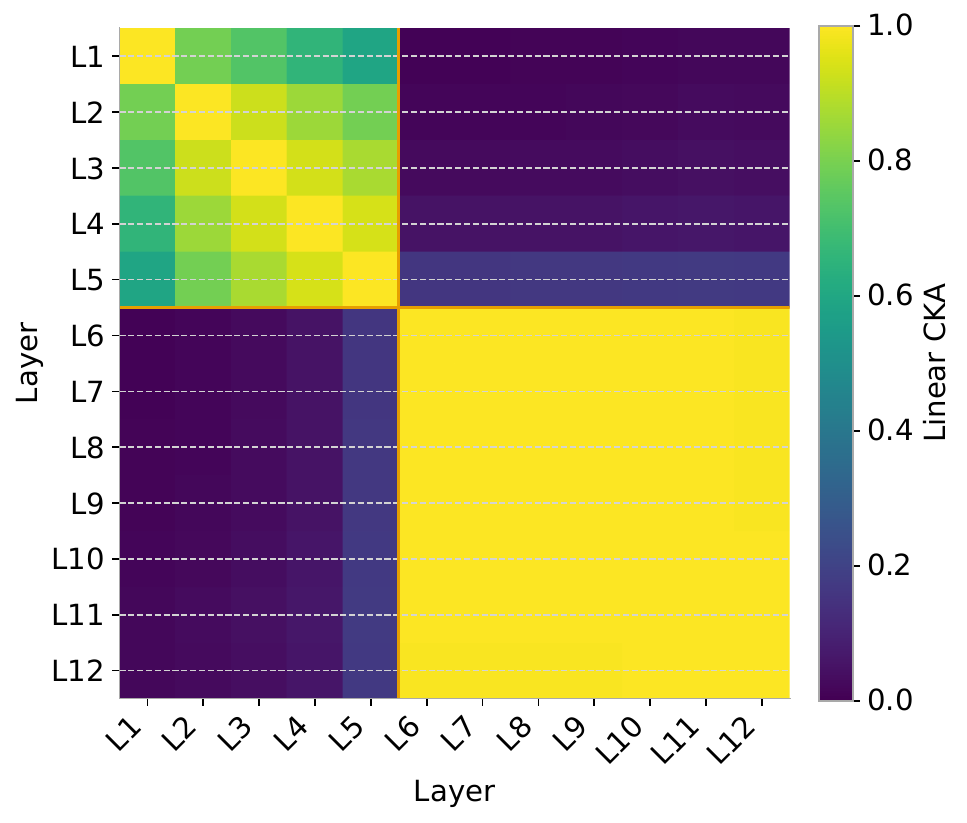}
        \caption{Mean Linear CKA across timesteps.}
        \label{fig:cka_heatmap}
    \end{subfigure}
    \hfill
    \begin{subfigure}[b]{0.48\textwidth}
        \centering
        \includegraphics[width=\textwidth]{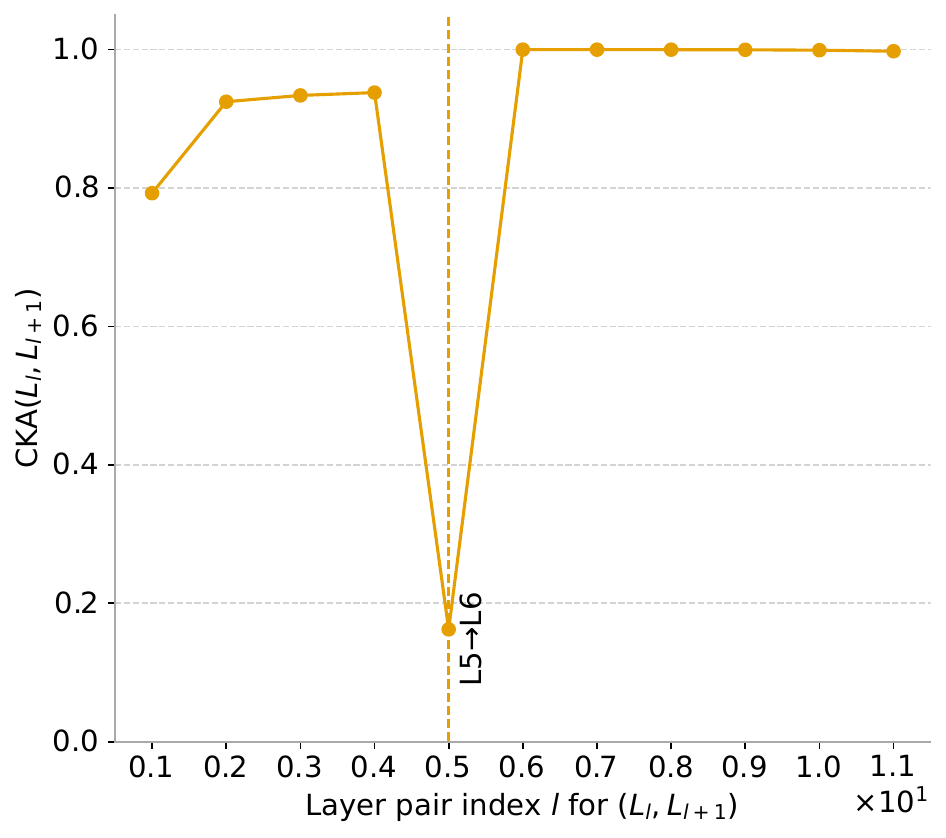}
        \caption{Consecutive-layer CKA.}
        \label{fig:cka_consecutive}
    \end{subfigure}
    \caption{\textbf{Representation similarity in a pretrained MDLM.}
    We compute Linear CKA between residual-stream activations of all transformer layers at timesteps $t\in\{0.1,0.3,0.5,0.7,0.9\}$, then average the similarities across timesteps.
    The heatmap shows a clear two-stage structure: layers 1--5 form an early block, layers 6--12 form a highly self-similar late block, and cross-block similarity is low.
    The consecutive-layer plot shows a sharp drop between layers 5 and 6, followed by near-saturated similarity among the later layers.}
    \label{fig:cka_layer_similarity}
\end{figure}
\FloatBarrier

\paragraph{Results and interpretation.}
Figure~\ref{fig:cka_layer_similarity} shows that the pretrained MDLM has a pronounced block structure in depth. Averaged across timesteps, Linear CKA identifies the strongest and most stable boundary between layers 5 and 6; the same boundary is recovered at all analyzed timesteps. Layers 1--5 form a moderately coherent early stage, with average within-block similarity $0.808$, while layers 6--12 form an extremely tight late stage, with average within-block similarity $0.998$. In contrast, similarity across the two blocks is very low, $0.060$, indicating that the later layers operate in a representation regime that is strongly separated from the earlier layers. The consecutive-layer profile makes this transition especially visible: similarity increases gradually through the early stack, drops sharply from $0.938$ between layers 4 and 5 to $0.162$ between layers 5 and 6, and then remains nearly saturated from layer 6 onward. This suggests that the deeper part of the pretrained MDLM is highly redundant and already behaves like repeated refinement in a shared representation space. This provides empirical support for replacing the later transformer stack with a shared fixed-point block.

\subsubsection{Layer mapping and initialization}

\begin{figure}[H]
    \centering
    \includegraphics[width=\textwidth]{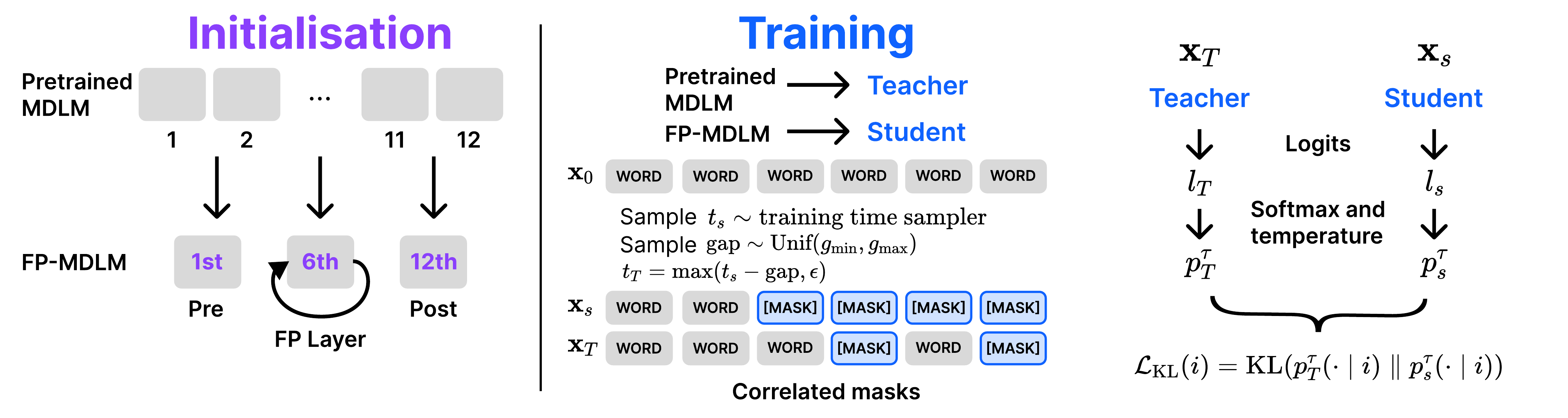}
    \caption{\textbf{Adapting an MDLM checkpoint into FP-MDLM.} (Left) FP-MDLM is initialized by mapping layers from a pretrained MDLM checkpoint to the preprocessing, fixed-point, and postprocessing blocks (see Appendix~\ref{appendix:analysis_pretrained_mdlm}). (Right) we then run a short adaptation stage with a teacher--student KL loss on logits, using correlated masks at two nearby noise levels, where the teacher input is less noisy than the student input.}
    \label{fig:adaptation-mapping}
\end{figure}

Motivated by the CKA analysis above, we initialize the converted FP-MDLM by mapping representative layers from the pretrained MDLM into the fixed-point architecture. Since layers 6-12 form a highly coherent late-stage block, we use layer 6 to initialize the shared fixed-point block. We keep the boundary layers explicit by mapping layer 1 to the preprocessing block and layer 12 to the postprocessing block (see Figure \ref{fig:adaptation-mapping}, Left).
\FloatBarrier
\FloatBarrier

\subsubsection{Correlated mask construction}
\label{appendix:correlated-masks}
For both logits-KL adaptation against a teacher checkpoint and the consistency loss $\mathcal{L}_{\mathrm{CONS}}$, we construct the student and cleaner masks in a correlated rather than independent way. We first sample a student noise level $t_s$, then obtain a cleaner level $t_c$ by subtracting a random gap $\Delta \sim \mathcal{U}[\mathrm{gap}_{\min}, \mathrm{gap}_{\max}]$. These two times are converted into keep probabilities $\alpha_s$ and $\alpha_c$, with $\alpha_c \geq \alpha_s$. For each token position, we draw a single uniform random value $u_i$ and reuse it for both branches:
\[
z_s^i =
\begin{cases}
\mask, & u_i < 1-\alpha_s,\\
x^i, & \text{otherwise},
\end{cases}
\qquad
z_c^i =
\begin{cases}
\mask, & u_i < 1-\alpha_c,\\
x^i, & \text{otherwise}.
\end{cases}
\]
Because the same random numbers are reused, the two masks are nested: the cleaner mask is a subset of the student mask. Thus, the cleaner branch always sees an equally clean or cleaner context. This avoids extra variance from unrelated masking patterns and makes the consistency comparison focus on a controlled change in corruption level. In the rare case where both masks would otherwise be identical, we unmask one cleaner position so that the cleaner branch is strictly less noisy. The consistency term is evaluated only on positions masked in the student input.

\subsubsection{Effect on the training loss when using or not initialization}
\FloatBarrier
\begin{figure}[H]
    \centering
    
    \includegraphics[width=0.5\textwidth]{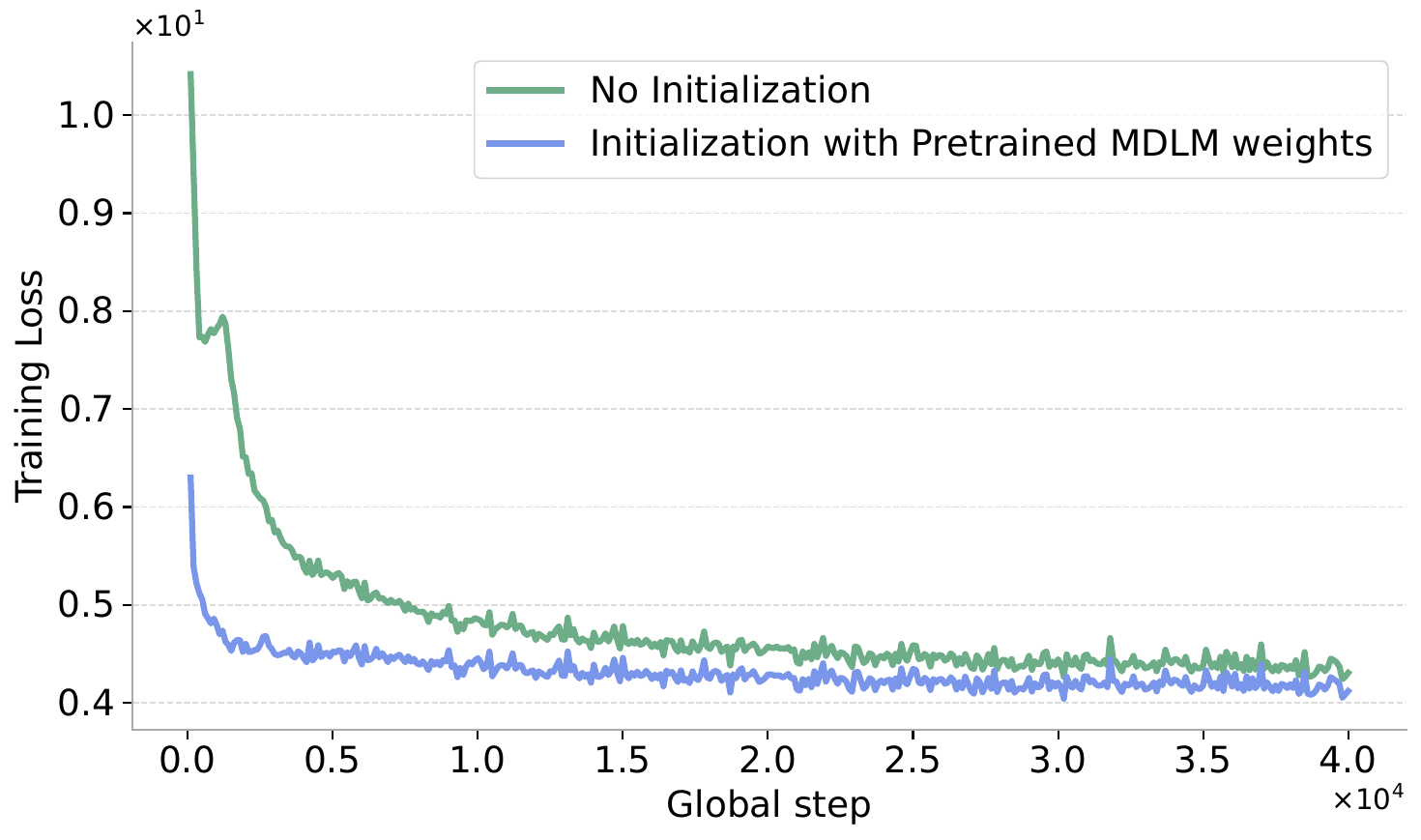}
    \caption{Comparison of the training loss between no initialization and initialization with pretrained MDLM weights when distilling the MDLM model into a FP model.}
    \label{fig:init_vs_noinit_loss}
\end{figure}
\FloatBarrier
Figure~\ref{fig:init_vs_noinit_loss} shows that initializing the FP model from pretrained MDLM weights provides a better starting point for short adaptation. The initialized run starts with a substantially lower training loss and remains below the non-initialized run throughout training. Both runs eventually approach similar loss values, but pretrained initialization reaches the low-loss regime much faster, which supports using layer mapping before the KL adaptation stage.

\section{Experimental Details}
\label{appendix:experimental_details}

\subsection{Hyperparameter Tuning Protocol}
\label{appendix:tuning_protocol_details}
We detail in this section how we tune the majority of the hyperparameters introduced by our methods.

\subsubsection{Fixed-Point MGM hyperparameters}

In order to tune the hyperparameters of our fixed-point models, we use a simple staged protocol. For language modeling, all tuning is done on OWT for 100k training steps with sequence length 128, which provides a fast and inexpensive proxy before running the full setup. We first tune the architecture of the fixed-point backbone, namely the number of preprocessing and postprocessing layers, while keeping the solver settings at the default values from \citet{bai_fixed_2024}. We then tune the solver iteration budget, and finally the learning rate. The full procedure is as follows:

\begin{enumerate}
  \item \textbf{Choice of tuning setup.}
  We first select the target setting for hyperparameter tuning. For language modeling, we tune on OWT for 100k steps with sequence length 128 in order to obtain fast and cheap comparisons.

  \item \textbf{Architecture tuning.}
  We tune the number of preprocessing and postprocessing layers in the fixed-point backbone, while keeping the fixed-point solver hyperparameters at the default values of \citet{bai_fixed_2024}. We evaluate a small grid of candidate architectures and retain the best-performing one.

  \item \textbf{Solver tuning.}
  With the architecture fixed, we tune the fixed-point solver budget, including the number of no-gradient and with-gradient iterations. This isolates the effect of the implicit solver from that of the backbone architecture.

  \item \textbf{Learning-rate tuning.}
  With both the architecture and solver settings fixed, we tune the learning rate over a logarithmic grid and select the value that gives the best validation performance.

  \item \textbf{Boundary check.}
  Whenever the best hyperparameter lies at the edge of the tested range, we extend the search range and repeat the evaluation until the selected value is not on the boundary.

  \item \textbf{Final selection.}
  Finally, we choose the configuration that performs best under this tuning protocol and use it for the full training runs.
\end{enumerate}

\subsubsection{Learning-rate and solver tuning}
\label{appendix:tuning_protocol_other_hp}

We tune the main optimization and solver hyperparameters through small-scale experiments before running the full training jobs. We first test the base learning rate used for MDLM. However, with the original learning-rate setting, the FP-MDLM training does not converge, as seen in previous work \citep{Prairie2026ParcaeSL}. We therefore progressively lower the learning rate until the training dynamics become stable and the validation loss shows consistent convergence.
We use the same gradient clipping as MDLM, as detailed in \citep{sahoo_simple_2024}.

\subsubsubsection{Adaptation hyperparameters}
For adaptation, we keep the optimizer and model hyperparameters fixed to the base FP-MDLM setup and tune only the adaptation-specific hyperparameters. In particular, we tune the consistency weight $\lambda$, the distillation temperature $\tau$, and the teacher-student noise-gap range $[\mathrm{gap}_{\min}, \mathrm{gap}_{\max}]$. In our base adaptation setting, we use logits-KL with $\lambda=0.1$, $\tau=1.5$, $\mathrm{gap}_{\min}=0.05$, and $\mathrm{gap}_{\max}=0.30$, apply the loss only on positions masked in the student input, and linearly warm up the consistency weight over the first 5k steps. We select these hyperparameters using a small validation sweep, choosing the setting with the best generative perplexity at matched training compute.

\subsection{Language Modeling Setup}
\label{appendix:language_modeling_details}

For FP-MDLM, each denoising step applies the explicit preprocessing block, then solves the implicit fixed-point layer, and finally applies the explicit postprocessing block. 

We evaluate sampling quality under different compute budgets by sweeping the total forward-pass budget and the number of denoising steps. Unless stated otherwise, we use the ancestral sampler with no additional noise-removal step for MDLM , \texttt{float64} logits before sampling, and compute sample quality with GPT-2 Large generative perplexity and sample entropy for text. We report both sampling quality and generation speed in tokens per second.

We do not include LM1B \citep{chelbaOneBillionWord2014} because our primary evaluation target is generation quality under limited sampling budgets, rather than likelihood modelling alone. LM1B is most commonly used to report validation/test perplexity, which would mainly evaluate likelihood estimation and would not directly exercise the core advantages of CoFRe: adaptive-depth iterative masked generation and reuse across denoising steps. We therefore focus on settings such as OWT generation, where sample quality and quality–cost trade-offs can be measured more directly.

\subsubsection{OpenWebText}

For language modeling, we evaluate on OWT with context length 1024, sentence packing, and the GPT-2 tokenizer. We reserve the last 100k documents for validation. The MDLM baseline follows \citet{sahoo_simple_2024}: a 12-layer Diffusion Transformer with hidden size 768, RoPE, dropout 0.1, Adam with learning rate $3\times10^{-4}$, global batch size 512, EMA decay 0.9999, and 1M training steps. FP-MDLM uses the same data, tokenizer, and objective, but replaces the middle transformer stack with a fixed-point block. We report generative perplexity using GPT-2 Large and unigram entropy across fixed transformer-block budgets. We use a decreasing order of number of sampling steps as we found that it performs best.

\subsubsection{Downstream evaluation}
\label{sub:downstream-eval}

We evaluate downstream performance with \texttt{lm-eval-harness} \citep{eval-harness}, following the masked-model evaluation protocol of \citet{deschenaux2024promisesoutlookschallengesdiffusion} and \citet{nieScalingMaskedDiffusion2025}. Because the harness is designed for autoregressive models, we adapt its scoring rule to masked generative models: each answer choice is scored using the variational likelihood bound available for MDLM and FP-MDLM/CoFRe, and the highest-scoring choice is selected.

\subsection{Image Modeling Setup}
\label{appendix:image_modeling_details}

For image generation, we evaluate on ImageNette at $256\times256$ resolution. Images are center-cropped, resized, and tokenized into a $16\times16$ grid of discrete latent codes, giving a sequence length of 256. We use the ImageFolder VQ-4096/XQGAN-4096 tokenizer~\citep{li2024imagefolder,li2024xq} and follow the MaskGIT/Halton setup of \citet{besnierHaltonSchedulerMasked2025}. We compare MaskGIT-Large and FP-MaskGIT under the same training and sampling protocol: AdamW optimizer, learning rate $5\times10^{-4}$, weight decay 0.03, cosine learning-rate schedule, 1500 warmup steps, batch size 128, classifier-free guidance, and Halton sampling. We use Transformer dropout 0.1 and class-label dropout 0.1 for classifier-free guidance. Following \citet{besnierHaltonSchedulerMasked2025}, the MaskGIT baseline uses one register \citep{darcetVisionTransformersNeed2024}. FP-MaskGIT keeps the same tokenizer, masked-token prediction objective, and decoding procedure as MaskGIT-Large, but replaces the middle transformer stack with a fixed-point block. We report FID, IS, training time, and peak VRAM usage. 

\paragraph{Halton sampler}
\label{appendix:halton}
While effective, \citet{besnierHaltonSchedulerMasked2025} identified a notable drawback: confidence-guided sampling typically leads to spatially clustered token decoding, as the denoiser is inherently more confident near already-populated regions. Because MGMs sample positions independently from the product of marginals $\prod_{\ell \in \mathcal{S}} p_\theta(x^\ell \mid \mathbf{z}_\tau)$ instead of the exact joint distribution, clusters of neighboring tokens are more likely to produce spatial inconsistencies. To address this, \citet{besnierHaltonSchedulerMasked2025} propose to use low-discrepancy sequences \citep{haltonRadicalInverse1964} to guarantee more uniform spatial coverage during decoding. This modification prevents extreme clustering and leads to improved FID and IS compared to standard confidence sampling.

\subsection{Sampling Precision}
\label{appendix:sampling_precision}

\citet{zhengMaskedDiffusionModels2025} found that when Masked Diffusion Models sample with low-precision logits, some logits can underflow. This can reduce the variety of sampled tokens and make Generative Perplexity look better than it really is. Because of this, we cast all logits to FP64 before sampling.

\subsection{Training Costs and Resources}
\label{appendix:resources}
All sampling experiments, for both text and images, are run on NVIDIA A100 GPUs with either 40GB or 80GB of memory. FP-MaskGIT training is also performed on A100 GPUs. For the text experiments, both MDLM and FP-MDLM are trained on 8 NVIDIA H200 GPUs with 141GB of memory per GPU. Unless otherwise stated, all reported latency, throughput, training-time, and VRAM measurements use these hardware settings.

\section{Metrics details}
\label{appendix:metrics}
In this section, we detail the main metrics used to monitor the performance of the baseline and proposed models and methods of this paper.

\subsection{Generative perplexity}
We evaluate generated text using \emph{generative perplexity}, following prior work on discrete diffusion language models \citep{louDiscreteDiffusionModeling2024, sahoo_simple_2024, deschenaux2024beyond}. This metric measures how well a strong autoregressive reference model predicts samples generated by our model. Concretely, we generate $N_{\mathrm{samp}}$ samples and score them with GPT-2 Large:
\begin{equation}
\mathrm{GenPPL}
=
\exp\!\left(
-\frac{1}{N_{\mathrm{samp}}}\sum_{n=1}^{N_{\mathrm{samp}}}
\frac{1}{L_n}\sum_{i=1}^{L_n}
\log p_{\mathrm{GPT\mbox{-}2\,Large}}(y_i^{(n)} \mid \mathbf{y}_{<i}^{(n)})
\right),
\end{equation}
where $L_n$ is the length of generated sample $\mathbf{y}^{(n)}$, and
$p_{\mathrm{GPT\mbox{-}2\,Large}}(y_i^{(n)} \mid \mathbf{y}_{<i}^{(n)})$
is the probability assigned by GPT-2 Large \citep{radford2019language} to token $y_i^{(n)}$ given its prefix.
Lower values indicate that generated text is more predictable under the reference language model and are therefore better.

As noted in prior work, sampling precision can affect this metric substantially. In particular, low-precision logits may artificially reduce token diversity and make generative perplexity appear better than it truly is. To avoid this issue, we cast logits to \texttt{float64} before sampling in all generative-perplexity evaluations \citep{zhengMaskedDiffusionModels2025}.

\subsection{Unigram Entropy}
Generative perplexity alone can reward degenerate text, for example if a model produces repetitive or low-diversity samples. To detect such failures, we also report the \emph{unigram entropy} of generated text, following prior work \citep{dieleman2022continuousdiffusioncategoricaldata, deschenauxPartitionGenerativeModeling2026}. For a generated sequence $\mathbf{y}^{(n)}$ of length $L_n$, let $c(v,\mathbf{y}^{(n)})$ denote the number of occurrences of token $v$. The unigram entropy is
\begin{equation}
H_{\mathrm{uni}}
=
-\frac{1}{N_{\mathrm{samp}}}\sum_{n=1}^{N_{\mathrm{samp}}}\sum_{v \in \mathcal{V}}
\frac{c(v,\mathbf{y}^{(n)})}{L_n}
\log \frac{c(v,\mathbf{y}^{(n)})}{L_n}.
\end{equation}
Higher values indicate more diverse token usage, while unusually low entropy can reveal collapse or repetitive generation. We therefore interpret unigram entropy jointly with generative perplexity rather than in isolation.

\subsection{Fréchet Inception Distance and Inception Score}
FID embeds real and generated images using a pretrained Inception network, fits a Gaussian distribution to each set of features, and computes the Fréchet distance between the two Gaussians:
$$
\mathrm{FID}
=
\|\mu_r - \mu_g\|_2^2
+
\mathrm{Tr}\left(\Sigma_r + \Sigma_g - 2(\Sigma_r \Sigma_g)^{1/2}\right),
$$
where $(\mu_r, \Sigma_r)$ and $(\mu_g, \Sigma_g)$ are the empirical feature mean and covariance of real and generated images. Lower FID indicates that generated images better match the real data distribution in Inception feature space. IS instead evaluates the class predictions of the Inception network on generated images:
$$
\mathrm{IS}
=
\exp\left(
\mathbb{E}_{x}
\mathrm{KL}\left(p(y \mid x) \,\|\, p(y)\right)
\right).
$$
It is high when individual generated images produce confident class predictions while the marginal class distribution remains diverse. 

Following the protocol used in the PGM paper (except the number of generated images), we compute these metrics on 10,000 generated images for efficiency, rather than the more common 50,000 \citep{deschenauxPartitionGenerativeModeling2026}. When comparing models, we keep the evaluation protocol fixed across all methods.

\subsection{Throughput and Latency}
We measure inference efficiency using both \emph{latency} and \emph{throughput}. Latency is the wall-clock time required to generate a batch of samples under a fixed sampling budget. Throughput is the amount of generated output per second, reported as tokens/s for language and images/s for image generation. Lower latency and higher throughput are better.
We report these metrics after a short warmup phase and average them over repeated runs on a single device. Unless stated otherwise, all models are evaluated with the same batch size, numerical precision, and hardware setup to ensure a fair comparison (which is A100 40 or 80GB and \texttt{float64} for sampling text, following \citep{zhengMaskedDiffusionModels2025}). More details regarding these results in Appendix~\ref{appendix:latency-res}

\subsection{Training time and VRAM used}
\label{appendix:training_time_vram}

We report training time as the wall-clock time required to complete the corresponding training run, measured from the first optimization step to the final checkpoint. Unless stated otherwise, this does not include offline preprocessing, dataset download, evaluation, or sampling. For post-training stages such as $\mathcal{L}_{\mathrm{CONS}}$, SDTT, or checkpoint adaptation, we report the additional number of training steps separately when relevant, since these stages are short compared to full pretraining.

We report VRAM as the peak GPU memory used during training, measured per GPU and expressed in GiB/GPU. For language modeling, MDLM and FP-MDLM are trained on 8 NVIDIA H200 GPUs. For image modeling, FP-MaskGIT is trained on NVIDIA A100 GPUs. Sampling experiments for both text and images are run on NVIDIA A100 GPUs with either 40GB or 80GB of memory. All training-time and VRAM comparisons are therefore intended to compare models within the same experimental setting and hardware configuration.

\section{Additional Results}
\label{appendix:additional_results}

\subsection{Extended OWT Generation Results}
\label{appendix:owt_full_results}

In this section we detail all the results for FP-MDLM, MDLM and CoFRe on OpenWebText.

\subsubsection{Samples}

We provide short uncurated samples to complement Gen. PPL and entropy. They illustrate typical low-budget failure modes: local fluency can be reasonable, but generations may drift semantically, repeat entities or phrases, and lose long-range coherence.

\begin{tcolorbox}[
    colback=gray!5,
    colframe=gray!40,
    coltitle=black,
    title=\textbf{Sample 1: repetition and semantic drift},
    fonttitle=\bfseries,
    boxrule=0.5pt,
    arc=2pt,
    left=4pt,
    right=4pt,
    top=4pt,
    bottom=4pt,
    breakable
]
\small
Researchers from the Medical Research Institute of Germany's Centre for Dermatology and Research Center for Brain Research, in Berlin recently, unearthed evidence from more than 50 decades in Berlin, when ARTISTS ate meals on seven different occasions -- dishes in toilet tissue drippings; scraps of 22-ounce paper bags and towels; glove boxes, and perfumed socks, wrapped around their bedsits.

Some say that while the press claims that Germans have forged a legend in the Hollywood musical \emph{Waters on the White Strip}, a portrait of Rolling Nirvana frontman Billie got published in \emph{Times Magazine}. Perhaps most astonishing is Jonathan Weber, a native German born to immigrant parents who grew up in Berlin, Germany, earning only \$185 for his meals.

The biscuit consists of a sparge of rare, handcrafted alloy of peppers, beef kidneys, pork, pork and oil. ``I have got the biscuits from German kitchens,'' he says. ``I also got the food from German restaurants. I even got ramen noodles from my own supermarket.''

\medskip
\textit{Failure mode: the sample remains locally grammatical, but drifts between unrelated topics and repeats food/Germany motifs with weak semantic consistency.}
\end{tcolorbox}

\begin{tcolorbox}[
    colback=gray!5,
    colframe=gray!40,
    coltitle=black,
    title=\textbf{Sample 2: fluent fragments with topic drift},
    fonttitle=\bfseries,
    boxrule=0.5pt,
    arc=2pt,
    left=4pt,
    right=4pt,
    top=4pt,
    bottom=4pt,
    breakable
]
\small
\texttt{\textless{}|endoftext|\textgreater{}}'s website, wellness.com.com, provides a nurturing health and wellness platform for professionals in the business, health and wellness industries. In the last years we have expanded our wellness Topics Page, post wellness information page on our website and Facebook page and has created a dedicated 24 hour monthly wellness page.

\texttt{\textless{}|endoftext|\textgreater{}}Schedstein's favorite musical savior

This Hammerstein's was the most memorable year of the millennium. The celebration of Hammerstein's included an ``open mic'' festival, performances, movie screenings, and a parade of reviews, all musical from around the world, from MTV to music festivals.

Hopefully this is a forever, this is a this. This is a red. Hopefully it will always be a red. Call it. I can call it that but that warm smile is on. The color of smile. That warm breath is the bottom of ``cheeks.''

\medskip
\textit{Failure mode: the generation contains fluent local fragments, but abruptly switches topics and degenerates into repetitive, low-information phrasing.}
\end{tcolorbox}

\subsubsection{MDLM vs FP-MDLM}

Table~\ref{tab:mdlm_fpmdlm_budget_grouped} compares MDLM and FP-MDLM across fixed transformer-block budgets on OWT. FP-MDLM substantially improves the low-budget regime: at budget 96, generative perplexity drops from 830.82 to 375.63, and at budget 192 from 343.33 to 273.28, while maintaining similar unigram entropy. This shows that replacing part of the denoiser with a fixed-point block improves the quality--cost trade-off when sampling compute is limited. At larger budgets, however, the standard MDLM becomes stronger, suggesting that the base FP-MDLM architecture alone mainly benefits the low-budget regime and requires additional regularization for strong high-budget generation.

Table~\ref{tab:validation_metrics} reports the corresponding validation perplexity and efficiency metrics. FP-MDLM uses substantially fewer parameters than MDLM, reducing the model size from 170M to 104M parameters. It also lowers training time and VRAM usage, from approximately 139h and 112.4 GiB/GPU to approximately 123h and 93.44 GiB/GPU, while having a slightly lower latency (Appendix~\ref{appendix:latency-res}). The validation perplexity is worse than MDLM, which is expected from the reduced parameter count and weight sharing, but the generation results in Table~\ref{tab:mdlm_fpmdlm_budget_grouped} show that this trade-off is favorable under low sampling budgets.

\FloatBarrier
\begin{table}[H]
\centering
\small
\caption{Generation quality across compute budgets for MDLM and FP-MDLM on OWT. The budget counts the total number of transformer-block forward passes. Training time and training VRAM are reported in the column headers. For MDLM, the budget is obtained by multiplying the number of denoising steps by 12, corresponding to the 12-layer backbone. FP-MDLM results are reported without reuse.}
\label{tab:mdlm_fpmdlm_budget_grouped}
\setlength{\tabcolsep}{1.2pt}
\renewcommand{\arraystretch}{0.72}
\sisetup{detect-weight=true, detect-inline-weight=math}

\begin{tabular*}{\linewidth}{@{\extracolsep{\fill}} r c c c c @{}}
\toprule
\textbf{Budget}
& \multicolumn{2}{c}{\shortstack{\textbf{MDLM}\\\scriptsize 12 transformer layers}}
& \multicolumn{2}{c}{\shortstack{\textbf{FP-MDLM}\\\scriptsize no reuse}} \\
\cmidrule(lr){2-3}
\cmidrule(lr){4-5}
& \textbf{Gen. PPL} $\downarrow$
& \textbf{Entropy} $\uparrow$
& \textbf{Gen. PPL} $\downarrow$
& \textbf{Entropy} $\uparrow$ \\
\midrule

96   & 830.8200 & 5.9100 & 375.6314 & 5.8102 \\
192  & 343.3300 & 5.8100 & 273.2752 & 5.7630 \\
384  & 196.7900 & 5.7500 & 215.1965 & 5.7259 \\
768  & 143.8800 & 5.7000 & 179.6546 & 5.7016 \\
1536 & 120.7700 & 5.6700 & 158.5044 & 5.6859 \\
3072 & 112.7000 & 5.6600 & 155.8161 & 5.6858 \\

\bottomrule
\end{tabular*}
\end{table}
\FloatBarrier

\FloatBarrier
\begin{table}[H]
\centering
\small
\caption{Validation perplexity, training time and VRAM on OpenWebText.}
\label{tab:validation_metrics}
\setlength{\tabcolsep}{1.2pt}
\renewcommand{\arraystretch}{0.72}
\sisetup{detect-weight=true, detect-inline-weight=math}

\begin{tabular*}{\linewidth}{@{\extracolsep{\fill}} l c c c c @{}}
\toprule
\textbf{Model} & \textbf{\#Params} & \textbf{Val. PPL $\downarrow$} & \textbf{Training time (h)} & \textbf{VRAM} \\
\midrule

\multicolumn{5}{l}{\textit{OWT (1024)}} \\

\quad MDLM
& 170M
& \textbf{23.07}
& $\approx 139$
& \textbf{112.4} GiB/GPU \\

\rowcolor{gray!15}
\quad FP-MDLM
& \textbf{104M}
& 27.45
& $\approx \mathbf{123}$
& \textbf{93.44} GiB/GPU \\

\bottomrule
\end{tabular*}
\end{table}
\FloatBarrier

\subsubsection{Initialized vs not initialized FP-MDLM adaptation}

Table~\ref{tab:gen_ppl_noinit_scratch} compares 40k-step FP-MDLM adaptation with and without initialization from a pretrained MDLM checkpoint. Pretrained initialization improves the no-reuse setting at every budget, reducing generative perplexity from 296.76 to 276.00 at budget 96 and from 149.63 to 139.00 at budget 768. This indicates that the layer-mapping initialization provides a better starting point for adapting the fixed-point architecture.

The effect is also visible when reuse is enabled. For both initialized and non-initialized models, reuse becomes more beneficial at medium and high budgets, and three-state reuse gives the best results in most of these regimes. With initialization and 3SR, the adapted FP-MDLM reaches the best overall perplexities at budgets 192, 384, and 768. These results suggest that pretrained initialization not only improves the adapted checkpoint itself, but also makes the resulting fixed-point states more reusable across denoising steps.

\FloatBarrier
\begin{table}[H]
\centering
\small
\caption{\textbf{Pretrained initialization improves short FP-MDLM adaptation.}
We compare 40k-step FP-MDLM adaptation with and without initialization from a pretrained MDLM checkpoint. For each sampling budget, we report the best generative perplexity and entropy across denoising-step sweeps under no reuse (NR), full reuse (R), and three-state reuse (3SR). Initialization improves generation quality at every budget and gives the best results with 3SR at medium and high budgets.}
\label{tab:gen_ppl_noinit_scratch}
\setlength{\tabcolsep}{1.2pt}
\renewcommand{\arraystretch}{0.72}
\sisetup{detect-weight=true, detect-inline-weight=math}

\begin{tabular*}{\linewidth}{@{\extracolsep{\fill}} l l l c c c c @{}}
\toprule
& & & \multicolumn{4}{c}{\textbf{Budget}} \\
\cmidrule(l){4-7}
\textbf{Model} & \textbf{Reuse} & \textbf{Metric} & \textbf{96} & \textbf{192} & \textbf{384} & \textbf{768}\\
\midrule

\multirow{6}{*}{\shortstack[l]{\textbf{No init}\\\scriptsize FP-MDLM\\\scriptsize @40k}}
& \multirow{2}{*}{NR}
& Gen PPL $\downarrow$ & 296.764 & 204.787 & 170.835 & 149.626 \\
&
& Entropy $\uparrow$   & 5.690 & 5.661 & 5.629 & 5.603 \\
\cmidrule{2-7}

& \multirow{2}{*}{R}
& Gen PPL $\downarrow$ & 336.057 & 201.521 & 161.845 & 131.800 \\
&
& Entropy $\uparrow$   & 5.705 & 5.662 & 5.637 & 5.590 \\
\cmidrule{2-7}

& \multirow{2}{*}{3SR}
& Gen PPL $\downarrow$ & 298.643 & 192.227 & 149.427 & 131.291 \\
&
& Entropy $\uparrow$   & 5.722 & 5.658 & 5.611 & 5.597 \\
\midrule

\multirow{6}{*}{\shortstack[l]{\textbf{Init}\\\scriptsize FP-MDLM\\\scriptsize @40k}}
& \multirow{2}{*}{NR}
& Gen PPL $\downarrow$ & 276.003 & 194.355 & 163.424 & 139.003 \\
&
& Entropy $\uparrow$   & 5.696 & 5.649 & 5.622 & 5.578 \\
\cmidrule{2-7}

& \multirow{2}{*}{R}
& Gen PPL $\downarrow$ & 312.912 & 195.589 & 149.934 & 130.443 \\
&
& Entropy $\uparrow$   & 5.702 & 5.656 & 5.617 & 5.588 \\
\cmidrule{2-7}

& \multirow{2}{*}{3SR}
& Gen PPL $\downarrow$ & 286.403 & 184.708 & 147.497 & 126.872 \\
&
& Entropy $\uparrow$   & 5.695 & 5.649 & 5.619 & 5.572 \\
\bottomrule
\end{tabular*}
\end{table}
\FloatBarrier

\subsubsection{Generation quality when adding different components of CoFRe}

In this section, we ablate the components of CoFRe and analyze how each one affects OWT generation quality. Detailed results are reported in Table~\ref{tab:fp_mdlm_component_ablation}.

\FloatBarrier
\begin{table}[H]
\centering
\small
\caption{\textbf{Component ablation for FP-MDLM generation on OpenWebText.}
We report generative perplexity and entropy across sampling budgets. The table compares MDLM, the base FP-MDLM checkpoint trained for 1M steps, the same model with inference-time reuse or three-state reuse, the checkpoint after cross-step consistency regularization, and FP-MDLM checkpoints obtained by adapting a pretrained MDLM with a 40k-step teacher-student KL loss on logits.}
\label{tab:fp_mdlm_component_ablation}
\setlength{\tabcolsep}{1.2pt}
\renewcommand{\arraystretch}{0.72}
\sisetup{detect-weight=true, detect-inline-weight=math}

\begin{tabular*}{\linewidth}{@{\extracolsep{\fill}} l l c c c c @{}}
\toprule
& & \multicolumn{4}{c}{\textbf{Budget}} \\
\cmidrule(l){3-6}
\textbf{Method} & \textbf{Metric} & \textbf{96} & \textbf{192} & \textbf{384} & \textbf{768}\\
\midrule

\multirow{2}{*}{MDLM}
& Gen PPL $\downarrow$ & 830.8200 & 343.3300 & 196.7900 & 143.8800 \\
& Entropy $\uparrow$   & 5.9100 & 5.8100 & 5.7500 & 5.7000 \\
\midrule

\multirow{2}{*}{FP-MDLM}
& Gen PPL $\downarrow$ & 375.6314 & 273.2752 & 215.1965 & 179.6546 \\
& Entropy $\uparrow$   & 5.8102 & 5.7630 & 5.7259 & 5.7016 \\
\cmidrule(lr){1-6}

\multirow{2}{*}{\shortstack[l]{FP-MDLM\\+ reuse}}
& Gen PPL $\downarrow$ & 516.677 & 253.210 & 229.409 & 269.007 \\
& Entropy $\uparrow$   & 5.815 & 5.729 & 5.728 & 2.209 \\
\cmidrule(lr){1-6}

\multirow{2}{*}{\shortstack[l]{FP-MDLM\\+ 3SR}}
& Gen PPL $\downarrow$ & 454.100 & 249.322 & 196.307 & 254.258 \\
& Entropy $\uparrow$   & 5.795 & 5.736 & 5.663 & 5.384 \\
\midrule

\multirow{2}{*}{\shortstack[l]{FP-MDLM\\+ $\mathcal{L}_{\mathrm{CONS}}$}}
& Gen PPL $\downarrow$ & 104.153 & 70.275 & 54.927 & 41.673 \\
& Entropy $\uparrow$   & 5.447 & 5.388 & 5.343 & 5.156 \\
\cmidrule(lr){1-6}

\multirow{2}{*}{\shortstack[l]{FP-MDLM\\+ $\mathcal{L}_{\mathrm{CONS}}$+ reuse}}
& Gen PPL $\downarrow$ & 117.592 & 68.550 & 50.195 & 37.567 \\
& Entropy $\uparrow$   & 5.438 & 5.387 & 5.283 & 5.142 \\
\cmidrule(lr){1-6}

\multirow{2}{*}{\shortstack[l]{FP-MDLM\\+ $\mathcal{L}_{\mathrm{CONS}}$+ 3SR}}
& Gen PPL $\downarrow$ & 101.791 & 65.182 & 48.755 & 37.846 \\
& Entropy $\uparrow$   & 5.434 & 5.380 & 5.283 & 5.142 \\
\midrule

\multirow{2}{*}{\shortstack[l]{Adapted\\FP-MDLM}}
& Gen PPL $\downarrow$ & 296.764 & 204.787 & 170.835 & 149.626 \\
& Entropy $\uparrow$   & 5.690 & 5.661 & 5.629 & 5.603 \\
\cmidrule(lr){1-6}

\multirow{2}{*}{\shortstack[l]{Adapted\\FP-MDLM+ reuse}}
& Gen PPL $\downarrow$ & 336.057 & 201.521 & 161.845 & 131.800 \\
& Entropy $\uparrow$   & 5.705 & 5.662 & 5.637 & 5.590 \\
\cmidrule(lr){1-6}

\multirow{2}{*}{\shortstack[l]{Adapted\\FP-MDLM+ 3SR}}
& Gen PPL $\downarrow$ & 298.643 & 192.227 & 149.427 & 131.291 \\
& Entropy $\uparrow$   & 5.722 & 5.658 & 5.611 & 5.597 \\

\bottomrule
\end{tabular*}
\end{table}
\FloatBarrier

Table~\ref{tab:fp_mdlm_component_ablation} decomposes the effect of the main CoFRe components. The base FP-MDLM substantially improves over MDLM in the very low-budget regime, reducing generative perplexity from 830.82 to 375.63 at budget 96. However, this advantage decreases at larger budgets, where the unregularized FP-MDLM remains worse than MDLM. This confirms that the fixed-point architecture alone improves the low-budget quality--cost trade-off, but is not sufficient for strong generation across all budgets.

Reuse alone is unstable on the base FP-MDLM checkpoint. Full reuse and 3SR improve some medium-budget results, but can hurt in other regimes; in particular, full reuse at budget 768 leads to severe degeneration, as reflected by both high generative perplexity and very low entropy. This supports the motivation for adding a regularization mechanism that makes representations smoother and more reusable across denoising steps.

Cross-step consistency is the largest contributor to generation quality. Adding $\mathcal{L}_{\mathrm{CONS}}$ reduces generative perplexity from 375.63 to 104.15 at budget 96 and from 179.65 to 41.67 at budget 768. Once the model is regularized with $\mathcal{L}_{\mathrm{CONS}}$, reuse becomes much more effective: 3SR gives the best results at budgets 96, 192, and 384, while full reuse is slightly better at budget 768. Finally, the adapted FP-MDLM rows show a complementary path to obtaining useful fixed-point denoisers from pretrained MDLM checkpoints with only a short teacher-student adaptation stage.

\subsubsection{Mauve score evaluation}

Following \citep{pillutla2021mauve,wang2025remasking}, we evaluate generation quality with MAUVE, which measures the distributional gap between model-generated and human-written text using divergence frontiers \citep{pillutla2021mauve, liu2021divergence}. Like \citep{wang2025remasking}, for each model and sampling budget, we generate $5{,}000$ samples and compare them against $5{,}000$ OpenWebText reference samples. We compute also MAUVE beside generative perplexity as a generation metric because it accounts for both sample quality and diversity, whereas perplexity alone can be uninformative for corrector-based samplers that may trade diversity for lower perplexity \citep{zhengMaskedDiffusionModels2025}. Table~\ref{tab:mdlm_cofre_mauve_5k} shows that CoFRe consistently outperforms MDLM across all budgets, with the strongest improvements at budgets $96$ and $192$, suggesting that cross-step consistency regularization and three-state reuse improve the quality--diversity trade-off of generated text. (More details in \citep{pillutla2021mauve} about MAUVE).

\FloatBarrier
\begin{table}[H]
\centering
\small
\caption{\textbf{MAUVE scores for MDLM and CoFRe generation on OpenWebText.}
We report MAUVE scores computed with $5{,}000$ generated samples and $5{,}000$ reference samples. CoFRe denotes FP-MDLM with cross-step consistency regularization and three-state reuse.}
\label{tab:mdlm_cofre_mauve_5k}
\setlength{\tabcolsep}{1.2pt}
\renewcommand{\arraystretch}{0.72}
\sisetup{detect-weight=true, detect-inline-weight=math}

\begin{tabular*}{\linewidth}{@{\extracolsep{\fill}} l l c c c c @{}}
\toprule
& & \multicolumn{4}{c}{\textbf{Budget}} \\
\cmidrule(l){3-6}
\textbf{Method} & \textbf{Metric} & \textbf{96} & \textbf{192} & \textbf{384} & \textbf{768}\\
\midrule

\multirow{1}{*}{MDLM}
& MAUVE $\uparrow$ & 0.010594 & 0.010176 & 0.010197 & 0.009641 \\
\midrule

\multirow{1}{*}{\shortstack[l]{CoFRe}}
& MAUVE $\uparrow$ & 0.013759 & 0.014188 & 0.012080 & 0.010704 \\

\bottomrule
\end{tabular*}
\end{table}
\FloatBarrier

\subsubsection{Downstream evaluation}
\label{appendix:downstream-eval}
Table~\ref{tab:mdlm_distill_results} shows that CoFRe is competitive with MDLM on downstream multiple-choice tasks. It improves on LAMBADA (39.45 vs.\ 38.52), ARC-easy (36.70 vs.\ 34.26), and BoolQ (60.21 vs.\ 49.42), while underperforming on ARC-challenge, OpenBookQA, PIQA, RACE, and SIQA. This mixed behaviour is expected: these tasks are evaluated through likelihood-based answer scoring, so CoFRe does not directly benefit from its main advantage, which appears during low-budget generation. Overall, the results suggest that CoFRe preserves reasonable likelihood-based downstream performance, while its strongest gains remain in the sampling regime.

\FloatBarrier
\begin{table}[H]
\centering
\small
\caption{Results on downstream evaluation tasks.}
\label{tab:mdlm_distill_results}
\setlength{\tabcolsep}{1.2pt}
\renewcommand{\arraystretch}{0.72}
\sisetup{detect-weight=true, detect-inline-weight=math}

\begin{tabular*}{\linewidth}{@{\extracolsep{\fill}} l c c c c c c c c @{}}
\toprule
& \textbf{LAMBADA}
& \textbf{ARC-e}
& \textbf{ARC-c}
& \textbf{BoolQ}
& \textbf{OBQA}
& \textbf{PIQA}
& \textbf{RACE}
& \textbf{SIQA} \\
\midrule

MDLM
& 38.52
& 34.26
& 24.66
& 49.42
& 28.60
& 58.27
& 28.04
& 38.84 \\

\shortstack[l]{CoFRe}
& 39.45
& 36.70
& 22.87
& 60.21
& 25.20
& 55.33
& 27.27
& 35.82 \\

\bottomrule
\end{tabular*}
\end{table}
\FloatBarrier

\subsubsection{Sampling with nucleus sampling}
As suggested by \citet{wang2025remasking, deschenauxPartitionGenerativeModeling2026}, nucleus sampling \citep{Holtzman2020The} can highly impact the generation of high-quality text
sequences. We therefore use this method (with top-p = 0.9) on both MDLM and CoFRe and compare these results in Table~\ref{tab:fp_mdlm_nucleus_sampling}.

\FloatBarrier
\begin{table}[H]
\centering
\small
\caption{\textbf{The effect of using nucleus sampling on MDLM and CoFRe generation on OpenWebText.}
We report generative perplexity and entropy across sampling budgets, comparing MDLM against CoFRe. Sampling is performed with nucleus sampling with $p=0.9$, as in \citep{wang2025remasking, deschenauxPartitionGenerativeModeling2026}.}
\label{tab:fp_mdlm_nucleus_sampling}
\setlength{\tabcolsep}{1.2pt}
\renewcommand{\arraystretch}{0.72}
\sisetup{detect-weight=true, detect-inline-weight=math}

\begin{tabular*}{\linewidth}{@{\extracolsep{\fill}} l l c c c c @{}}
\toprule
& & \multicolumn{4}{c}{\textbf{Budget}} \\
\cmidrule(l){3-6}
\textbf{Method} & \textbf{Metric} & \textbf{96} & \textbf{192} & \textbf{384} & \textbf{768}\\
\midrule
\multirow{2}{*}{MDLM}
& Gen PPL $\downarrow$ & 830.8200 & 343.3300 & 196.7900 & 143.8800 \\
& Entropy $\uparrow$   & 5.9100 & 5.8100 & 5.7500 & 5.7000 \\
\midrule
\multirow{2}{*}{MDLM+nucleus}
& Gen PPL $\downarrow$ & 292.957 & 119.284 & 69.479 & 51.187 \\
& Entropy $\uparrow$   & 5.609 & 5.538 & 5.481 & 5.433 \\
\midrule

\multirow{2}{*}{\shortstack[l]{FP-MDLM\\+ $\mathcal{L}_{\mathrm{CONS}}$+ 3SR}}
& Gen PPL $\downarrow$ & 101.791 & 65.182 & 48.755 & 37.846 \\
& Entropy $\uparrow$   & 5.434 & 5.380 & 5.283 & 5.142 \\
\midrule

\multirow{2}{*}{\shortstack[l]{FP-MDLM\\+ $\mathcal{L}_{\mathrm{CONS}}$+ 3SR + nucleus}}
& Gen PPL $\downarrow$ & 39.518 & 30.113 & 29.199 & 28.111 \\
& Entropy $\uparrow$   & 5.108 & 5.057 & 5.069 & 5.055 \\

\bottomrule
\end{tabular*}
\end{table}
\FloatBarrier

Table~\ref{tab:fp_mdlm_nucleus_sampling} shows that nucleus sampling substantially improves generation quality for both MDLM and CoFRe, but also introduces a clear tradeoff. For MDLM, using top-$p=0.9$ reduces Gen PPL from $830.82$ to $292.96$ at budget $96$, and from $143.88$ to $51.19$ at budget $768$. However, this improvement comes with lower unigram entropy, which decreases from $5.91$ to $5.61$ at budget $96$ and from $5.70$ to $5.43$ at budget $768$. The same pattern is observed for CoFRe: adding nucleus sampling to CoFRe with $\mathcal{L}_{\mathrm{CONS}}$ and 3SR further reduces Gen PPL across all budgets, reaching $28.11$ at budget $768$, while also lowering unigram entropy. Overall, nucleus sampling improves the Gen PPL-quality tradeoff, but does so by making generations less diverse according to unigram entropy.

\subsection{Using consistency loss on MDLM}
We next compare CoFRe against several MDLM-based baselines to isolate the effect of each component. In particular, we evaluate the original MDLM, MDLM trained with the same cross-step consistency regularization, MDLM with \citep{deschenaux2024beyond}, and our full CoFRe method. This comparison allows us to assess whether the gains come only from consistency regularization, from improved sampling, or from the full CoFRe formulation.

\FloatBarrier
\begin{table}[H]
\centering
\small
\caption{\textbf{Comparison of MDLM variants and CoFRe generation on OpenWebText.}
We report generative perplexity and unigram entropy across sampling budgets, comparing MDLM, MDLM with cross-step consistency regularization, MDLM with SDTT, and CoFRe, defined as FP-MDLM with cross-step consistency regularization and three-state reuse.}
\label{tab:mdlm_sdtt_cofre_comparison}
\setlength{\tabcolsep}{1.2pt}
\renewcommand{\arraystretch}{0.72}
\sisetup{detect-weight=true, detect-inline-weight=math}

\begin{tabular*}{\linewidth}{@{\extracolsep{\fill}} l l c c c c @{}}
\toprule
& & \multicolumn{4}{c}{\textbf{Budget}} \\
\cmidrule(l){3-6}
\textbf{Method} & \textbf{Metric} & \textbf{96} & \textbf{192} & \textbf{384} & \textbf{768}\\
\midrule

\multirow{2}{*}{MDLM}
& Gen PPL $\downarrow$ & 830.820 & 343.330 & 196.790 & 143.880 \\
& Entropy $\uparrow$   & 5.910 & 5.810 & 5.750 & 5.700 \\
\midrule

\multirow{2}{*}{\shortstack[l]{MDLM\\+ $\mathcal{L}_{\mathrm{CONS}}$}}
& Gen PPL $\downarrow$ & 621.121 & 250.313 & 141.397 & 102.731 \\
& Entropy $\uparrow$   & 5.840 & 5.757 & 5.690 & 5.636 \\
\midrule

\multirow{2}{*}{\shortstack[l]{MDLM\\+ SDTT}}
& Gen PPL $\downarrow$ & 193.050 & 89.170 & 62.290 & 47.040 \\
& Entropy $\uparrow$   & 5.580 & 5.530 & 5.490 & 5.450 \\
\midrule

\multirow{2}{*}{\shortstack[l]{CoFRe\\{\footnotesize(FP-MDLM + $\mathcal{L}_{\mathrm{CONS}}$ + 3SR)}}}
& Gen PPL $\downarrow$ & 101.791 & 65.182 & 48.755 & 37.846 \\
& Entropy $\uparrow$   & 5.434 & 5.380 & 5.283 & 5.142 \\

\bottomrule
\end{tabular*}
\end{table}
\FloatBarrier

Table~\ref{tab:mdlm_sdtt_cofre_comparison} shows that each additional modelling or sampling component improves Gen PPL over the MDLM baseline. Cross-step consistency regularization alone reduces MDLM Gen PPL by approximately $25.2\%$, $27.1\%$, $28.1\%$, and $28.6\%$ across budgets $96$, $192$, $384$, and $768$, respectively. SDTT yields substantially larger gains, reducing Gen PPL to $193.05$ at budget $96$ and $47.04$ at budget $768$. CoFRe performs best across all budgets, reaching $101.79$, $65.18$, $48.76$, and $37.85$ Gen PPL. Compared to MDLM, this corresponds to relative reductions of approximately $87.8\%$, $81.0\%$, $75.2\%$, and $73.7\%$. These gains come with a reduction in unigram entropy, from $5.91$-$5.70$ for MDLM to $5.43$-$5.14$ for CoFRe, highlighting the quality-diversity tradeoff induced by the proposed method.

\subsection{Comparison against other distilled or accelerated diffusion language models}
\label{appendix:comparison_more_methods}

We next compare CoFRe against other distilled or accelerated diffusion language models. In particular, we evaluate PGM 6/6, PGM 6/6 with SDTT \citep{deschenauxPartitionGenerativeModeling2026}, IDLM-MDLM \citep{liIDLMInverseDistilled2026}, and our full CoFRe method. For PGM and IDLM-MDLM, we report the effective sampling budget as $12 \times$ the number of sampling steps. We emphasize that CoFRe is not primarily designed as a distillation or acceleration method: rather, its goal is to provide a general generation framework that improves sample quality while reducing the effective cost of training and the cost of generation, especially in low-budget regimes. This comparison allows us to assess how CoFRe compares not only to MDLM-based variants, but also to methods that explicitly target faster sampling or improved diffusion language model generation.

\FloatBarrier
\begin{figure}[H]
\centering
\includegraphics[width=0.72\linewidth]{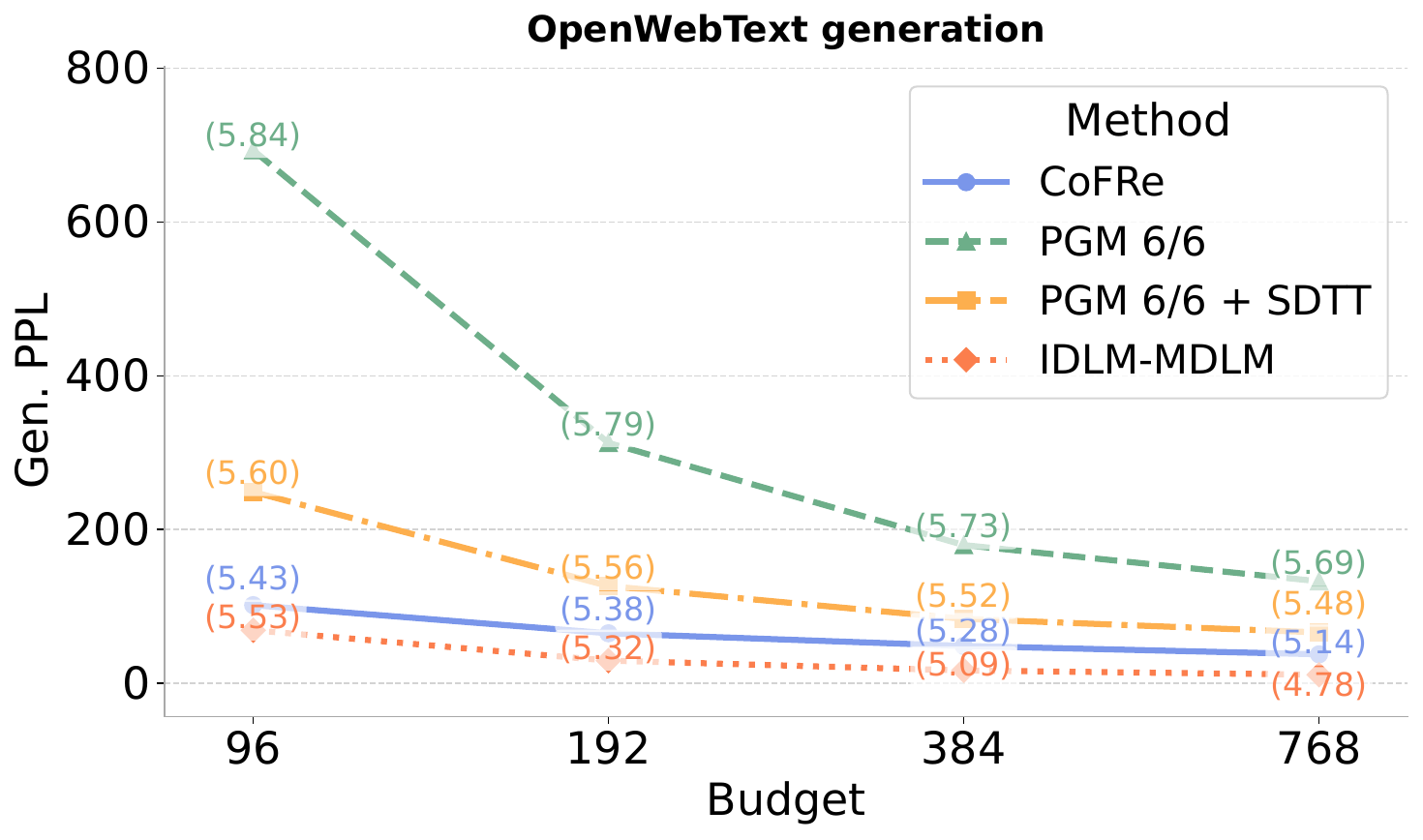}
\caption{\textbf{Generative perplexity as a function of sampling budget on OpenWebText.}
We compare CoFRe against PGM 6/6, PGM 6/6 with SDTT, and IDLM-MDLM.}
\label{fig:gen_ppl_budget_comparison}
\end{figure}
\FloatBarrier

\FloatBarrier
\begin{table}[H]
\centering
\small
\caption{\textbf{Comparison of CoFRe, PGM, and IDLM-MDLM on OpenWebText.}
We report generative perplexity and entropy across sampling budgets. For PGM and IDLM-MDLM,
the budget is computed as $12 \times$ the number of sampling steps.}
\label{tab:cofre_pgm_idlm_budget_comparison}
\setlength{\tabcolsep}{2.5pt}
\renewcommand{\arraystretch}{0.78}
\sisetup{detect-weight=true, detect-inline-weight=math}

\begin{tabular*}{\linewidth}{@{\extracolsep{\fill}} l l c c c c @{}}
\toprule
& & \multicolumn{4}{c}{\textbf{Budget}} \\
\cmidrule(l){3-6}
\textbf{Method} & \textbf{Metric} & \textbf{96} & \textbf{192} & \textbf{384} & \textbf{768} \\
\midrule

\multirow{2}{*}{\shortstack[l]{CoFRe\\{\scriptsize FP-MDLM + $\mathcal{L}_{\mathrm{CONS}}$ + 3SR}}}
& Gen PPL $\downarrow$ & 101.791 & 65.182 & 48.755 & 37.846 \\
& Entropy $\uparrow$   & 5.434 & 5.380 & 5.283 & 5.142 \\
\midrule

\multirow{2}{*}{PGM 6/6}
& Gen PPL $\downarrow$ & 693.513 & 312.812 & 179.928 & 132.786 \\
& Entropy $\uparrow$   & 5.843 & 5.785 & 5.732 & 5.688 \\
\cmidrule(lr){1-6}

\multirow{2}{*}{PGM 6/6 + SDTT}
& Gen PPL $\downarrow$ & 249.076 & 126.604 & 83.952 & 66.316 \\
& Entropy $\uparrow$   & 5.599 & 5.561 & 5.524 & 5.484 \\
\cmidrule(lr){1-6}

\multirow{2}{*}{IDLM-MDLM}
& Gen PPL $\downarrow$ & 69.165 & 29.659 & 16.843 & 11.312 \\
& Entropy $\uparrow$   & 5.529 & 5.323 & 5.089 & 4.782 \\

\bottomrule
\end{tabular*}
\end{table}
\FloatBarrier

Table~\ref{tab:cofre_pgm_idlm_budget_comparison} and Figure~\ref{fig:gen_ppl_budget_comparison} show that CoFRe substantially improves over the PGM baselines across all sampling budgets. Compared to PGM 6/6, CoFRe reduces Gen PPL across all budgets. Applying SDTT to PGM gives a strong improvement over the undistilled PGM baseline, reducing Gen PPL from $693.51$ to $249.08$ at budget $96$ and from $132.79$ to $66.32$ at budget $768$. Nevertheless, CoFRe remains better than PGM 6/6 + SDTT at every budget, with relative reductions of approximately $59.1\%$, $48.5\%$, $41.9\%$, and $42.9\%$.

IDLM-MDLM achieves the lowest Gen PPL across all budgets, reaching $69.17$, $29.66$, $16.84$, and $11.31$. Compared to CoFRe, this corresponds to additional reductions of approximately $32.1\%$, $54.5\%$, $65.5\%$, and $70.1\%$. However, these improvements are accompanied by a stronger reduction in unigram entropy at larger budgets: IDLM-MDLM entropy decreases from $5.53$ at budget $96$ to $4.78$ at budget $768$, whereas CoFRe remains between $5.43$ and $5.14$. Overall, CoFRe provides a strong improvement over accelerated PGM variants while preserving higher entropy than IDLM-MDLM at medium and large budgets, again highlighting the quality-diversity tradeoff among accelerated diffusion language model samplers.

\subsection{Extended Image Modeling Results}
\label{appendix:image_full_results}

\subsubsection{Samples}
\FloatBarrier
\begin{figure}[H]
    \centering
    \includegraphics[width=0.155\linewidth]{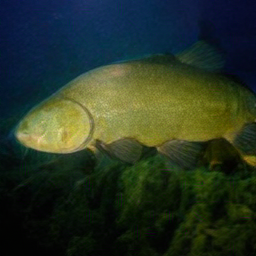}\hfill
    \includegraphics[width=0.155\linewidth]{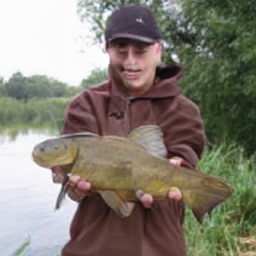}\hfill
    \includegraphics[width=0.155\linewidth]{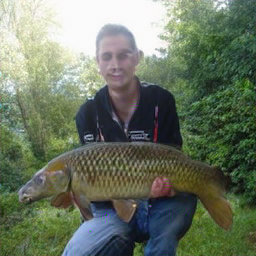}\hfill
    \includegraphics[width=0.155\linewidth]{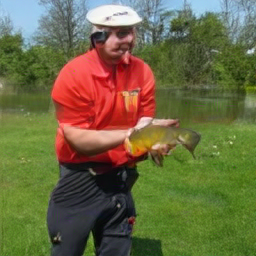}\hfill
    \includegraphics[width=0.155\linewidth]{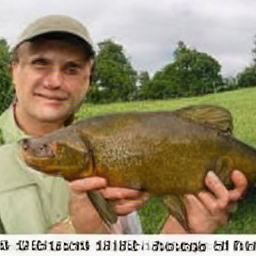}\hfill
    \includegraphics[width=0.155\linewidth]{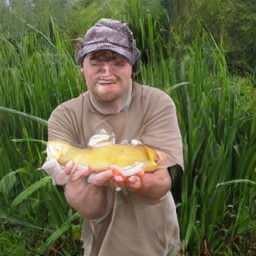}
    \caption{Generated samples using CoFRe with a budget of 460.}
    \label{fig:generated_samples}
\end{figure}
\FloatBarrier
\subsubsection{MaskGIT-Large vs MaskGIT-12 vs FP-MaskGIT}

\FloatBarrier
\begin{table}[H]
\centering
\small
\caption{Generation quality vs.\ compute budget, where the budget counts the total number of transformer-block forward passes. Training time and training VRAM are reported in the column headers; percentages indicate reductions relative to MaskGIT-Large. Models are trained on Imagenette (10-class subset of ImageNet) for 50k training steps. MaskGIT-Large uses 24 transformer layers, MaskGIT-12 uses 12 transformer layers, and FP-MaskGIT uses 4 pre- and 4 post-transformer layers. All models use a global batch size (GBS) of 128. For 24-layer MaskGIT-Large, a budget of 48 corresponds to 2 decoding steps ($2\times24$), while for 12-layer MaskGIT-12 it corresponds to 4 decoding steps ($4\times12$).}
\label{tab:maskgit_budget_grouped}
\setlength{\tabcolsep}{1.2pt}
\renewcommand{\arraystretch}{0.72}
\sisetup{detect-weight=true, detect-inline-weight=math}

\begin{tabular*}{\linewidth}{@{\extracolsep{\fill}} r c c c c c c @{}}
\toprule
\textbf{Budget}
& \multicolumn{2}{c}{\shortstack{\textbf{MaskGIT-Large}\\\scriptsize 24 transformer layers\\\scriptsize Train: 17h\,46m \quad VRAM: 72.45\,GiB}}
& \multicolumn{2}{c}{\shortstack{\textbf{FP-MaskGIT}\\\scriptsize 4 pre / 4 post layers\\\scriptsize Train: 9h\,08m (-48.6\%)\\\scriptsize VRAM: 35.74\,GiB (-50.7\%)}}
& \multicolumn{2}{c}{\shortstack{\textbf{MaskGIT-12}\\\scriptsize 12 transformer layers\\\scriptsize Train: 7h\,40m (-56.8\%)\\\scriptsize VRAM: 28.14\,GiB (-61.2\%)}} \\
\cmidrule(lr){2-3}
\cmidrule(lr){4-5}
\cmidrule(lr){6-7}
& \textbf{FID} $\downarrow$
& \textbf{IS} $\uparrow$
& \textbf{FID} $\downarrow$
& \textbf{IS} $\uparrow$
& \textbf{FID} $\downarrow$
& \textbf{IS} $\uparrow$ \\
\midrule

48
& 174.0856
& 9.2860
& \textbf{100.0964}
& \textbf{15.0196}
& 128.9871
& 12.6302 \\

96
& 117.6439
& 13.3696
& \textbf{57.5823}
& 15.8495
& 77.0653
& \textbf{15.9587} \\

192
& 54.6172
& \textbf{16.0220}
& \textbf{32.0072}
& 15.0822
& 50.9830
& 15.3803 \\

240
& 44.4930
& \textbf{15.6458}
& \textbf{27.6470}
& 14.6582
& 45.6384
& 15.2925 \\

384
& 30.0202
& 14.6473
& \textbf{24.1946}
& 14.4567
& 38.0946
& \textbf{15.2823} \\

480
& 27.3975
& \textbf{14.2319}
& \textbf{22.1171}
& 13.8718
& 36.0931
& 15.2505 \\

\bottomrule
\end{tabular*}
\end{table}
\FloatBarrier

\subsubsection{Generation quality when adding different components of CoFRe}

\FloatBarrier
\begin{table}[H]
\centering
\small
\caption{\textbf{Component ablation for FP-MaskGIT generation on ImageNette.}
We report FID and Inception Score across sampling budgets. The table compares MaskGIT baselines, the FP-MaskGIT backbone with different reuse variants, and FP-MaskGIT with cross-step consistency regularization. CoFRe corresponds to the consistency-trained FP-MaskGIT model with reuse.}
\label{tab:fp_maskgit_component_ablation}
\setlength{\tabcolsep}{1.2pt}

\sisetup{detect-weight=true, detect-inline-weight=math}

\begin{tabular*}{\linewidth}{@{\extracolsep{\fill}} l l c c c c @{}}
\toprule
& & \multicolumn{4}{c}{\textbf{Budget}} \\
\cmidrule(l){3-6}
\textbf{Method} & \textbf{Metric} & \textbf{48} & \textbf{96} & \textbf{192} & \textbf{384}\\
\midrule

\multirow{2}{*}{MaskGIT-Large}
& FID $\downarrow$ & 174.0856 & 117.6439 & 54.6172 & 30.0202 \\
& IS $\uparrow$   & 9.2860 & 13.3696 & 16.0220 & 14.6473 \\
\midrule

\multirow{2}{*}{MaskGIT-12}
& FID $\downarrow$ & 128.9871 & 77.0653 & 50.9830 & 38.0946 \\
& IS $\uparrow$   & 12.6302 & 15.9587 & 15.3803 & 15.2823 \\
\midrule

\multirow{2}{*}{\shortstack[l]{FP-MaskGIT\\no reuse}}
& FID $\downarrow$ & 100.0964 & 57.5823 & 32.0072 & 24.1946 \\
& IS $\uparrow$   & 15.0196 & 15.8495 & 15.0822 & 14.4567 \\
\cmidrule(lr){1-6}

\multirow{2}{*}{\shortstack[l]{FP-MaskGIT\\+ reuse}}
& FID $\downarrow$ & 101.5056 & 55.2582 & 31.7141 & 24.0396 \\
& IS $\uparrow$   & 14.4204 & 15.5139 & 14.5876 & 	13.9587 \\
\cmidrule(lr){1-6}

\multirow{2}{*}{\shortstack[l]{FP-MaskGIT\\+ 3SR}}
& FID $\downarrow$ & 102.3251 & 54.4212 & 31.2308 & 23.5458 \\
& IS $\uparrow$   & 14.432 & 15.4369 & 14.912 & 14.0228 \\
\midrule

\multirow{2}{*}{\shortstack[l]{FP-MaskGIT\\+ $\mathcal{L}_{\mathrm{CONS}}$+ no reuse}}
& FID $\downarrow$ & 107.5012 & 56.7336 & 31.3828 & 24.6839 \\
& IS $\uparrow$   & 13.3413 & 16.3229 & 15.0487 & 14.0886 \\
\cmidrule(lr){1-6}

\multirow{2}{*}{\shortstack[l]{FP-MaskGIT\\+ $\mathcal{L}_{\mathrm{CONS}}$+ reuse}}
& FID $\downarrow$ & 97.8401 & 54.5651 & 30.0072 & 23.4257 \\
& IS $\uparrow$   & 15.0866 & 16.2692 & 14.5894 & 14.2665 \\
\cmidrule(lr){1-6}

\multirow{2}{*}{\shortstack[l]{FP-MaskGIT\\+ $\mathcal{L}_{\mathrm{CONS}}$+ reuse, CoFRe}}
& FID $\downarrow$ & 96.7331 & 51.0077 & 27.6242 & 22.8381 \\
& IS $\uparrow$   & 14.4074 & 15.9572 & 15.0822 & 14.4567 \\

\bottomrule
\end{tabular*}
\end{table}
\FloatBarrier

\subsection{Component Ablation for FP-MGM}
\label{appendix:component_ablation}

We use small-scale proxy runs to select the main FP-MGM design choices before running the full experiments. For language, we ablate the number of explicit preprocessing/postprocessing layers, the number of fixed-point iterations used with and without gradients, the learning rate, and gradient clipping. Unless stated otherwise, these ablations are run for 100k training steps and are intended to compare configurations under the same compute setting, not to match the final full-scale results.

\paragraph{Language-modeling ablations on LM1B.}
Table~\ref{tab:lm1b_fp_ablation} reports FP-MDLM ablations on LM1B without sentence packing. The main trend is that very shallow stochastic fixed-point training is cheapest, while larger stochastic solver budgets improve perplexity at moderate cost. Using a deterministic large solver budget, $\mathcal{U}(12,12)$ without gradients and $\mathcal{U}(12,12)$ with gradients, gives the best validation perplexity among FP-MDLM variants, but it is substantially slower and more memory-intensive. The stochastic setting $\mathcal{U}(0,4)$ without gradients and $\mathcal{U}(3,6)$ with gradients gives the best FP-MDLM test perplexity, and provides a better cost--quality trade-off. We therefore use this stochastic solver setting as the default for the larger FP-MDLM runs.

\FloatBarrier
\begin{table}[H]
\centering
\small
\caption{\textbf{FP-MDLM solver and architecture ablation on LM1B.}
Models are trained for 100k steps without sentence packing. $\mathcal{U}(a,b)$ denotes a discrete uniform distribution over solver iterations. We report validation perplexity and test perplexity on 500 samples.}
\label{tab:lm1b_fp_ablation}
\setlength{\tabcolsep}{0.8pt}
\renewcommand{\arraystretch}{0.92}
\sisetup{detect-weight=true, detect-inline-weight=math}

\begin{tabular*}{\linewidth}{@{\extracolsep{\fill}} l c c c c c c c c c @{}}
\toprule
\textbf{Model}
& \textbf{Pre/Post}
& \textbf{No-grad}
& \textbf{Grad}
& \textbf{LR}
& \textbf{Clip}
& \textbf{Time}
& \textbf{VRAM}
& \textbf{Val. PPL $\downarrow$}
& \textbf{Test PPL $\downarrow$} \\
\midrule

FP-MDLM & 2/2 & $\mathcal{U}(0,2)$  & $\mathcal{U}(1,3)$  & $1{\times}10^{-4}$ & 0.5 & 8h05 & 36.0GB  & 50.66  & 48.057 \\
FP-MDLM & 2/2 & $\mathcal{U}(0,4)$  & $\mathcal{U}(3,6)$  & $1{\times}10^{-4}$ & 0.5 & 9h53 & 41.0GB  & 46.54  & \textbf{44.885} \\
FP-MDLM & 2/2 & $\mathcal{U}(12,12)$ & $\mathcal{U}(12,12)$ & $1{\times}10^{-4}$ & 0.5 & 16h49 & 56.2GB & 42.50  & 45.136 \\
FP-MDLM & 1/1 & $\mathcal{U}(0,2)$  & $\mathcal{U}(1,3)$  & $1{\times}10^{-4}$ & 0.5 & \textbf{7h05} & \textbf{30.7GB} & 47.37 & 50.740 \\
FP-MDLM & 1/1 & $\mathcal{U}(0,4)$  & $\mathcal{U}(3,6)$  & $1{\times}10^{-4}$ & 0.5 & 8h53 & 36.18GB & 44.60  & 46.620 \\
FP-MDLM & 1/1 & $\mathcal{U}(0,3)$  & $\mathcal{U}(2,4)$  & $1{\times}10^{-4}$ & 0.5 & 7h52 & 34.4GB  & 49.19  & 47.690 \\
FP-MDLM & 1/1 & $\mathcal{U}(0,4)$  & $\mathcal{U}(3,6)$  & $2{\times}10^{-4}$ & 0.5 & 8h51 & 36.18GB & 44.06  & 45.926 \\
FP-MDLM & 1/1 & $\mathcal{U}(0,4)$  & $\mathcal{U}(3,6)$  & $1{\times}10^{-4}$ & 1.0 & 8h50 & 36.18GB & 44.60  & 46.773 \\
FP-MDLM & 1/1 & $\mathcal{U}(0,4)$  & $\mathcal{U}(3,6)$  & $2{\times}10^{-4}$ & 1.0 & 8h50 & 36.18GB & 42.795 & 45.175 \\
\midrule
MDLM DiT & -- & -- & -- & $3{\times}10^{-4}$ & -- & 11h02 & 44.0GB & \textbf{42.24} & 48.522 \\

\bottomrule
\end{tabular*}
\end{table}
\FloatBarrier

\paragraph{Language-modeling ablations on OWT.}
Table~\ref{tab:owt128_fp_ablation} shows the corresponding proxy experiment on OWT with sequence length 128. The FP-MDLM variants use substantially fewer parameters than the MDLM DiT baseline. Among the FP-MDLM variants, configuration B, with learning rate $2{\times}10^{-4}$ and gradient clipping 1.0, gives the best validation and test perplexity. It also improves generative perplexity over configuration A across all reported sampling budgets. MDLM remains stronger in likelihood at this small scale, but FP-MDLM is competitive in low-budget generation and uses fewer parameters. For FP-MDLM, we use $K_{\mathrm{pre}}=1$, $K_{\mathrm{fp}}=1$, and $K_{\mathrm{post}}=1$.

\FloatBarrier
\begin{table}[H]
\centering
\tiny
\caption{\textbf{FP-MDLM proxy ablation on OWT with sequence length 128.}
Models are trained for 100k steps. Generation cells report generative perplexity / unigram entropy.}
\label{tab:owt128_fp_ablation}
\setlength{\tabcolsep}{0.8pt}
\renewcommand{\arraystretch}{0.72}
\sisetup{detect-weight=true, detect-inline-weight=math}

\begin{tabular*}{\linewidth}{@{\extracolsep{\fill}} l c c c c c c c c c c c @{}}
\toprule
\textbf{Model}
& \textbf{Pre/Post}
& \textbf{LR}
& \textbf{Clip}
& \textbf{Params}
& \textbf{Val. PPL $\downarrow$}
& \textbf{Test PPL $\downarrow$}
& \textbf{B128}
& \textbf{B64}
& \textbf{B32}
& \textbf{B16}
& \textbf{B8} \\
\midrule

FP-MDLM A
& 1/1
& $1{\times}10^{-4}$
& 0.5
& 104M
& 58.98
& 55.390
& 185.42 / 4.32
& 216.24 / 4.33
& 263.59 / 4.34
& 400.29 / 4.38
& 744.83 / 4.41 \\

FP-MDLM B
& 1/1
& $2{\times}10^{-4}$
& 1.0
& 104M
& 53.278
& 50.287
& 170.79 / 4.31
& 201.23 / 4.32
& 230.00 / 4.34
& 380.00 / 4.38
& 637.00 / 4.41 \\

\midrule

MDLM DiT
& --
& $3{\times}10^{-4}$
& 1.0
& 169M
& \textbf{47.10}
& \textbf{45.211}
& \textbf{133.15} / 4.29
& \textbf{150.62} / 4.31
& \textbf{193.946} / 4.33
& \textbf{285.50} / 4.36
& \textbf{552.71} / 4.38 \\

\bottomrule
\end{tabular*}
\end{table}
\FloatBarrier

\paragraph{Image-generation ablation.}
For image generation, we use the same fixed-point replacement idea inside MaskGIT and compare the resulting FP-MaskGIT to fixed-depth MaskGIT baselines. Table~\ref{tab:imagenette_arch_ablation} reports the completed ImageNette runs. FP-MaskGIT with a 4/4 explicit pre/post architecture substantially reduces training cost relative to MaskGIT-Large, from 17h46m to 9h08m and from 72.45GiB to 35.74GiB. It also improves FID at all reported budgets. Compared with the smaller 12-layer MaskGIT baseline, FP-MaskGIT is more expensive to train but gives better FID across the reported budgets, supporting the choice of the 4/4 FP-MaskGIT configuration for the main image experiments. For FP-MaskGIT, we use $K_{\mathrm{pre}}=4$, $K_{\mathrm{fp}}=1$, and $K_{\mathrm{post}}=4$.

\FloatBarrier
\begin{table}[H]
\centering
\small
\caption{\textbf{Architecture ablation for FP-MaskGIT on ImageNette.}
We compare MaskGIT baselines against FP-MaskGIT with different numbers of explicit preprocessing and postprocessing layers. Smaller FP-MaskGIT variants perform progressively worse than the 4/4 configuration. For each model, we report FID and IS across sampling budgets.}
\label{tab:imagenette_arch_ablation}
\setlength{\tabcolsep}{1.0pt}
\renewcommand{\arraystretch}{0.72}
\sisetup{detect-weight=true, detect-inline-weight=math}

\begin{tabular*}{\linewidth}{@{\extracolsep{\fill}} l l c c c c c c @{}}
\toprule
\textbf{Model} & \textbf{Metric} & \textbf{48} & \textbf{96} & \textbf{192} & \textbf{240} & \textbf{384} & \textbf{480} \\
\midrule

\multirow{2}{*}{MaskGIT-Large}
& FID $\downarrow$ & 174.0856 & 117.6439 & 54.6172 & 44.4930 & 30.0202 & 27.3975 \\
& IS  $\uparrow$   & 9.2860   & 13.3696  & \textbf{16.0220} & \textbf{15.6458} & 14.6473 & \textbf{14.2319} \\
\midrule

\multirow{2}{*}{MaskGIT-12}
& FID $\downarrow$ & 128.9871 & 77.0653 & 50.9830 & 45.6384 & 38.0946 & 32.8731 \\
& IS  $\uparrow$   & 12.6302  & \textbf{15.9587} & 15.3803 & 15.2925 & \textbf{15.2823} & 14.1865 \\
\midrule

\multirow{2}{*}{FP-MaskGIT 1/1}
& FID $\downarrow$ & 121.7644 & 72.1945 & 46.2390 & 41.1405 & 34.6196 & 30.1841 \\
& IS  $\uparrow$   & 12.7667 & 13.4721 & 12.8199 & 12.4595 & 12.2882 & 11.7910 \\
\midrule

\multirow{2}{*}{FP-MaskGIT 2/2}
& FID $\downarrow$ & 114.5418 & 67.3238 & 41.4951 & 36.6427 & 31.1446 & 27.4951 \\
& IS  $\uparrow$   & 13.5176 & 14.2646 & 13.5740 & 13.1924 & 13.0110 & 12.4846 \\
\midrule

\multirow{2}{*}{FP-MaskGIT 3/3}
& FID $\downarrow$ & 107.3191 & 62.4530 & 36.7511 & 32.1448 & 27.6696 & 24.8061 \\
& IS  $\uparrow$   & 14.2686 & 15.0570 & 14.3281 & 13.9253 & 13.7339 & 13.1782 \\
\midrule

\multirow{2}{*}{FP-MaskGIT 4/4}
& FID $\downarrow$ & \textbf{100.0964} & \textbf{57.5823} & \textbf{32.0072} & \textbf{27.6470} & \textbf{24.1946} & \textbf{22.1171} \\
& IS  $\uparrow$   & \textbf{15.0196} & 15.8495 & 15.0822 & 14.6582 & 14.4567 & 13.8718 \\
\bottomrule
\end{tabular*}
\end{table}
\FloatBarrier

Overall, these ablations motivate the default FP-MGM recipe used in the main experiments: a stochastic fixed-point solver rather than a fully deterministic deep solve, a compact explicit pre/post architecture for language, and a 4/4 explicit pre/post architecture for image generation. The proxy results also show the main trade-off that carries through the full experiments: fixed-point denoisers reduce parameter and memory cost, but require careful solver-budget and reuse choices to obtain strong generation quality.

\subsection{Cross-step consistency recovers low-budget generation quality}
\label{appendix:consistency-reg}

In masked generative models, the loss is typically applied only to final predictions at masked positions, so intermediate states are supervised only indirectly. A common extension is \emph{consistency} regularization, which aligns predictions or hidden states across nearby denoising steps. This can be done in output space, for example with KL on logits, or in representation space, for example with MSE or cosine similarity on hidden states. Targets may come from the same model with stop-gradient or from a teacher such as an EMA copy. This adds auxiliary supervision between denoising states while leaving the original masked modeling objective unchanged.

A related idea is self-distillation through time (SDTT), which trains a student with fewer denoising steps to match a teacher with more steps. If $p_\theta^{(m)}$ is generation with $m$ steps and $p_\nu^{(k)}$ the student model with $k<m$, SDTT minimizes
$ \mathbb{E}_{\mathbf{z}_0 \sim \mathcal{D},\, \mathbf{z}_t \sim q_t(\mathbf{z}_t \mid \mathbf{z}_0)} \, \delta\!\left(\mathbf{x}_\nu(\mathbf{z}_t, t), \tilde{\mathbf{x}}^{\text{teacher}}_\theta(\mathbf{z}_t, t, m/k)\right), $
where $\delta$ is typically a divergence such as KL, $\operatorname{sg}$ denotes stop-gradient, and $\tilde{\mathbf{x}}^{\text{teacher}}_\theta$ aggregates teacher predictions over multiple steps. After training, one student step can approximate several teacher steps. Consistency regularization thus provides local temporal supervision, whereas SDTT directly compresses the denoising process.

\paragraph{Experimental Settings}
For the cross-step consistency regularization experiments, we compare two FP-MDLM models with the same 1M-step base checkpoint, followed by a 30k consistency post-training stage. The baseline is trained from scratch using only the MDLM loss. The consistency model starts from the same training setup, but adds a consistency objective late in training. We explore several consistency objectives and found the best results with  $\mathcal{L}_{\mathrm{MSE}}$ introduced Sec. \ref{method:consistency}; the comparison is reported in Appendix \ref{appendix:loss-type-consistency}. Specifically, the consistency weight is introduced at step 1M, increased linearly from 0 to 0.1 over 5k steps, and then kept at 0.1 until step $1030000$. We evaluate both models using GPT-2 Large generative perplexity and sample entropy, and report results both with and without solution reuse.

\paragraph{Results}
Table~\ref{tab:fp_mdlm_component_ablation} shows that cross-step consistency regularization is the main driver of generation quality. Without reuse, adding $\mathcal{L}_{\mathrm{CONS}}$ improves generative perplexity at every budget, from 375.6 to 104.2 at budget 96 and from 179.7 to 41.7 at budget 768. This is a large improvement over the baseline FP-MDLM, while maintaining non-degenerate entropy across all budgets.

Reuse becomes useful once the model has been regularized with $\mathcal{L}_{\mathrm{CONS}}$. In the baseline checkpoint, reuse and 3SR are inconsistent and can hurt generation quality, especially at the largest budget. After consistency regularization, however, reuse further improves the low-perplexity model at budgets 192, 384, and 768. The best results are obtained with 3SR at budgets 96, 192, and 384, reaching generative perplexities of 101.8, 65.2, and 48.8, while standard reuse is slightly better at budget 768 with 37.6. \textbf{Overall, $\mathcal{L}_{\mathrm{CONS}}$ turns FP-MDLM into a much stronger generator, and reuse provides additional gains once the fixed-point denoiser has been regularized.}

\paragraph{Lagged logit analysis.}
To understand the improvement from $\mathcal{L}_{\mathrm{CONS}}$, we compare masked-token logits at a student denoising step $s$ with logits at cleaner future steps $s+\ell$ on shared contexts. The consistency-trained model has lower lagged logit KL than the baseline across all sampling-step, solver-budget, and lag settings, reducing the KL by $15.2\%$ on average. The effect is strongest for nearby denoising states, with a $19.0\%$ average reduction at lag $1$, and remains positive through lag $4$. This suggests that $\mathcal{L}_{\mathrm{CONS}}$ acts as a cross-time self-distillation signal: it makes student-step logits more aligned with cleaner future predictions, which helps explain the large generative perplexity gains in Tables~\ref{tab:fp_mdlm_component_ablation} and \ref{tab:fp_maskgit_component_ablation}.
\FloatBarrier
\begin{figure}[H]
    \centering
    \begin{subfigure}[b]{0.48\textwidth}
        \centering
        \includegraphics[width=\textwidth]{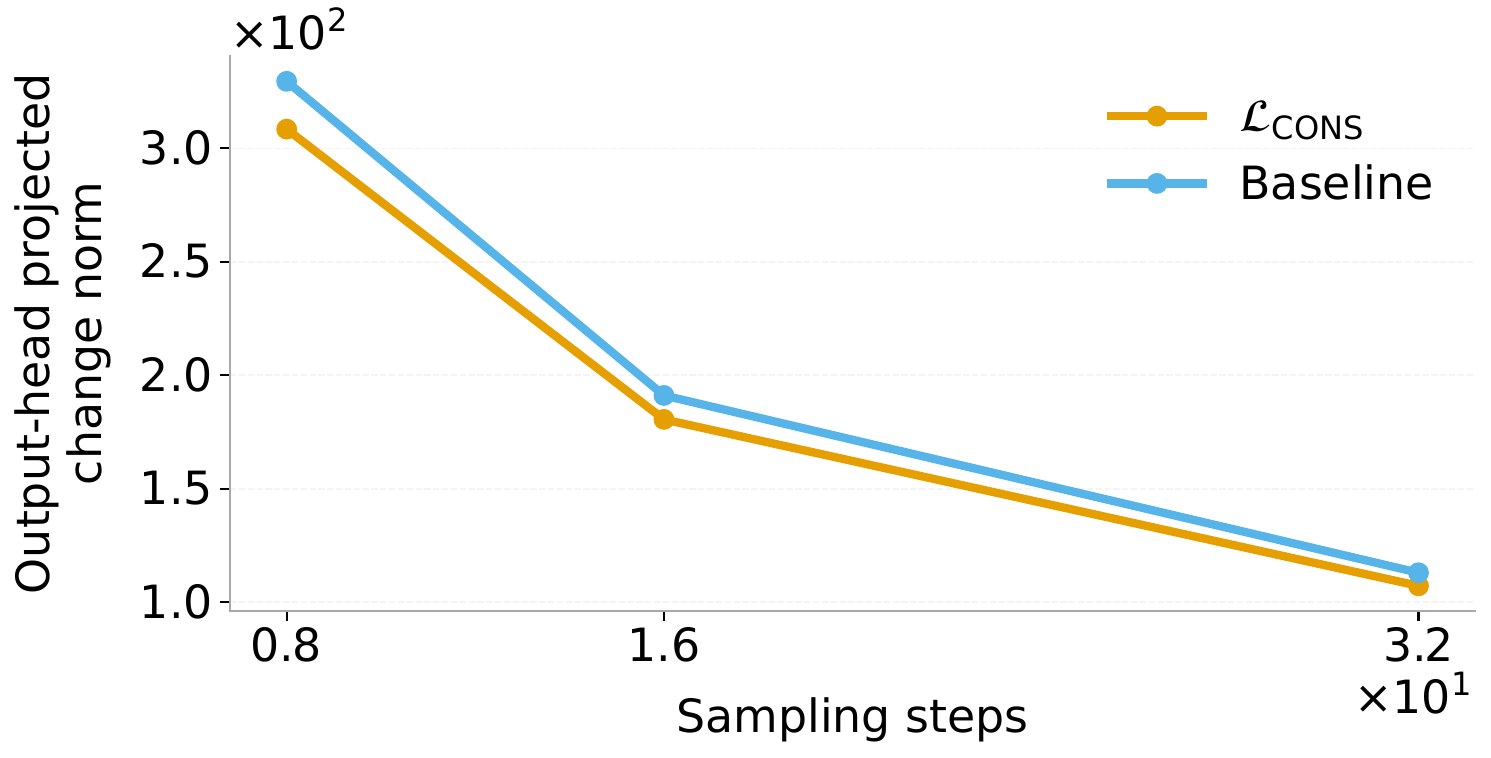}

        \label{fig:lagged_l}
    \end{subfigure}
    \hfill 
    \begin{subfigure}[b]{0.48\textwidth}
        \centering
        \includegraphics[width=\textwidth]{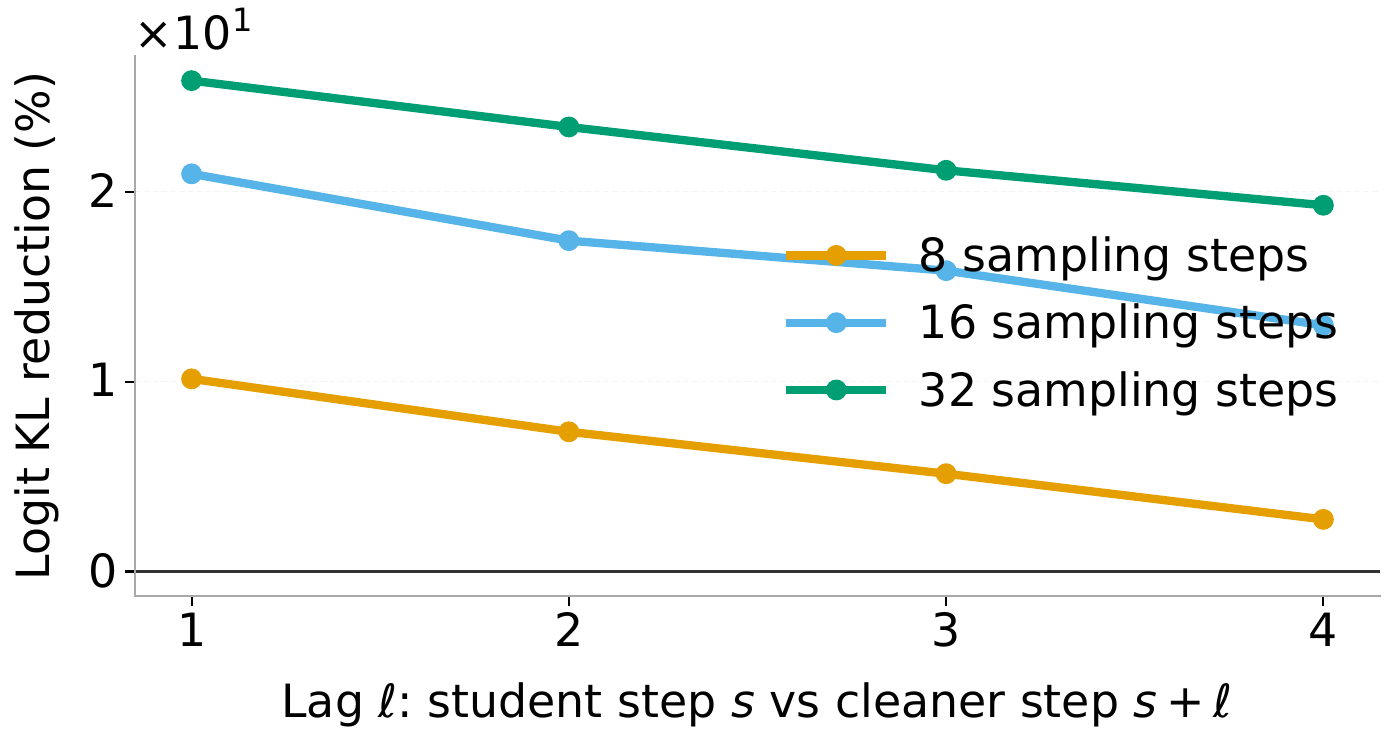}

        \label{fig:lagged_r}
    \end{subfigure}
    
    \caption{\textbf{Lagged logit analysis.}
(Left) output-head-projected hidden-state changes decrease as the number of sampling steps increases, for both the baseline and the $\mathcal{L}_{\mathrm{CONS}}$ model. (Right) relative reduction in lagged logit KL from $\mathcal{L}_{\mathrm{CONS}}$ compared to the baseline, measured between a student denoising step $s$ and a cleaner future step $s+\ell$. The consistency-trained model reduces lagged logit KL across lags and sampling-step settings, with the strongest gains at smaller lags.}
\label{fig:lagged_logit_analysis}
    \label{fig:lagged_consistency}
\end{figure}

\FloatBarrier

\subsection{Consistency Loss Training Dynamics}
\label{appendix:lcons_training_dynamics}

In practice, we find that extending the consistency stage for too long can degrade generation quality by over-sharpening the model. We therefore use validation perplexity as an early stopping signal: starting from the pre-$\mathcal{L}_{\mathrm{CONS}}$ checkpoint, we select the first checkpoint whose validation perplexity exceeds the pre-$\mathcal{L}_{\mathrm{CONS}}$ value by 15\%. This rule gave the best empirical trade-off between generative perplexity and entropy. We linearly warm up the consistency term at the beginning of post-training to avoid an abrupt change in the optimization objective.

\begin{figure}[H]
    \centering
    \begin{subfigure}[b]{0.46\textwidth}
        \centering
        \includegraphics[width=0.9\textwidth]{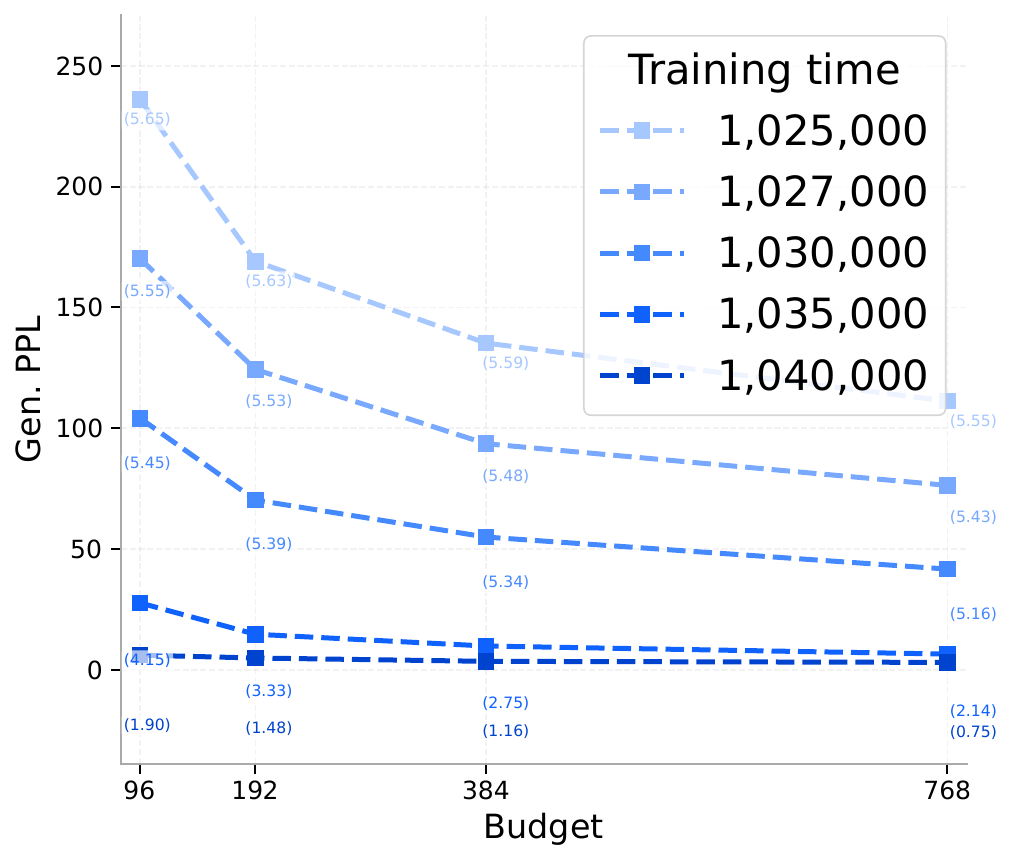}

        \label{fig:model_collapse_gen_ppl}
    \end{subfigure}
    \hfill 
    \begin{subfigure}[b]{0.52\textwidth}
        \centering
        \includegraphics[width=\textwidth]{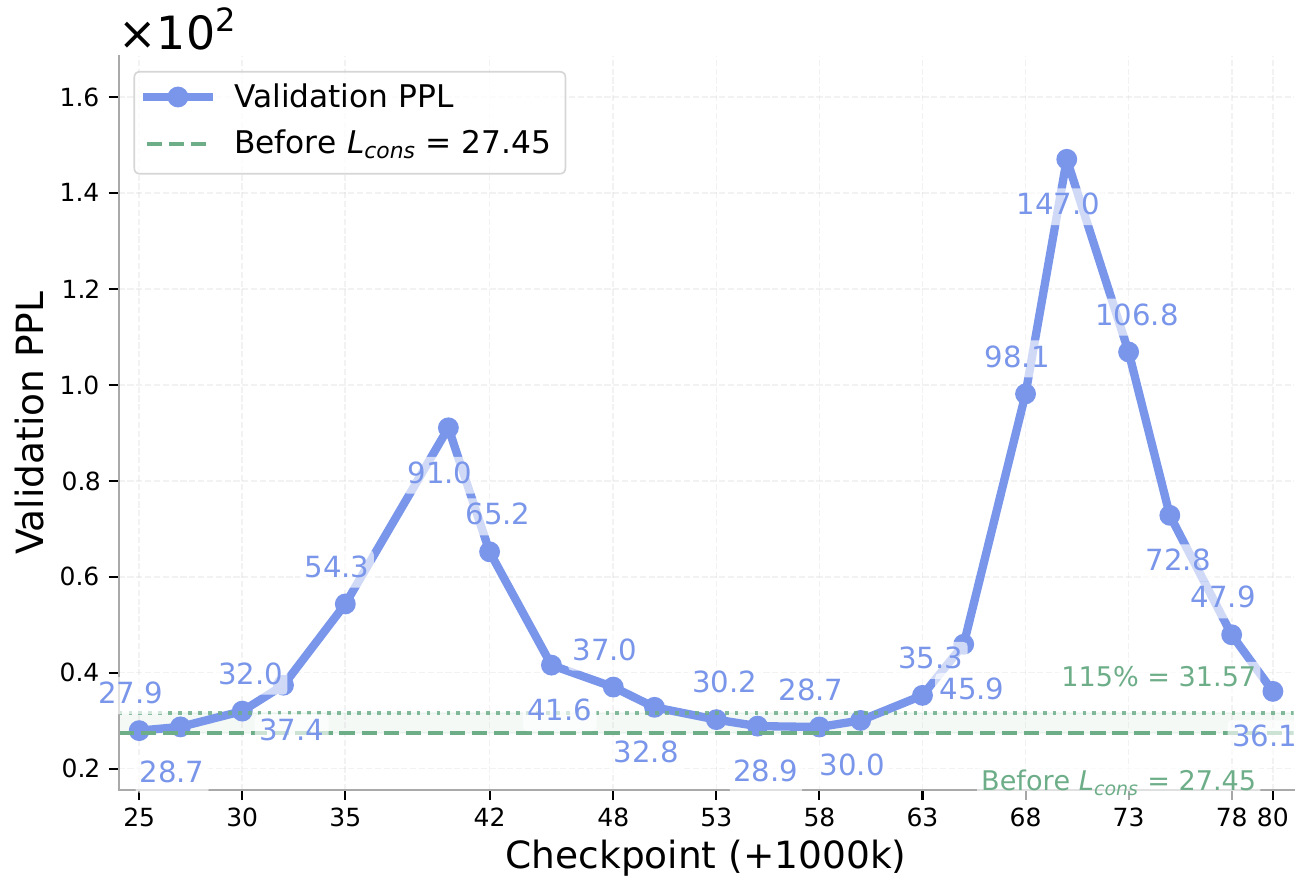}

        \label{fig:model_collapse_ppl}
    \end{subfigure}
    
    \caption{\textbf{Checkpoint selection for the $\mathcal{L}_{\mathrm{CONS}}$ post-training stage.}
(Left) Generative perplexity across budgets improves rapidly during early consistency training, but later checkpoints over-sharpen the model, as reflected by collapsing entropy values shown in parentheses. Sampling is done without warm-start (e.g. no reuse)
(Right) Validation perplexity is not monotonic: it first rises above the 15\% threshold, then can decrease again at later checkpoints. Since these later decreases do not recover sample diversity, we use the first checkpoint whose validation perplexity exceeds the pre-$\mathcal{L}_{\mathrm{CONS}}$ value by 15\% as our stopping rule. This empirically balances generation quality and sample diversity.}
    \label{fig:model_collapse}
\end{figure}

Figure~\ref{fig:model_collapse} illustrates this behavior. Generative perplexity improves rapidly during early consistency training, but later checkpoints become increasingly over-sharpened, as shown by the sharp drop in sample entropy. The validation perplexity curve is not monotonic: after first increasing beyond the 15\% threshold, it can later decrease again. However, these later decreases do not correspond to recovered sample diversity, so selecting a later checkpoint based only on validation perplexity would be misleading. We therefore use the first threshold crossing as a conservative stopping rule. This is qualitatively similar to the caution in SDTT \citep{deschenaux2024beyond}, where repeated distillation rounds can accumulate approximation error because each student becomes the next teacher; in both cases, validation metrics and sample diversity should be monitored rather than relying only on the distillation loss.

\FloatBarrier
\FloatBarrier

\subsection{Ablations on the Consistency Loss Type}
\label{appendix:loss-type-consistency}

We ablate the form of the consistency objective for FP-MaskGIT on ImageNette. Starting from the same FP-MaskGIT checkpoint, we compare output-space consistency with KL on logits, representation-space consistency with MSE or cosine distance, and variants that use an EMA teacher. We also test whether adding this consistency loss at the beginning of the training improves the final model. For each loss, we evaluate both no-reuse and fixed-point reuse at the same sampling budgets.

Table~\ref{tab:fp_maskgit_consistency_results_updated} shows that representation-space losses are generally more effective than output-space KL for improving generation quality. In particular, hidden-state MSE with reuse gives the best FID at budgets 48 and 192, while hidden-state cosine with EMA gives the best FID at budget 96. These results support our choice of hidden-state consistency as the main regularizer: it provides a stronger training signal for the fixed-point representation than directly matching logits, and it interacts well with reuse during sampling.

After identifying the three best loss configurations on FP-MaskGIT, we transfer them to FP-MDLM, 120k training steps at 128 length, to test whether they also work in a different modality. We first run this ablation on FP-MaskGIT because it is cheaper to train and evaluate than FP-MDLM, and because its image-generation metrics are more direct than proxy metrics such as generative perplexity and unigram entropy. These configuration are: Hidden state MSE (no EMA), Hidden state cosine (with EMA), Hidden state MSE (with EMA) (Table~\ref{tab:image_gen_ppl_entropy_results_128}). Once we found the best configuration on FP-MDLM, 120k training steps at 128 length, we test if it extends to longer sequences, i.e. 1024.

\FloatBarrier
\begin{table}[H]
\centering
\small
\caption{Ablation on the consistency loss type for FP-MaskGIT.}
\label{tab:fp_maskgit_consistency_results_updated}
\setlength{\tabcolsep}{1.2pt}
\renewcommand{\arraystretch}{0.72}
\sisetup{detect-weight=true, detect-inline-weight=math}

\begin{tabular*}{\linewidth}{@{\extracolsep{\fill}} l l l c c c @{}}
\toprule
& & & \multicolumn{3}{c}{\textbf{Budget}} \\
\cmidrule(l){4-6}
\textbf{Method} & \textbf{Reuse} & \textbf{Metric} & \textbf{48} & \textbf{96} & \textbf{192} \\
\midrule

\multirow{4}{*}{\shortstack[l]{\textbf{Baseline}\\\scriptsize FP-MaskGIT + 10k\\\scriptsize no KL loss}}
& \multirow{2}{*}{No reuse}
& FID $\downarrow$ & 102.3041 & 56.9086 & 32.3490 \\
&
& IS $\uparrow$ & \underline{14.6703} & 15.9080 & 15.0480 \\
\cmidrule(lr){2-6}
& \multirow{2}{*}{With reuse}
& FID $\downarrow$ & 101.5056 & 55.2582 & 31.7141 \\
&
& IS $\uparrow$ & 14.4204 & 15.5139 & 14.5876 \\
\midrule

\multirow{4}{*}{\shortstack[l]{\textbf{Cross-step logit KL}\\\scriptsize no EMA teacher}}
& \multirow{2}{*}{No reuse}
& FID $\downarrow$ & 101.7078 & 55.8024 & 31.4500 \\
&
& IS $\uparrow$ & 14.5171 & 16.0273 & 14.8362 \\
\cmidrule(lr){2-6}
& \multirow{2}{*}{With reuse}
& FID $\downarrow$ & 98.4032 & 53.7002 & 29.4128 \\
&
& IS $\uparrow$ & 14.5321 & 16.0719 & 15.0317 \\
\midrule

\multirow{4}{*}{\shortstack[l]{\textbf{Cross-step logit KL}\\\scriptsize EMA teacher}}
& \multirow{2}{*}{No reuse}
& FID $\downarrow$ & 102.2678 & 56.5554 & 31.2910 \\
&
& IS $\uparrow$ & 14.0976 & 16.0031 & 14.8055 \\
\cmidrule(lr){2-6}
& \multirow{2}{*}{With reuse}
& FID $\downarrow$ & 100.4105 & 54.2807 & 29.7377 \\
&
& IS $\uparrow$ & 14.2942 & 15.9352 & 14.5184 \\
\midrule

\multirow{4}{*}{\shortstack[l]{\textbf{Latent proj cosine}\\\scriptsize no EMA}}
& \multirow{2}{*}{No reuse}
& FID $\downarrow$ & 112.2268 & 55.3464 & 30.5902 \\
&
& IS $\uparrow$ & 13.0748 & 15.9799 & 15.0260 \\
\cmidrule(lr){2-6}
& \multirow{2}{*}{With reuse}
& FID $\downarrow$ & 102.4396 & 54.3828 & 30.4941 \\
&
& IS $\uparrow$ & 14.2156 & 15.7187 & 14.9456 \\
\midrule

\multirow{4}{*}{\shortstack[l]{\textbf{Latent proj cosine}\\\scriptsize with EMA}}
& \multirow{2}{*}{No reuse}
& FID $\downarrow$ & 110.7216 & 54.8109 & 32.3299 \\
&
& IS $\uparrow$ & 13.1563 & 15.6277 & 14.9067 \\
\cmidrule(lr){2-6}
& \multirow{2}{*}{With reuse}
& FID $\downarrow$ & 102.5607 & 53.8555 & 30.6918 \\
&
& IS $\uparrow$ & 14.0288 & 15.5334 & 14.9674 \\
\midrule

\multirow{4}{*}{\shortstack[l]{\textbf{Hidden state MSE}\\\scriptsize no EMA}}
& \multirow{2}{*}{No reuse}
& FID $\downarrow$ & 107.5012 & 56.7336 & 31.3828 \\
&
& IS $\uparrow$ & 13.3413 & \textbf{16.3229} & 15.0487 \\
\cmidrule(lr){2-6}
& \multirow{2}{*}{With reuse}
& FID $\downarrow$ & \textbf{97.8401} & 51.0077 & \textbf{27.6242} \\
&
& IS $\uparrow$ & \textbf{15.0866} & \underline{16.2692} & 14.5894 \\
\midrule

\multirow{4}{*}{\shortstack[l]{\textbf{Hidden state MSE}\\\scriptsize with EMA}}
& \multirow{2}{*}{No reuse}
& FID $\downarrow$ & 110.4611 & 54.2080 & 30.6044 \\
&
& IS $\uparrow$ & 13.4316 & 15.5567 & 14.9206 \\
\cmidrule(lr){2-6}
& \multirow{2}{*}{With reuse}
& FID $\downarrow$ & \underline{97.9066} & 51.1808 & 28.3367 \\
&
& IS $\uparrow$ & 14.6514 & 16.1668 & 14.5935 \\
\midrule

\multirow{4}{*}{\shortstack[l]{\textbf{Hidden state cosine}\\\scriptsize no EMA}}
& \multirow{2}{*}{No reuse}
& FID $\downarrow$ & 110.4582 & 56.6083 & 31.1203 \\
&
& IS $\uparrow$ & 13.7846 & 15.8706 & 14.9106 \\
\cmidrule(lr){2-6}
& \multirow{2}{*}{With reuse}
& FID $\downarrow$ & 98.7089 & 54.5677 & 29.5129 \\
&
& IS $\uparrow$ & 14.4814 & 15.7316 & 14.4452 \\
\midrule

\multirow{4}{*}{\shortstack[l]{\textbf{Hidden state cosine}\\\scriptsize with EMA}}
& \multirow{2}{*}{No reuse}
& FID $\downarrow$ & 112.5791 & 54.7753 & 29.7427 \\
&
& IS $\uparrow$ & 13.0173 & 15.9610 & 14.9954 \\
\cmidrule(lr){2-6}
& \multirow{2}{*}{With reuse}
& FID $\downarrow$ & 98.3099 & \textbf{50.2515} & 28.7888 \\
&
& IS $\uparrow$ & 14.4018 & 15.4198 & 14.7025 \\
\midrule

\multirow{4}{*}{\shortstack[l]{\textbf{Pretraining + hidden state MSE}\\\scriptsize without EMA teacher}}
& \multirow{2}{*}{No reuse}
& FID $\downarrow$ & 114.1953 & 57.1398 & 31.8960 \\
&
& IS $\uparrow$ & 14.0358 & 16.1214 & \textbf{15.4513} \\
\cmidrule(lr){2-6}
& \multirow{2}{*}{With reuse}
& FID $\downarrow$ & 108.5110 & \underline{50.3101} & 28.9113 \\
&
& IS $\uparrow$ & 14.4205 & 16.0090 & 14.8960 \\
\midrule

\multirow{4}{*}{\shortstack[l]{\textbf{Pretraining + hidden state MSE}\\\scriptsize with EMA teacher}}
& \multirow{2}{*}{No reuse}
& FID $\downarrow$ & 116.0607 & 56.6419 & 30.7830 \\
&
& IS $\uparrow$ & 13.3924 & 16.1979 & 15.0050 \\
\cmidrule(lr){2-6}
& \multirow{2}{*}{With reuse}
& FID $\downarrow$ & 108.8888 & 51.9627 & \underline{27.8129} \\
&
& IS $\uparrow$ & 14.3478 & 15.9366 & 14.7259 \\
\midrule

\multirow{4}{*}{\shortstack[l]{\textbf{Pretraining + hidden state cosine}\\\scriptsize without EMA teacher}}
& \multirow{2}{*}{No reuse}
& FID $\downarrow$ & 112.3853 & 55.5995 & 32.6594 \\
&
& IS $\uparrow$ & 13.6743 & 15.9413 & 15.1818 \\
\cmidrule(lr){2-6}
& \multirow{2}{*}{With reuse}
& FID $\downarrow$ & 104.7088 & 53.0664 & 31.0667 \\
&
& IS $\uparrow$ & 14.5433 & 15.9216 & 15.0648 \\
\midrule

\multirow{4}{*}{\shortstack[l]{\textbf{Pretraining + hidden state cosine}\\\scriptsize with EMA teacher}}
& \multirow{2}{*}{No reuse}
& FID $\downarrow$ & 118.5230 & 57.9347 & 32.6617 \\
&
& IS $\uparrow$ & 13.3390 & 16.1273 & 15.2384 \\
\cmidrule(lr){2-6}
& \multirow{2}{*}{With reuse}
& FID $\downarrow$ & 110.3087 & 53.6505 & 30.9416 \\
&
& IS $\uparrow$ & 13.9843 & 16.0756 & \underline{15.3464} \\

\bottomrule
\end{tabular*}
\end{table}
\FloatBarrier

\FloatBarrier
\begin{table}[H]
\centering
\small
\caption{Generation quality across budgets on OWT, 120k training steps, 128 sequence length. Baseline is 120k training steps. Consistency post training uses 100k pretraining steps followed by 20k post-training adaptation steps.}
\label{tab:image_gen_ppl_entropy_results_128}
\setlength{\tabcolsep}{1.2pt}
\renewcommand{\arraystretch}{0.72}
\sisetup{detect-weight=true, detect-inline-weight=math}

\begin{tabular*}{\linewidth}{@{\extracolsep{\fill}} l l l c c c c @{}}
\toprule
& & & \multicolumn{4}{c}{\textbf{Budget}} \\
\cmidrule(l){4-7}
\textbf{Method} & \textbf{Reuse} & \textbf{Metric} & \textbf{96} & \textbf{192} & \textbf{384} & \textbf{768}\\
\midrule

\multirow{4}{*}{\shortstack[l]{\textbf{Baseline}\\\scriptsize FP-MDLM @120k}}
& \multirow{2}{*}{No reuse}
& Gen PPL $\downarrow$ & 421.5819 & 337.1057 & 262.6982 & 245.1979\\
&
& Entropy $\uparrow$   & 4.4004 & 4.3862 & 4.3688 & 4.3571\\
\cmidrule(lr){2-7}

& \multirow{2}{*}{With reuse}
& Gen PPL $\downarrow$ & 425.2954 & 310.0746 & 253.1428 & 225.9641\\
&
& Entropy $\uparrow$   & 4.4033 & 4.3931 & 4.3651 & 4.3455\\
\midrule

\multirow{4}{*}{\shortstack[l]{\textbf{Hidden state MSE}\\\scriptsize w/o EMA teacher}}
& \multirow{2}{*}{No reuse}
& Gen PPL $\downarrow$ & 433.9153 & 338.9548 & 274.9060 & 227.5296\\
&
& Entropy $\uparrow$   & 4.4047 & 4.3851 & 4.3782 & 4.3658\\
\cmidrule(lr){2-7}

& \multirow{2}{*}{With reuse}
& Gen PPL $\downarrow$ & 385.6467 & 266.6036 & 226.6500 & 215.4439\\
&
& Entropy $\uparrow$   & 4.3992 & 4.3763 & 4.3668 & 4.3705\\
\midrule

\multirow{4}{*}{\shortstack[l]{\textbf{Hidden state cosine}\\\scriptsize w/ EMA teacher}}
& \multirow{2}{*}{No reuse}
& Gen PPL $\downarrow$ & 435.8042 & 334.9215 & 277.8421 & 243.6187\\
&
& Entropy $\uparrow$   & 4.4051 & 4.3832 & 4.3791 & 4.3618\\
\cmidrule(lr){2-7}

& \multirow{2}{*}{With reuse}
& Gen PPL $\downarrow$ & 418.2375 & 306.8842 & 249.7164 & 225.9032\\
&
& Entropy $\uparrow$   & 4.4014 & 4.3841 & 4.3681 & 4.3673\\
\midrule

\multirow{4}{*}{\shortstack[l]{\textbf{Hidden state MSE}\\\scriptsize w/ EMA teacher}}
& \multirow{2}{*}{No reuse}
& Gen PPL $\downarrow$ & 436.9225 & 332.7336 & 279.9604 & 251.2712\\
&
& Entropy $\uparrow$   & 4.4063 & 4.3819 & 4.3804 & 4.3579\\
\cmidrule(lr){2-7}

& \multirow{2}{*}{With reuse}
& Gen PPL $\downarrow$ & 426.5467 & 317.1626 & 254.6232 & 228.4749\\
&
& Entropy $\uparrow$   & 4.4022 & 4.3878 & 4.3692 & 4.3661\\

\bottomrule
\end{tabular*}
\end{table}
\FloatBarrier

\subsection{Tradeoff Between Denoising Steps and Fixed-Point Iterations}
\label{appendix:tradeoff-steps-iter}

This sweep studies how generation quality changes when compute is allocated differently between denoising steps and fixed-point iterations. We measure this tradeoff on FP-MDLM + $\mathcal{L}_{\mathrm{CONS}}$ + 3SR, which corresponds to our CoFRe model. The heatmap shows that these two sources of compute are not interchangeable: using very few denoising steps gives poor generative perplexity even when many FP iterations are used, while increasing the number of denoising steps substantially improves sample quality. Once enough denoising steps are used, additional FP iterations still help, but with smaller gains. The entropy heatmap shows that the best generative perplexity values are not obtained by collapsing sample diversity, since entropy remains in a similar non-degenerate range across the strongest configurations. Overall, the sweep indicates that the quality--diversity tradeoff depends on how inference compute is split between outer denoising and inner fixed-point solving, supporting the use of non-uniform depth allocation at sampling time.
\FloatBarrier
\begin{figure}[H]
    \centering
    \includegraphics[width=\textwidth]{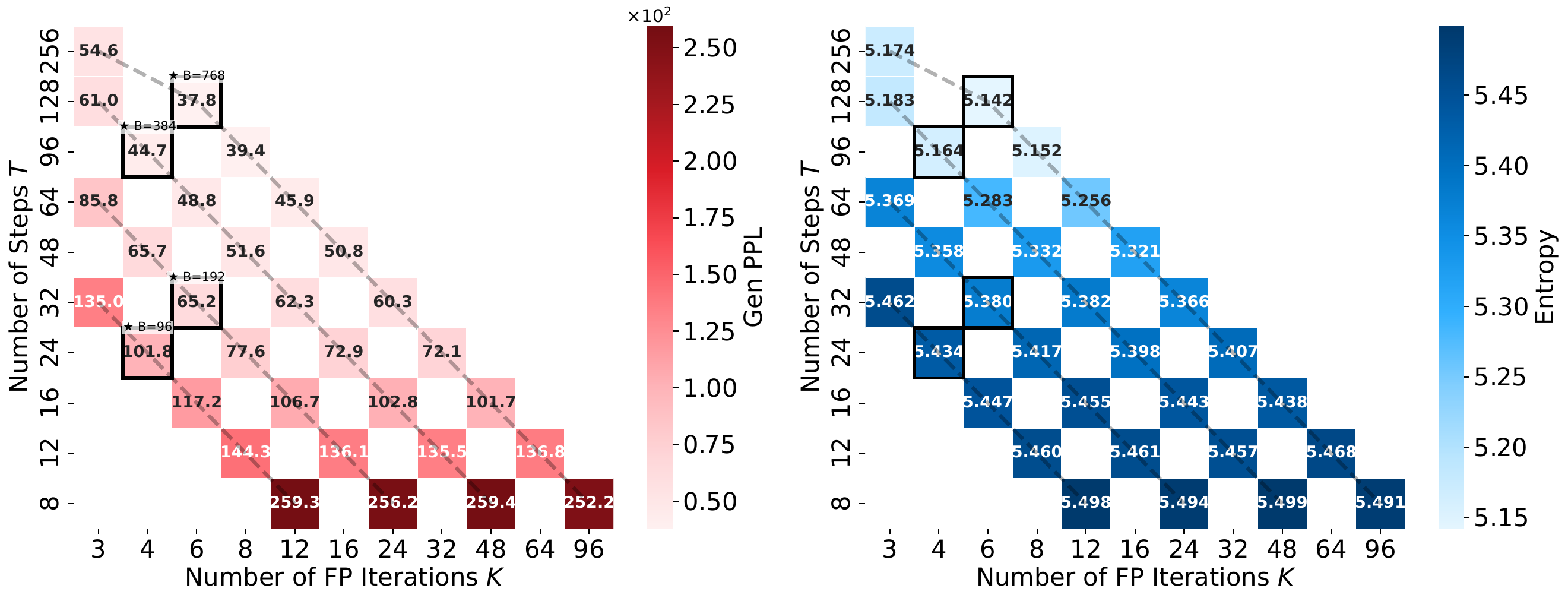}
    \caption{\textbf{Tradeoff between denoising steps and fixed-point iterations.} We sweep the number of denoising steps and FP iterations for FP-MDLM with consistency regularization and three-state reuse. The left heatmap reports generative perplexity, where lower is better, and the right heatmap reports sample entropy. Each annotated cell corresponds to one evaluated allocation; blank cells were not evaluated. Allocating compute to very few denoising steps leads to poor generative perplexity even with many FP iterations, whereas configurations with sufficiently many denoising steps and moderate FP depth achieve the best quality while maintaining non-collapsed entropy.}
    \label{fig:heatmap_tradeoff}
\end{figure}
\FloatBarrier

\subsection{Different budget allocation strategies}
\label{appendix:budget_allocation_strategies}

\begin{figure}[H]
    \centering

    \begin{subfigure}{0.32\textwidth}
        \centering
        \includegraphics[width=\linewidth]{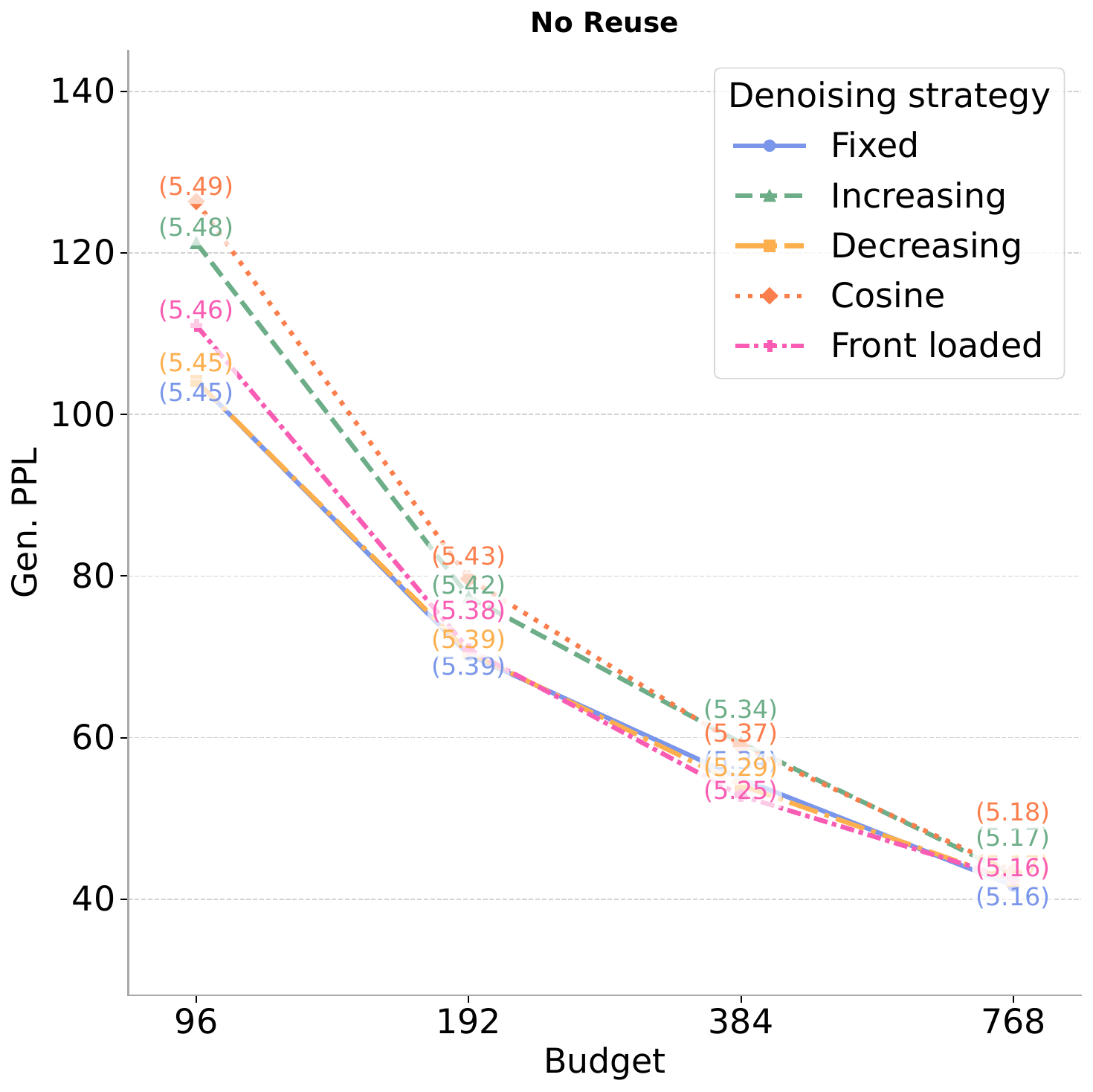}
        \caption{No Reuse}
        \label{fig:denoising_budget_nr}
    \end{subfigure}
    \hfill
    \begin{subfigure}{0.32\textwidth}
        \centering
        \includegraphics[width=\linewidth]{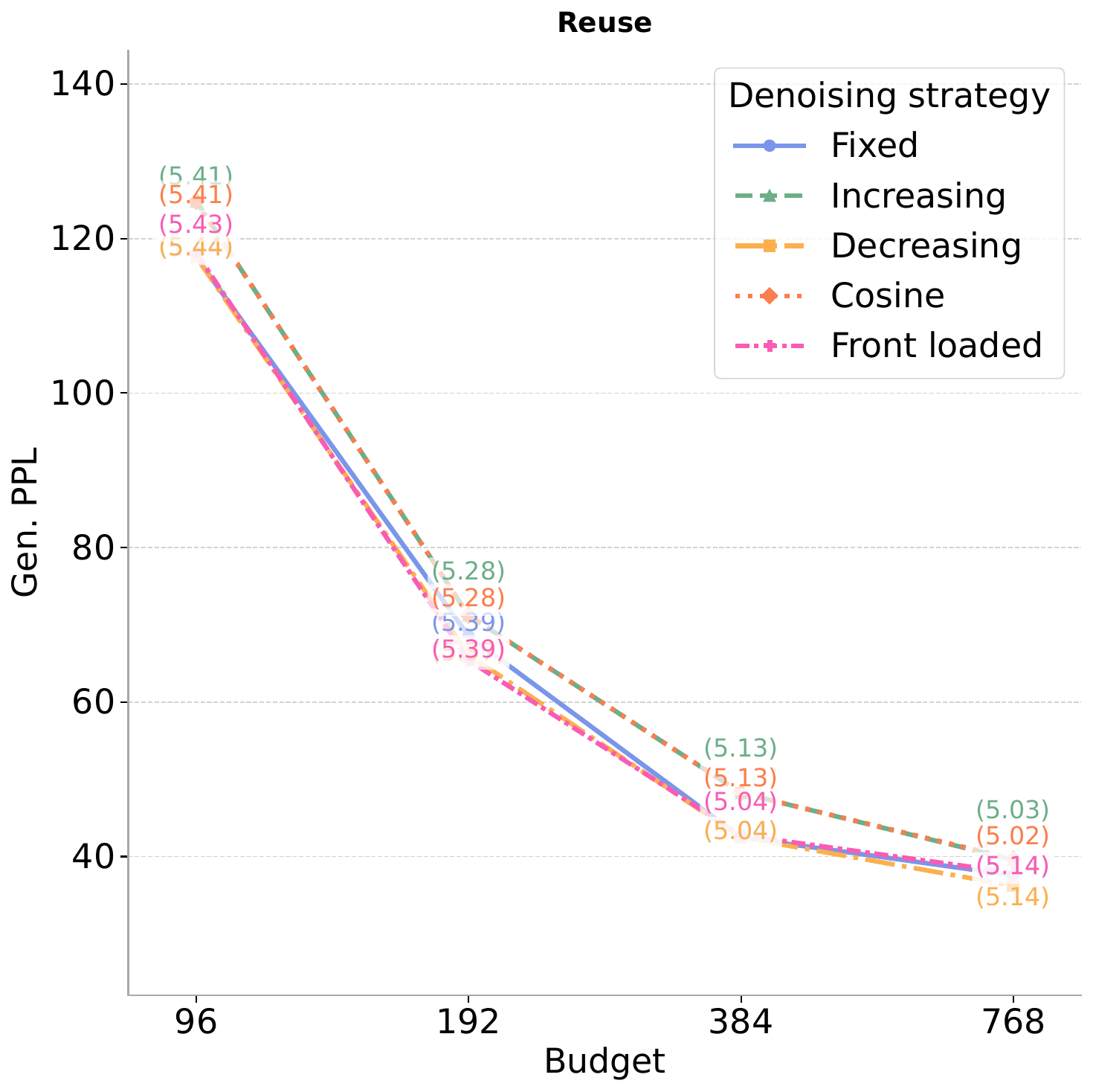}
        \caption{Reuse}
        \label{fig:denoising_budget_r}
    \end{subfigure}
    \hfill
    \begin{subfigure}{0.32\textwidth}
        \centering
        \includegraphics[width=\linewidth]{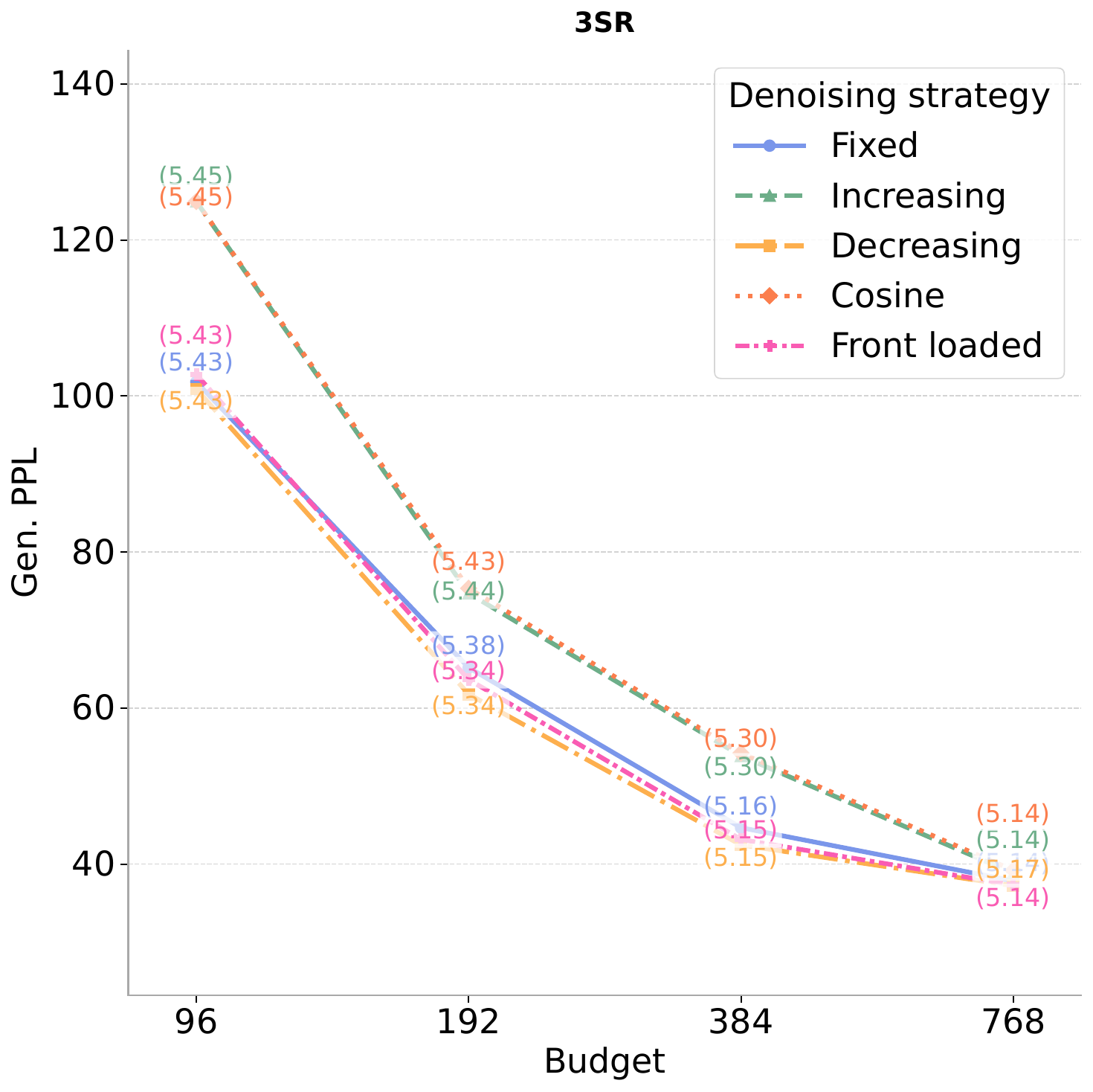}
        \caption{3SR}
        \label{fig:denoising_budget_3sr}
    \end{subfigure}

    \caption{\textbf{Generative perplexity as a function of budget for different denoising strategies.} Entropy values are shown in parentheses. We observe that decreasing schedules perform best overall.}
    \label{fig:denoising_budget_all}
\end{figure}

We also ablate how the fixed-point iteration budget should be distributed across denoising steps for FP-MDLM+$L_{\mathrm{CONS}}$. Keeping the checkpoint and total forward-pass budget fixed, we compare five allocation strategies: fixed, decreasing, increasing, cosine, and front-loaded schedules. We evaluate each schedule under the three initialization regimes used throughout the paper: no reuse, full reuse, and 3SR. Table~\ref{tab:denoising_strategy_overall} and Figure~\ref{fig:denoising_budget_all} show that decreasing schedules perform best overall, with the lowest mean generative perplexity, the best average rank, and the most wins across initialization-budget settings. Fixed allocation is a strong second, while increasing and cosine schedules consistently underperform. This suggests that FP-MDLM benefits from spending more fixed-point computation early in the denoising trajectory, when the sequence is most corrupted and the denoising problem is hardest.

\FloatBarrier
\begin{table}[H]
\centering
\small
\caption{Overall comparison of denoising strategies. Lower Gen. PPL and lower average rank are better. Wins count the number of init-budget settings where the strategy obtains the best Gen. PPL. More than 12 wins are possible in total because ties count as wins for all tied strategies.}
\label{tab:denoising_strategy_overall}
\setlength{\tabcolsep}{1.2pt}
\renewcommand{\arraystretch}{0.72}
\sisetup{detect-weight=true, detect-inline-weight=math}

\begin{tabular*}{\linewidth}{@{\extracolsep{\fill}} l c c c @{}}
\toprule
\textbf{Strategy}
& \textbf{Mean Gen. PPL $\downarrow$}
& \textbf{Avg. rank $\downarrow$}
& \textbf{Wins} \\
\midrule

\textbf{Decreasing} & \textbf{64.65} & \textbf{1.33} & \textbf{8 / 12} \\
Fixed              & 65.56          & 2.00          & 5 / 12 \\
Front loaded       & 65.69          & 2.42          & 2 / 12 \\
Increasing         & 73.10          & 4.08          & 0 / 12 \\
Cosine             & 73.84          & 4.58          & 0 / 12 \\

\bottomrule
\end{tabular*}
\end{table}
\FloatBarrier

\FloatBarrier

\subsection{Latency and generation quality for language modeling}
\label{appendix:latency-res}
We measure generation-only sampling latency, defined as the wall-clock time from fully masked token IDs to final generated token IDs. The timed region includes all denoising-loop computation: model forward passes, fixed-point iterations, reuse/CoFRe (3SR) initialization, mask-confidence updates, categorical sampling, and loop logic. We synchronize CUDA before and after sampling, and exclude decoding, external Gen. PPL evaluation, file I/O, model loading, and warmup/compilation effects. All methods are evaluated under matched transformer-block budgets with the same ancestral-cache sampler, batch size, sequence length, precision, tokenizer, device, and number of samples. We use two warmup batches followed by 20 timed batches.

\begin{figure}[H]
    \centering
    \includegraphics[width=0.4\textwidth]{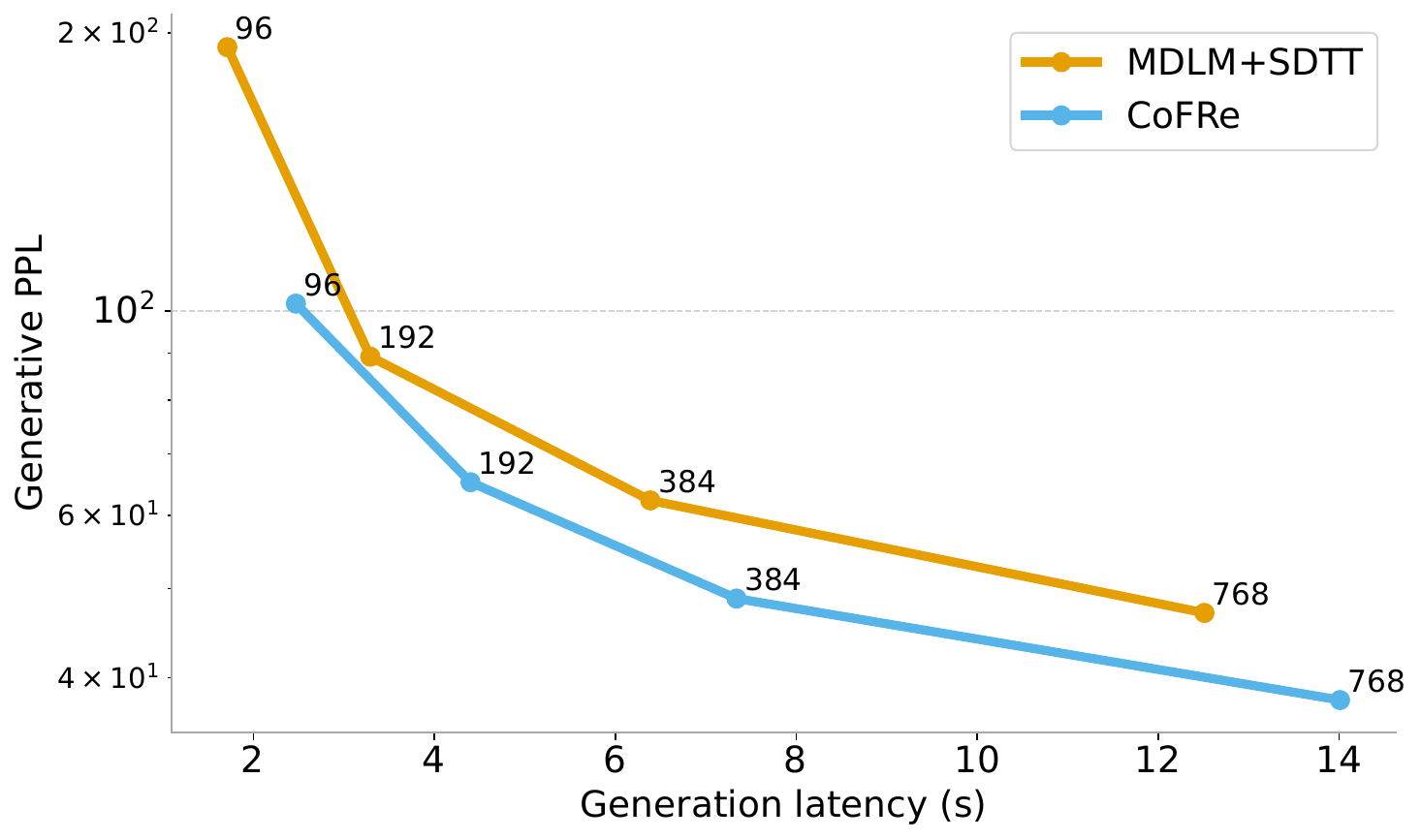}
    \caption{\textbf{Generation quality as a function of wall-clock sampling latency on OWT.}
We report generation-only latency, measured from fully masked token IDs to final generated token IDs, excluding decoding and external Gen. PPL evaluation. Points are annotated by their transformer-block budget. CoFRe is modestly slower than MDLM+SDTT at matched budget, but reaches substantially lower Gen. PPL at lower wall-clock latency in the low- and medium-budget regimes.}
    \label{fig:latency_vs_genppl}
\end{figure}

Table~\ref{tab:sampling_latency_quality} and Figure~\ref{fig:latency_vs_genppl} show that CoFRe is modestly slower than MDLM+SDTT at equal transformer-block budget, with slowdowns between $1.12\times$ and $1.45\times$, but achieves substantially better generation quality. For example, at budget 96, CoFRe reduces Gen. PPL from 193.05 to 101.791 with latency increasing from 1.71s to 2.47s. At budget 384, it improves Gen. PPL from 62.29 to 48.755 with only a $1.15\times$ slowdown. Overall, CoFRe is not faster than MDLM+SDTT at equal budget in the current implementation, but provides a better latency--quality trade-off, reaching Gen. PPL below 100 in 4.40-14.01s, while MDLM+SDTT does not reach this quality at any tested budget.

\FloatBarrier
\begin{table}[H]
\centering
\caption{Generation-only sampling latency and quality on OWT. Latency is measured from fully masked token IDs to final generated token IDs, excluding tokenizer decoding and external Gen. PPL evaluation.}
\label{tab:sampling_latency_quality}
\small
\setlength{\tabcolsep}{5pt}
\begin{tabular*}{\linewidth}{@{\extracolsep{\fill}} r c c c c c @{}}
\toprule
\textbf{Budget}
& \multicolumn{2}{c}{\textbf{Latency (s) $\downarrow$}}
& \textbf{Slowdown}
& \multicolumn{2}{c}{\textbf{Gen. PPL $\downarrow$}} \\
\cmidrule(lr){2-3}
\cmidrule(lr){5-6}
& \textbf{MDLM+SDTT}
& \textbf{CoFRe}
& \textbf{CoFRe / MDLM+SDTT}
& \textbf{MDLM+SDTT}
& \textbf{CoFRe} \\
\midrule
96  & 1.71  & 2.47  & 1.45$\times$ & 193.05 & 101.791 \\
192 & 3.29  & 4.40  & 1.34$\times$ & 89.17 & 65.182  \\
384 & 6.39  & 7.34  & 1.15$\times$ & 62.29 & 48.755  \\
768 & 12.51 & 14.01 & 1.12$\times$ & 47.04 & 37.846 \\
\bottomrule
\end{tabular*}
\end{table}

\FloatBarrier

\FloatBarrier

\subsection{Fixed-point residual analysis.}
\begin{figure}[H]
    \centering
    \includegraphics[width=0.48\textwidth]{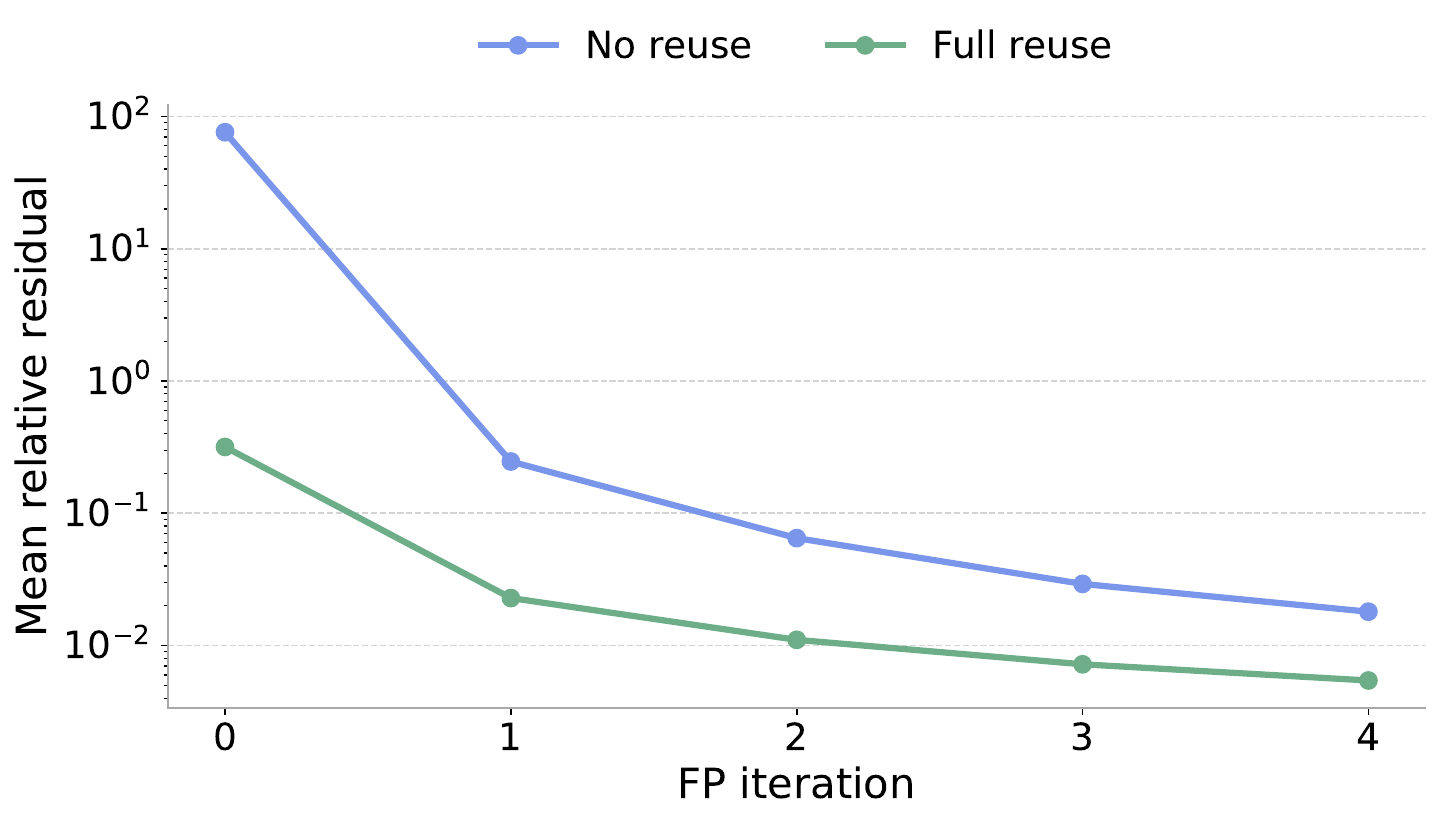}
    \includegraphics[width=0.48\textwidth]{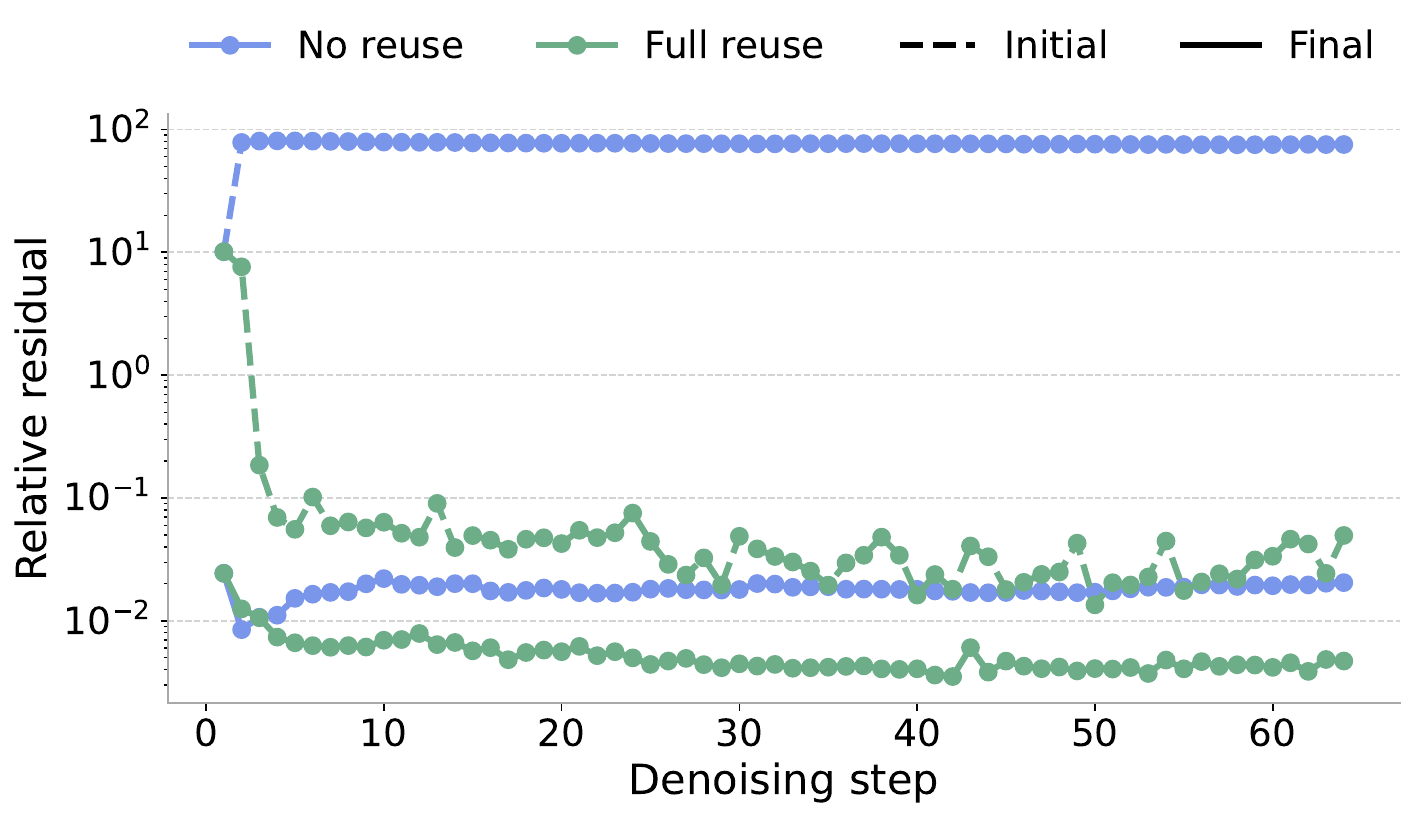}
    \caption{\textbf{Fixed-point residual analysis.}
    (Left) Mean relative residual decreases with FP iterations, showing that the repeated block approaches a fixed point. Full reuse starts much closer to the solution than no reuse. 
    (Right) Across denoising steps, reuse strongly reduces the initial residual and yields lower final residuals under the same iteration budget. Residuals validate the solver and warm-start mechanism, while generation ablations are needed to compare full reuse and 3SR quality.}
    \label{fig:fp_residual_analysis}
\end{figure}

We measure the relative hidden residual
\[
r^{(n)}_t =
\frac{\|F_\theta(h^{(n)}_t; \tilde h_t, t)-h^{(n)}_t\|_2}
{\|h^{(n)}_t\|_2}
\]
across FP iterations and denoising steps. As shown in Figure~\ref{fig:fp_residual_analysis}, residuals decrease steadily with iterations, confirming that the repeated block behaves as an iterative fixed-point solver. Without reuse, the mean residual drops from $75.97$ to $0.0180$ after four iterations; with full reuse, it starts much closer and drops from $0.317$ to $0.00545$. Excluding the first step, full reuse reduces the median initial residual from $76.68$ to $0.0385$, i.e. it starts about $1984\times$ closer to the current fixed point. These results validate the solver and the benefit of warm starts. However, lower residual does not necessarily imply better generation: full reuse is numerically closest to equilibrium, but can over-reuse stale states, motivating the token-aware 3SR rule.

\FloatBarrier
\FloatBarrier

\subsection{Effect of Training Duration on Generation Quality}

In this section, we compare the effect of the training duration on generation quality. We compare here MDLM trained at 100k steps, MDLM trained at 1M steps (checkpoint from \citep{sahoo_simple_2024}), FP-MDLM trained at 100k steps, FP-MDLM trained at 1M steps and FP-MDLM trained at 100k + $\mathcal{L}_{\mathrm{CONS}}$ +3SR (so basically CoFRe at 100k training steps). For all the models, we evaluate both the generative perplexity and the unigram entropy. We report the results in Table~\ref{tab:fp_mdlm_training_stage_comparison}.

\FloatBarrier
\begin{table}[H]
\centering
\small
\caption{\textbf{Training-stage comparison for MDLM and FP-MDLM generation on OpenWebText.}
We report generative perplexity and entropy across sampling budgets. The table compares MDLM and FP-MDLM checkpoints at different training stages, together with the FP-MDLM checkpoint trained with cross-step consistency regularization and evaluated with three-state reuse.}
\label{tab:fp_mdlm_training_stage_comparison}
\setlength{\tabcolsep}{1.2pt}
\renewcommand{\arraystretch}{0.72}
\sisetup{detect-weight=true, detect-inline-weight=math}

\begin{tabular*}{\linewidth}{@{\extracolsep{\fill}} l l c c c c @{}}
\toprule
& & \multicolumn{4}{c}{\textbf{Budget}} \\
\cmidrule(l){3-6}
\textbf{Method} & \textbf{Metric} & \textbf{96} & \textbf{192} & \textbf{384} & \textbf{768}\\
\midrule

\multirow{2}{*}{\shortstack[l]{MDLM\\@100k}}
& Gen PPL $\downarrow$ & 642.425 & 277.504 & 177.962 & 134.952 \\
& Entropy $\uparrow$   & 5.866 & 5.779 & 5.730 & 5.693 \\
\cmidrule(lr){1-6}

\multirow{2}{*}{MDLM}
& Gen PPL $\downarrow$ & 830.820 & 343.330 & 196.790 & 143.880 \\
& Entropy $\uparrow$   & 5.910 & 5.810 & 5.750 & 5.700 \\
\midrule

\multirow{2}{*}{\shortstack[l]{FP-MDLM\\@100k}}
& Gen PPL $\downarrow$ & 379.484 & 268.113 & 213.801 & 180.315 \\
& Entropy $\uparrow$   & 5.790 & 5.755 & 5.722 & 5.694 \\
\cmidrule(lr){1-6}

\multirow{2}{*}{FP-MDLM}
& Gen PPL $\downarrow$ & 375.631 & 273.275 & 215.197 & 179.655 \\
& Entropy $\uparrow$   & 5.810 & 5.763 & 5.726 & 5.702 \\
\cmidrule(lr){1-6}

\multirow{2}{*}{\shortstack[l]{FP-MDLM @100k\\+ $\mathcal{L}_{\mathrm{CONS}}$ + 3SR}}
& Gen PPL $\downarrow$ & 96.802 & 61.891 & 49.669 & 37.717 \\
& Entropy $\uparrow$   & 5.519 & 5.411 & 5.409 & 5.246 \\

\bottomrule
\end{tabular*}
\end{table}
\FloatBarrier

We observe that the generative quality after 100k steps closely match those at 1M steps, suggesting that 100k steps are likely sufficient to test new algorithms or framework when pretraining a model. If a model fails by that time, it is unlikely to succeed at 1M steps. Further analysis is needed to determine whether this reflects limitations of the metrics (generative perplexity and entropy may not capture differences at that scale) or constraints imposed by model capacity. \citet{pynadath2026generativefrontiersevaluationmatters} observes similar results at 50k steps when analyzing the generative frontiers of these models.

\newpage

\end{document}